\documentclass{article}
\usepackage{arxiv}

\usepackage{amsmath}
\usepackage{unicode-math}

\AtBeginDocument{\renewcommand{\mathbf}[1]{\symbfup{#1}}}

\usepackage{booktabs,array,longtable,tabularx,multirow}
\usepackage[labelsep=period]{caption}
\usepackage{graphicx}
\usepackage{float}

\usepackage[hidelinks]{hyperref}
\usepackage{etoolbox}
\makeatletter
\patchcmd{\@maketitle}{\center{\today}}{}{}{}
\makeatother
\renewcommand{\keywordname}{\textbf{Keywords:}}
\title{Factorized axis convolutional gated recurrent unit with dynamic adaptive pooling for remaining useful life prediction of rolling bearings}

\author{
	Hanbyeol Park \\
	Department of Industrial Engineering\\
	Pusan National University\\
	Pusan, Republic of Korea\\
	\texttt{pb104@pusan.ac.kr} \\
	\And
	Jungho Choo \\
	Department of Industrial Engineering\\
	Pusan National University\\
	Pusan, Republic of Korea\\
	\texttt{jhchoo@pusan.ac.kr} \\
	\And
	Hyerim Bae \\
	Department of Industrial Engineering\\
	Pusan National University\\
	Pusan, Republic of Korea\\
	\texttt{hrbae@pusan.ac.kr} \\
}

\begin{document}
\maketitle

\begin{abstract}
    Convolutional neural networks (CNN) are widely used to predict the remaining useful life (RUL) of rolling bearings from time--frequency representations (TFRs) of vibration signals. However, during degradation, characteristic structures in TFRs align predominantly along the frequency or time axis, making it challenging for conventional CNN isotropic kernels to capture directional structure. Furthermore, global average pooling (GAP) averages across axes, potentially obscuring the locations and concentrations of salient activations. This study introduces a factorized-axis convolutional gated recurrent unit (GRU) that employs multiscale anisotropic convolution and a dual-axis convolution block attention module to enhance directional features and highlight salient time--frequency regions. Dynamic adaptive pooling (DAP) adaptively aggregates the time--frequency-axis information from the extracted feature maps, whereas a GRU captures temporal dynamics in the latent representations and Monte Carlo dropout enables predictive uncertainty estimation. Experiments on two public bearing datasets demonstrate that the proposed model outperforms existing RUL prediction methods across operating conditions. Ablation experiments demonstrate that the factorized axis-wise design achieves lower mean errors than conventional isotropic kernels. DAP yields clear improvements on one dataset while matching GAP on the other, highlighting the importance of anisotropic feature extraction and adaptive feature aggregation for TFR-based RUL prediction.
\end{abstract}


\section{Introduction}
\label{sec:introduction}

Rolling bearings are essential components in rotating machinery such as electric motors and aerospace equipment \cite{Zhang_2023_Digital}. They facilitate relative motion, transmit loads, and minimize friction \cite{Rejith_2023_Bearings}. Over extended periods of operation, these bearings are subjected to mechanical and thermal stresses, variations in lubrication, and adverse environmental conditions \cite{Nandi_2005_Condi}. The interplay of these factors can result in progressive performance degradation and, ultimately, failure \cite{Gao_2024_Long}. Bearing failures constitute a significant proportion of rotating machinery breakdowns \cite{Nandi_2005_Condi}, leading to reduced operational efficiency and, in some cases, posing significant safety risks. Consequently, the accurate prediction of the remaining useful life (RUL) of rolling bearings is vital for ensuring reliable and efficient machinery operation, enhancing safety \cite{Hou_2022_High}.

RUL prediction methodologies are generally classified into physics-based \cite{QIU_2002_DAMAGE,Gazizulin_2015_Towards,Gabrielli_2024_Phys}, data-driven \cite{Deng_2024_Hybrid,Niazi_2024_Multi,Zhong_2025_RULMulti}, or hybrid \cite{Lv_2024_Hybrid,Chen_2025_Hybrid,He_2025_Phys} approaches. Physics-based methods model failure and degradation mechanisms through physical equations, estimating system states and model parameters based on observed data \cite{Zhong_2025_RULMulti}. When the underlying physical models and loading conditions adequately reflect the actual degradation process, these methods offer high interpretability and predictive performance \cite{Cubillo_2016_Review}. However, their performance is sensitive to model-form errors, simplifying assumptions, and parameter uncertainties, and their development requires specialized expertise in failure mechanisms \cite{An_2015_Prac,Li_2024_Review}. By contrast, data-driven approaches do not rely on explicit physics-based degradation models; instead, they infer relationships between historical condition-monitoring data and health states or RUL \cite{Deng_2024_Hybrid}. Their reliability depends on the quantity, quality, and representativeness of the training data, as well as their ability to generalize to previously unseen operating conditions \cite{Fink_2020_Poten,Ayman_2025_Feat}. Deep learning (DL) models can learn nonlinear representations from high-dimensional sensor time series, enabling the integration of feature extraction and RUL regression within a unified end-to-end architecture \cite{Li_2018_DCNN}. These approaches alleviate the need for constructing system-specific physical degradation models and effectively capture complex nonlinear relationships \cite{Ayman_2025_Feat}. Hybrid methods, which combine physical models, constraints, and prior knowledge with data-driven techniques, aim to address the limitations inherent in each individual approach \cite{Chen_2025_Hybrid}. The performance of hybrid models is contingent upon the accuracy of the incorporated physical knowledge and the integration strategy employed. However, these methods often retain much of the complexity associated with constructing physical models and require substantial domain expertise.

Data-driven RUL prediction leverages various input representations, including raw vibration signals \cite{Wang_2019_Deepse}, handcrafted time- and frequency-domain feature vectors \cite{Zhao_2021_Feature}, and time--frequency representations (TFRs) \cite{Niazi_2024_Multi}. Raw vibration signals preserve the waveform dynamics; however, simple statistical features may overlook important temporal patterns and spectral evolution, particularly given the nonstationary nature of bearing vibration during degradation \cite{Ma_2020_DeepW}. Handcrafted time- and frequency-domain features can capture variations in average energy and frequency components \cite{Yoo_2018_Novel}; however, their effectiveness depends on the manual selection of informative features tailored to specific signal characteristics \cite{Ayman_2025_Feat}. TFRs provide joint localization of spectral energy in both time and frequency domains, making them well-suited for representing the nonstationary and impulsive behavior observed in bearing vibration signals during degradation \cite{Feng_2013_Recent}. For example, Liu et al. \cite{Liu_2022_SALCNN} employed a short-time Fourier transform (STFT)-based TFR as input to an end-to-end RUL prediction framework, integrating a long short-term memory (LSTM) network and convolutional block attention module (CBAM) with a convolutional neural network (CNN). The STFT characterizes the time--frequency structure of nonstationary signals through windowed local frequency analysis. Compared with a conventional CNN, the resulting CNN--LSTM--CBAM architecture has demonstrated enhanced predictive performance by more effectively modeling temporal dependencies in vibration sequences. However, STFT computed with a fixed window imposes a single, global time--frequency resolution, necessitating a trade-off between temporal and frequency resolution \cite{Shi_2025_Dynamic}. This limitation has motivated subsequent studies using continuous wavelet transform (CWT)-based TFRs. For example, Yoo et al. \cite{Yoo_2018_Novel} constructed CWT-based TFRs, generated a health indicator using a CNN, and combined this with Gaussian process-based degradation modeling to predict the RUL. Niazi et al. \cite{Niazi_2024_Multi} advanced spatiotemporal modeling of multidimensional features extracted from the time, frequency, and time--frequency domains by combining a convolutional LSTM (ConvLSTM)-based CWT feature extractor featuring a parallel neural network, transformer, and recurrent neural network. Deng et al. \cite{Deng_2024_Hybrid} further improved bearing RUL prediction by constructing CWT-based TFRs and introducing a novel architecture that integrates a multilayer perceptron (MLP), DeepAR, and a transformer. Notably, their approach leveraged DeepAR to explicitly account for the predictive uncertainty. Overall, research on TFR-based RUL prediction has progressed toward the construction of multiscale TFRs and the application of DL models, particularly CNN-based architectures, to learn degradation-related representations and temporal dependencies.

Despite these advancements, many existing studies continue to utilize conventional CNN-based feature extractors \cite{Liu_2022_SALCNN}, which typically employ square kernels. These kernels span an equal number of grid positions along the time and frequency axes, thereby failing to explicitly encode axis-specific scale priors. Because the two axes correspond to distinct physical quantities, degradation-related patterns may demonstrate different characteristic extents along either axis. Consequently, applying identical kernel dimensions to both axes can limit the ability of the network to capture axis-dependent features. Furthermore, conventional CNNs typically employ global average pooling (GAP) to generate fixed-length representations. GAP computes an unweighted mean across all time--frequency locations, causing activations that are localized within a limited time interval or frequency bands to be diluted across the entire feature map, regardless of their relevance to degradation. These architectural limitations highlight the need for axis-wise convolution and input-dependent weighted aggregation for TFR-based bearing RUL prediction.

\begin{figure}[htbp]
	\centering
	\includegraphics[width=\linewidth]{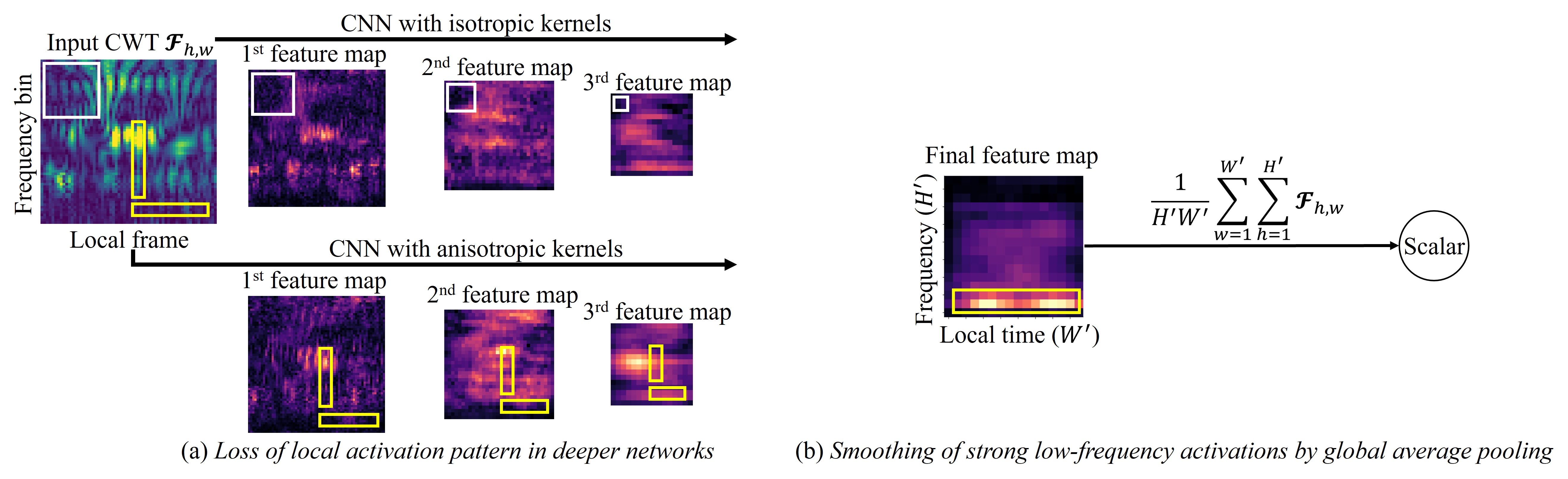}
	\caption{Motivation for axis-wise kernels and input-dependent aggregation in TFR feature maps. (a) Isotropic convolutional kernels tend to progressively weaken strong, localized activations, whereas anisotropic kernels preserve them more effectively. (b) GAP assigns equal weights to all time--frequency locations and thus does not distinguish between degradation-relevant and irrelevant frequency regions.}
	\label{fig:figure1}
\end{figure}

The motivations for these architectural choices are shown in Figure~\ref{fig:figure1}. A comparative analysis of the feature maps obtained from the same TFR input using square and anisotropic kernels is shown in Figure~\ref{fig:figure1}(a). The example demonstrates that localized activation in the mid-frequency region became attenuated in deeper layers when square kernels were used. This qualitative comparison motivated the use of different kernel extents along the two axes. The aggregation of the final feature map into a vector is shown in Figure~\ref{fig:figure1}(b). The GAP averages over both spatial axes within each channel, resulting in a single scalar per channel and eliminating explicit information regarding the locations of strong activations, including those in the low-frequency region, as shown in the figure.

Motivated by these observations, this study introduces the factorized-axis convolutional gated recurrent unit (FAAC-GRU). Model (i) extracts local degradation-related features using multi-scale anisotropic convolution (MSAC), with different kernel extents along the time and frequency axes; (ii) emphasizes salient directional regions through a dual-axis convolutional block attention module (DCBAM); and (iii) aggregates the resulting feature maps using dynamic adaptive pooling (DAP), which applies input-dependent attention weights for adaptive feature integration. Collectively, MSAC, DCBAM, and DAP constitute the FAAC block. A shared FAAC block processes each segment-wise TFR to extract embeddings, which are then concatenated to form a representation for each observation record. Temporal dependencies across these record-level representations are modeled using a GRU, and a regression head subsequently maps the GRU output to a scalar RUL estimate. During inference, Monte Carlo (MC) dropout is employed in the regression head to quantify prediction uncertainty arising from dropout-induced variability. The proposed approach is benchmarked against several comparison methods on two publicly available bearing datasets under multiple operating conditions.

The remainder of this paper is organized as follows. Section~\ref{sec:preliminary} reviews the mathematical foundations of CWT, CBAM, and MC dropout. Section~\ref{sec:method} details the input preprocessing steps and formalizes the proposed FAAC-GRU architecture. Section~\ref{sec:experiments} outlines the experimental protocol and presents the results. Section~\ref{sec:ablat} provides detailed ablation studies of the FAAC-GRU. Finally, Section~\ref{sec:conclus} concludes the paper and outlines directions for future research.

\section{Preliminary}
\label{sec:preliminary}

This section summarizes the definitions and computations of the CWT, CBAM, and MC dropout, which underpin the input representation and architectural design of the proposed model.

\subsection{Continuous wavelet transform}

CWT is a signal processing technique for analyzing singularities and periodic structures in signals \cite{Mallat_1992_Singular}. By representing a signal through translated and scaled versions of a mother wavelet, the CWT enables the examination of localized signal variations across multiple scales \cite{Wei_2025_RUL}. For a vibration record comprising \(P\) samples, let \(x(u)\) denote the continuous-time representation. With \(a > 0\) as the scale parameter, \(b\) as the translation parameter, and \(\psi\) as the mother wavelet, the CWT coefficient is defined as follows:

\begin{equation}
    W(a,b) = \frac{1}{\sqrt{a}}\int_{- \infty}^{\infty}{x(u)\psi^{*}\left( \frac{u - b}{a} \right)du}
    \tag{1}
    \label{eq:1}
\end{equation}

where \(u\) denotes the integration variable and * denotes complex conjugation. In this study, the Morlet wavelet is employed, and the input representation is constructed from the coefficient magnitude \(\left| W(a,b) \right|\), referred to as a magnitude scalogram. The Morlet wavelet is expressed as follows:

\begin{equation}
    \psi(u) = \pi^{- \frac{1}{4}}e^{i\omega_{0}u}e^{- \frac{u^{2}}{2}}
    \tag{2}
    \label{eq:2}
\end{equation}

where \(\psi(u)\) represents the wavelet function of Morlet, \(\pi^{- \frac{1}{4}}\) represents the normalization coefficient, \(i\) represents the imaginary unit, \(\omega_{0}\) represents the angular frequency parameter of the wavelet, \(u\) represents the time.

\subsection{Convolutional block attention module}

The CBAM is a feature-gating mechanism that sequentially applies channel attention (CA) and spatial attention (SA) to enhance feature representations. The CA module weights each feature channel, whereas the SA module emphasizes the spatial locations. These operations can enhance feature representations by emphasizing informative features and suppressing less-relevant responses \cite{Woo_2018_CBAM}.

\begin{figure}[htbp]
	\centering
	\includegraphics[width=0.7\linewidth]{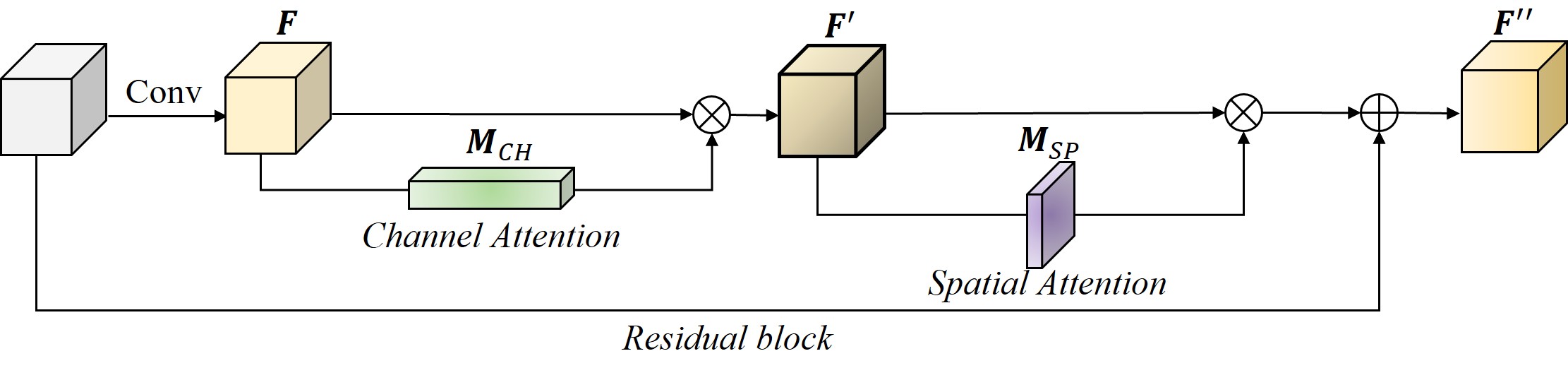}
	\caption{Architecture of CBAM.}
	\label{fig:figure2}
\end{figure}

\begin{figure}[htbp]
	\centering
	\includegraphics[width=\linewidth]{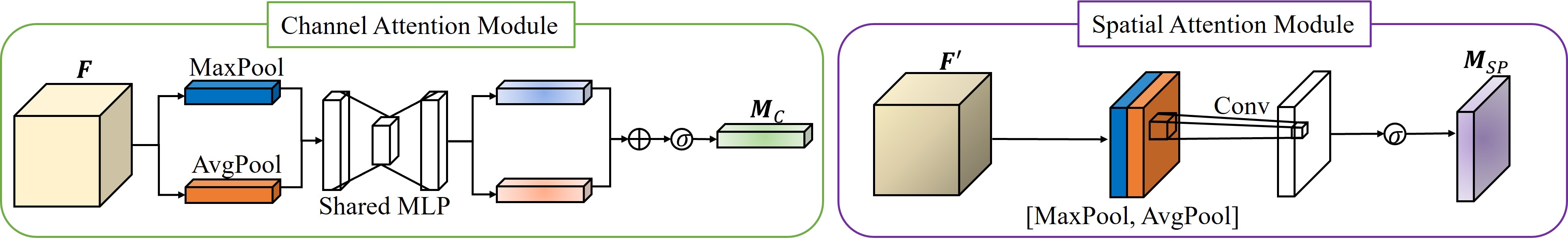}
	\caption{Architectures of the CA and SA modules.}
	\label{fig:figure3}
\end{figure}

Given an intermediate feature map \(\mathbf{F} \in \mathbb{R}^{C \times H \times W}\) as input, CBAM sequentially infers a one-dimensional channel attention map \(\mathbf{M}_{C} \in \mathbb{R}^{C \times 1 \times 1}\) and a two-dimensional spatial attention map \(\mathbf{M}_{S} \in \mathbb{R}^{1 \times H \times W}\) as shown in Figure~\ref{fig:figure2}. The overall attention process is expressed as follows:

\begin{equation}
    \mathbf{F}^{'} = \mathbf{M}_{C}\left( \mathbf{F} \right) \odot \mathbf{F}
    \tag{3}
    \label{eq:3}
\end{equation}

\begin{equation}
    \mathbf{F}^{''} = \mathbf{M}_{S}\left( \mathbf{F}^{'} \right) \odot \mathbf{F}^{'}
    \tag{4}
    \label{eq:4}
\end{equation}

where \(\odot\) denotes element-wise multiplication. The CA weights are broadcast over spatial dimensions, whereas the SA weights are broadcast over channel dimensions. The final refined feature map is denoted as \(\mathbf{F}^{''}\). The computation process for the two attention modules is shown in Figure~\ref{fig:figure3}.

The CA module employs both GAP and global max pooling (GMP), defined as follows:

\begin{equation}
    GAP\left( \mathbf{F} \right) = \frac{1}{HW}\sum_{h = 1}^{H}{\sum_{w = 1}^{W}\mathbf{F}_{c,h,w}} \in \mathbb{R}^{C \times 1 \times 1}
    \tag{5}
    \label{eq:5}
\end{equation}

\begin{equation}
    GMP\left( \mathbf{F} \right) = \max_{1 \leq h \leq H,1 \leq w \leq W}\mathbf{F}_{c,h,w} \in \mathbb{R}^{C \times 1 \times 1}
    \tag{6}
    \label{eq:6}
\end{equation}

For each channel \(c\), GAP averages all \((H \times W)\) spatial locations to produce a scalar. For each channel \(c\), the GMP selects the maximum activation over the same \((H \times W)\) spatial extent.

CA is computed as follows:

\begin{equation}
    \mathbf{M}_{CH}\left( \mathbf{F} \right) = \sigma\left( MLP\left( GAP\left( \mathbf{F} \right) \right) + MLP\left( GMP\left( \mathbf{F} \right) \right) \right) \in \mathbb{R}^{C \times 1 \times 1}
    \tag{7}
    \label{eq:7}
\end{equation}

\begin{equation}
    = \sigma\left( \mathbf{W}_{1}\left( ReLU\left( \mathbf{W}_{0}\left( \mathbf{F}_{avg}^{c} \right) \right) \right) + \mathbf{W}_{1}\left( ReLU\left( \mathbf{W}_{0}\left( \mathbf{F}_{\max}^{c} \right) \right) \right) \right)
    \tag{8}
    \label{eq:8}
\end{equation}

where \(\sigma\) denotes the sigmoid function and \(\mathbf{W}_{0} \in \mathbb{R}^{C/r \times C}\) and \(\mathbf{W}_{1} \in \mathbb{R}^{C \times C/r}\) represent the weight matrices of the shared MLP. The reduction ratio \(r\) controls the dimensionality of the hidden layer. Weights \(\mathbf{W}_{0}\) and \(\mathbf{W}_{1}\) were shared between the average- and max-pooled inputs. Rectified linear unit (ReLU) activation was applied after the transformation with \(\mathbf{W}_{0}\).

To generalize the aggregation operations used in SA and their generalization to other axes, we define axis-wise average pooling (AAP) and axis-wise max pooling (AMP). Pooling along the channel axis is expressed as follows:

\begin{equation}
    AAP\left( \mathbf{F};C \right) = \frac{1}{C}\sum_{c = 1}^{C}\mathbf{F}_{c,h,w} \in \mathbb{R}^{1 \times H \times W}
    \tag{9}
    \label{eq:9}
\end{equation}

\begin{equation}
    AMP\left( \mathbf{F};C \right) = \max_{1 \leq c \leq C}\mathbf{F}_{c,h,w} \in \mathbb{R}^{1 \times H \times W}
    \tag{10}
    \label{eq:10}
\end{equation}

These operations aggregate information across the channel dimension. Analogous pooling can be performed along the height or width axes, yielding average-pooled outputs \(AAP\left( \mathbf{F};H \right) \in \mathbb{R}^{C \times 1 \times W}\) and \(AAP\left( \mathbf{F};W \right) \in \mathbb{R}^{C \times H \times 1}\), respectively. Max-pooling is similarly generalized. The SA is computed as follows:

\begin{equation}
    \mathbf{M}_{SP}\left( \mathbf{F} \right) = \sigma\left( f^{7 \times 7}\left( \left\lbrack AAP\left( \mathbf{F};C \right)\| AMP\left( \mathbf{F};C \right) \right\rbrack \right) \right) \in \mathbb{R}^{1 \times H \times W}
    \tag{11}
    \label{eq:11}
\end{equation}

\begin{equation}
    = \sigma\left( f^{7 \times 7}\left( \left\lbrack \mathbf{F}_{avg}^{c}\|\mathbf{F}_{\max}^{c} \right\rbrack \right) \right)
    \tag{12}
    \label{eq:12}
\end{equation}

where \(f^{7 \times 7}\) denotes the convolution with \(7 \times 7\) kernel. \(\|\) denotes concatenation along the channel axis. The overall CBAM architecture is shown in Figure~\ref{fig:figure2}, whereas the CA and SA modules are shown in Figure~\ref{fig:figure3}.

\subsection{Monte Carlo dropout}

The MC dropout provides an approximation of the Bayesian interpretation of dropout, enabling the estimation of uncertainty associated with neural network weights \cite{Gal2016MCdrop}. During inference, dropout remains active, and multiple stochastic forward passes are performed for the same input. Under the variational interpretation of dropout, each stochastic forward pass corresponds to sampling effective network weights from an approximate variational distribution, thereby propagating weight uncertainty to the predictions \cite{Xiao_2022_Self}.

Let \(\mathcal{W}\) denote the neural network weights and \(\mathcal{D =}(X,Y)\) represent the training dataset. In Bayesian inference, the predictive distribution for a new input \(x^{*}\) is obtained from the posterior predictive distribution.

\begin{equation}
	p(y^{*} \mid x^{*}, \mathcal{D})
	= \int
	p(y^{*} \mid x^{*}, \mathcal{W})\,
	p(\mathcal{W} \mid \mathcal{D})\,
	\mathrm{d}\mathcal{W}
	\tag{13}
	\label{eq:13}
\end{equation}

Because the exact posterior \(p(\mathcal{W} \mid \mathcal{D})\) is generally intractable, it is approximated by MC dropout using a variational distribution \(q_{\theta}(\mathcal{W})\) induced by dropout masks. The variational objective corresponds to the negative evidence lower bound.

\begin{equation}
	\mathcal{L}_{VI}(\theta)
	= -\mathbb{E}_{q_{\theta}(\mathcal{W})}
	\left[ \log p(Y \mid X, \mathcal{W}) \right]
	+ KL\left(q_{\theta}(\mathcal{W}) \,\|\, p(\mathcal{W})\right)
	\tag{14}
	\label{eq:14}
\end{equation}

For the regression task considered in this study, we assumed the following Gaussian observation model:

\begin{equation}
	p(y_{n} \mid x_{n}, \mathcal{W})
	= \mathcal{N}\left(y_{n}; f_{\mathcal{W}}(x_{n}), \sigma^{2}I\right)
	\tag{15}
	\label{eq:15}
\end{equation}

When the observation variance is fixed, the negative log-likelihood denotes the scaled squared error loss plus an additive constant.

\begin{equation}
	-\log p(y_{n} \mid x_{n}, \mathcal{W})
	= \frac{d}{2}\log(2\pi\sigma^{2})
	+ \frac{1}{2\sigma^{2}}
	\left\|y_{n} - f_{\mathcal{W}}(x_{n})\right\|_{2}^{2}
	\tag{16}
	\label{eq:16}
\end{equation}

Approximating the Kullback--Leibler (KL) divergence term with weight decay yields the following practical dropout training objective:

\begin{equation}
    \mathcal{L}(\theta) = \frac{1}{N}\sum_{n = 1}^{N}{\mathcal{l}_{reg}\left( y_{n},{\widehat{y}}_{n} \right)} + \lambda\sum_{l = 1}^{L}\left\| \theta_{l} \right\|_{2}^{2}
    \tag{17}
    \label{eq:17}
\end{equation}

where \(\mathcal{l}_{reg}\) denotes regression loss. For fixed \(\sigma^{2}\), the Gaussian negative log-likelihood is proportional to the squared-error loss up to an additive constant. In Equation~\eqref{eq:17}, \(N\) denotes the number of training examples, \(L\) represents the number of parameterized layers, and \(\lambda\) controls weight decay.

\section{Methodology}
\label{sec:method}

The proposed methodology comprises segment-wise CWT preprocessing, shared FAAC feature extraction, ordered segment fusion, GRU-based temporal encoding, and an RUL regression head. For each observation record, \(C = 5\) contiguous segments are extracted to generate segment-wise magnitude scalograms \cite{Deng_2024_Hybrid}. The shared FAAC parameters were applied to each segment at each observation time. The five-segment embeddings were concatenated in their original order to form a record-level latent representation. Representations of the latest \(L\) observations form the input sequence for the GRU.

The notations employed in this section are listed in Table~\ref{tab:1}.

\begingroup
\small
\setlength{\tabcolsep}{3pt}
\renewcommand{\arraystretch}{1.2}
\begin{longtable}{@{}>{\centering\arraybackslash}p{0.18\textwidth}>{\raggedright\arraybackslash}p{0.78\textwidth}@{}}
    \caption{Notation utilized in the proposed method.}\label{tab:1}\\
    \toprule
    Notation & Description \\
    \midrule
    \endfirsthead
    \toprule
    Notation & Description \\
    \midrule
    \endhead
    \(n\) & Bearing index. \\
    \(t\) & Observation index. \\
    \(T_{n}\) & Failure-time index of bearing \(n\). \\
    \(P\) & Number of samples in a single raw vibration record. \\
    \(\mathbf{x}_{n}(t)\) & Raw vibration record of bearing \(n\) at observation \(t\), \(\mathbf{x}_{n}(t) \in \mathbb{R}^{P}\). \\
    \(C\) & Number of ordered contiguous segments within each vibration record; \(C = 5\). \\
    \(P_{c}\) & Number of samples in segment \(c\). \\
    \(\mathbf{x}_{n}^{c}(t)\) & The \(c\)-th contiguous segment of \(\mathbf{x}_{n}(t)\), \(\mathbf{x}_{n}^{c}(t) \in \mathbb{R}^{P_{c}}\). \\
    \(M\) & Number of CWT scales. \\
    \({\widetilde{\mathbf{S}}}_{n}^{c}(t)\) & Magnitude scalogram of segment \(c\) prior to bilinear interpolation, \({\widetilde{\mathbf{S}}}_{n}^{c}(t) \in \mathbb{R}^{M \times P_{c}}\). \\
    \(H\) & Frequency-axis size of a resized segment scalogram. \\
    \(W\) & Local-time-axis size of a resized segment scalogram. \\
    \(\mathbf{S}_{n}^{c}(t)\) & Resized single-channel scalogram of segment \(c\), \(\mathbf{S}_{n}^{c}(t) \in \mathbb{R}^{1 \times H \times W}\). \\
    \(\mathbf{S}_{n}(t)\) & Ordered stack of the \(C\) segment-wise scalograms at observation \(t\), \(\mathbf{S}_{n}(t) \in \mathbb{R}^{C \times 1 \times H \times W}\). \\
    \(L\) & Number of observation records included in a causal input window. \\
    \(\tau_{t,j}\) & Observation index assigned to the \(j\)-th position of the causal window ending at \(t\). \\
    \(\mathbf{X}_{n}(t)\) & Sequential scalogram input ending at observation \(t\), \(\mathbf{X}_{n}(t) \in \mathbb{R}^{L \times C \times 1 \times H \times W}\). \\
    \(\Phi_{\theta}\) & FAAC feature extractor shared across all observation periods and segments, \(\theta\) denotes the trainable parameters of the shared FAAC feature extractor. \\
    \(d\) & Dimension of a segment-level embedding. \\
    \(\mathbf{e}_{n}^{c}(t)\) & FAAC embedding of segment \(c\) at observation \(t\), \(\mathbf{e}_{n}^{c}(t) \in \mathbb{R}^{d}\). \\
    \(\mathbf{l}_{n}(t)\) & Ordered concatenation of \(C\) segment embeddings at observation \(t\), \(\mathbf{l}_{n}(t) \in \mathbb{R}^{Cd}\). \\
    \(\mathbf{Z}_{n}(t)\) & Sequence of \(L\) record-level representations supplied to the GRU, \(\mathbf{Z}_{n}(t) \in \mathbb{R}^{L \times Cd}\). \\
    \(d_{h}\) & Dimension of the GRU hidden state. \\
    \(r_{n}(t)\) & Ground-truth normalized RUL of bearing \(n\) at observation \(t\). \\
    \({\widehat{r}}_{n}(t)\) & Predicted normalized RUL of bearing \(n\) at observation \(t\). \\
    \({\widetilde{r}}_{n}(t)\) & Causally EWMA-smoothed RUL prediction of bearing \(n\) at observation \(t\). \\
    \bottomrule
\end{longtable}
\endgroup

\subsection{Data preprocessing: segment-wise input construction}

For bearing \(n\), the raw vibration record collected at observation \(t\) contained \(P\) samples and is denoted as \(\mathbf{x}_{n}(t) = \left\lbrack \left\lbrack x_{n}(t) \right\rbrack_{1},\ldots,\left\lbrack x_{n}(t) \right\rbrack_{P} \right\rbrack^{T} \in \mathbb{R}^{P}\), where \(n = 1,\ldots,N\) index bearings, \(t = 1,\ldots,T_{n}\) index observations, and \(T_{n}\) denotes the failure time index of bearing \(n\). The quantity \(\left\lbrack x_{n}(t) \right\rbrack_{p}\) denotes the \(p\)-th sample in the record.

To construct representations of local time--frequency behavior within a vibration record, \(\mathbf{x}_{n}(t)\) was divided into \(C = 5\) contiguous segments in the sample order \cite{Deng_2024_Hybrid}. The segment boundaries satisfy the following conditions:

\begin{equation}
    0 = p_{0} < p_{1} < \cdots < p_{C} = P,\ \ P_{c} = p_{c} - p_{c - 1},\ \ c = 1,\ldots,C
    \tag{18}
    \label{eq:18}
\end{equation}

where \(p_{c}\) denotes the final sample index of segment \(c\) and \(P_{c}\) represents the number of samples in that segment. Therefore, segment \(c\) is expressed as follows:

\begin{equation}
    \mathbf{x}_{n}^{c}(t) = \left\lbrack \left\lbrack x_{n}(t) \right\rbrack_{p_{c - 1} + 1},\ldots,\left\lbrack x_{n}(t) \right\rbrack_{p_{c}} \right\rbrack^{T} \in \mathbb{R}^{P_{c}}
    \tag{19}
    \label{eq:19}
\end{equation}

The CWT in Equation~\ref{eq:1} is applied independently to each segment \(\mathbf{x}_{n}^{c}(t)\). Let \(a_{m}\) denote scale \(m\) and \(b_{q}\) denote the translation position \(q\) within the segment. The magnitude scalogram is defined as follows:

\begin{equation}
    {\widetilde{\mathbf{S}}}_{n}^{c}(t)\lbrack m,q\rbrack = \left| W_{n}^{c}\left( t;a_{m},b_{q} \right) \right|,\ \ m = 1,\ldots,M,\ \ q = 1,\ldots,P_{c}
    \tag{20}
    \label{eq:20}
\end{equation}

Therefore, \({\widetilde{\mathbf{S}}}_{n}^{c}(t) \in \mathbb{R}^{M \times P_{c}}\), where \(M\) represents the number of CWT scales and \(q\) indexes local translation positions within the segment.

To obtain a unified spatial size regardless of the segment length \(P_{c}\), the magnitude scalogram \({\widetilde{\mathbf{S}}}_{n}^{c}(t)\) was resized to \(H \times W\) using bilinear interpolation:

\begin{equation}
    \mathbf{S}_{n}^{c}(t) = \mathcal{B}_{H \times W}\left( {\widetilde{\mathbf{S}}}_{n}^{c}(t) \right) \in \mathbb{R}^{1 \times H \times W}
    \tag{21}
    \label{eq:21}
\end{equation}

where \(\mathcal{B}_{H \times W}( \cdot )\) denotes bilinear interpolation. The dimensions \(H\) and \(W\) represent the frequency and within-segment local time axes of the resized scalogram, respectively, and a leading dimension of size 1 denotes a single input channel. Hence, \(\mathbf{S}_{n}^{c}(t)\) denotes the single-channel magnitude scalogram for bearing \(n\) at observation \(t\) and segment \(c\).

The \(C\) segment-wise scalograms from the same observation were stacked in their original order to form an observation-level tensor.

\begin{equation}
    \mathbf{S}_{n}(t) = {Stack}_{c = 1}^{C}\left( \mathbf{S}_{n}^{c}(t) \right) \in \mathbb{R}^{C \times 1 \times H \times W}
    \tag{22}
    \label{eq:22}
\end{equation}

The first axis of \(\mathbf{S}_{n}(t)\), with size \(C\), represents the ordered segment positions within the vibration record.

To predict the RUL at observation \(t\), a causal input window of length \(L\) was constructed, including the current observation. The observation index assigned to position \(j\) in this window is defined as follows:

\begin{equation}
    \tau_{t,j} = \max(1,t - L + j),\ \ j = 1,\ldots,L
    \tag{23}
    \label{eq:23}
\end{equation}

Given that \(\tau_{t,j} \leq t\) and \(\tau_{t,L} = t\), the window contains only information available at or before the current observation.

The sequential scalogram input is then expressed as follows:

\begin{equation}
    \mathbf{X}_{n}(t) = {Stack}_{j = 1}^{L}\left( \mathbf{S}_{n}\left( \tau_{t,j} \right) \right) \in \mathbb{R}^{L \times C \times 1 \times H \times W}
    \tag{24}
    \label{eq:24}
\end{equation}

\begin{figure}[htbp]
	\centering
	\includegraphics[width=\linewidth]{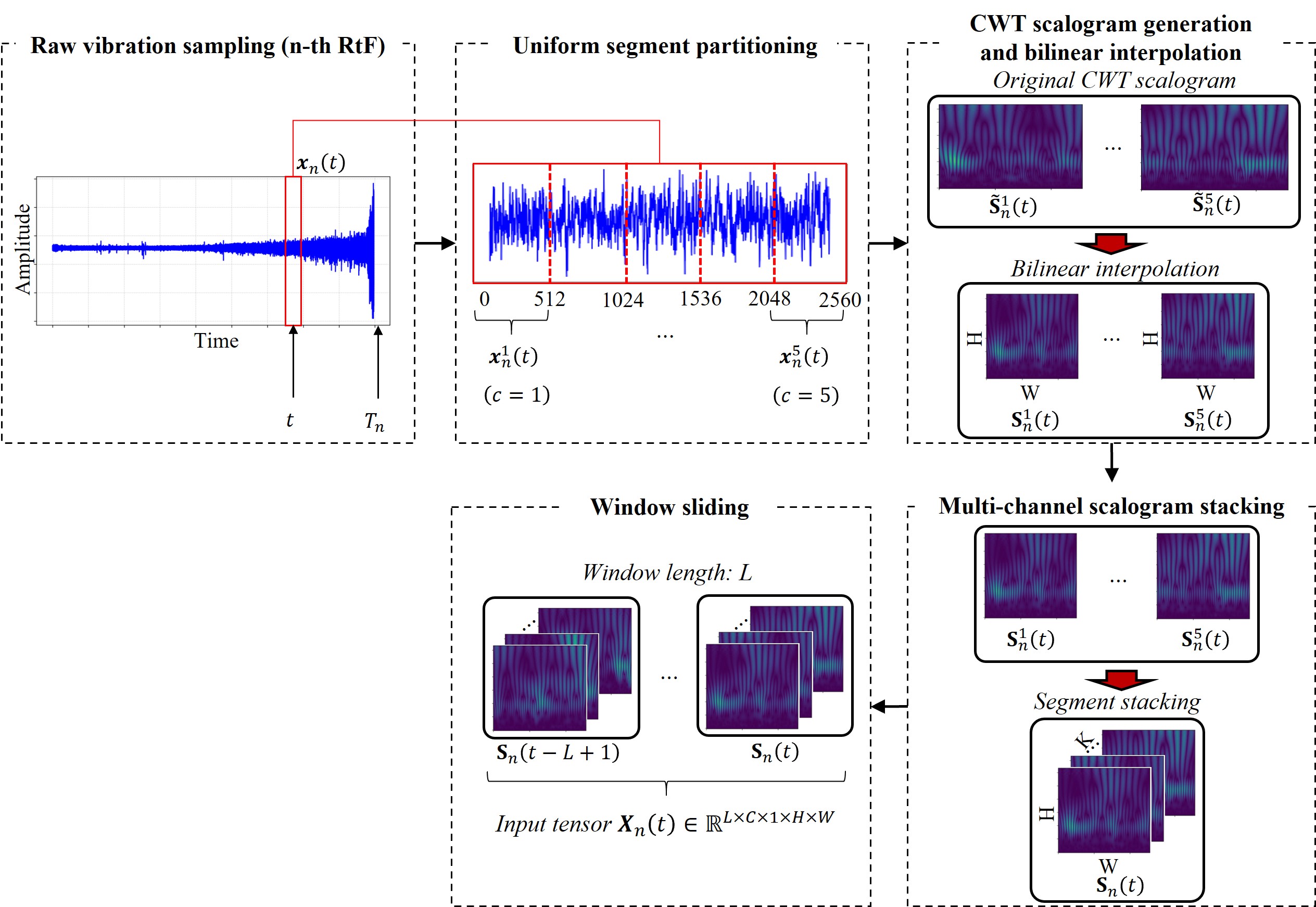}
	\caption{Segment-wise input construction. Each raw record is divided into five ordered contiguous segments, and each segment is independently transformed into a single-channel magnitude scalogram. The segment-wise scalograms from the latest \(L\) observations are stacked to construct \(\mathbf{X}_{n}(t)\).}
	\label{fig:figure4}
\end{figure}

Its elements satisfy \(\mathbf{X}_{n}(t)\lbrack j,c,:,:,:\rbrack = \mathbf{S}_{n}^{c}\left( \tau_{t,j} \right)\). The input tensor \(\mathbf{X}_{n}(t)\) lies in \(\mathbb{R}^{L \times C \times 1 \times H \times W}\); its five axes represent the causal window position, within-record segment position, input channel, frequency, and within-segment local time. When \(t < L\), the first observation is repeated at the beginning of the window to maintain an input length of \(L\). The preprocessing procedure is shown in Figure~\ref{fig:figure4}.

For bearing \(n\) with failure time index \(T_{n}\), the ground-truth normalized RUL at observation \(t\) is defined as follows:

\begin{equation}
    r_{n}(t) = \frac{T_{n} - t}{T_{n} - 1} \in \lbrack 0,1\rbrack,\ \ t = 1,\ldots,T_{n}
    \tag{25}
    \label{eq:25}
\end{equation}

This definition requires at least two observations per bearing and assigns a normalized RUL of 1 to the initial observation and 0 to the failure observation.

\subsection{Shared FAAC feature extractor}

The shared FAAC feature extractor, denoted as \(\Phi_{\theta}\) maps each segment-level scalogram to a \(d\)-dimensional embedding. This extractor comprises the MSAC, DCBAM, and DAP. MSAC extracts time--frequency features using anisotropic kernels with different receptive field sizes. Subsequently, DCBAM sequentially refines the channel and spatial responses along the frequency and local time axes. Finally, DAP aggregates the resulting feature map through both frequency-to-time and time-to-frequency pathways.

\begin{figure}[htbp]
	\centering
	\includegraphics[width=\linewidth]{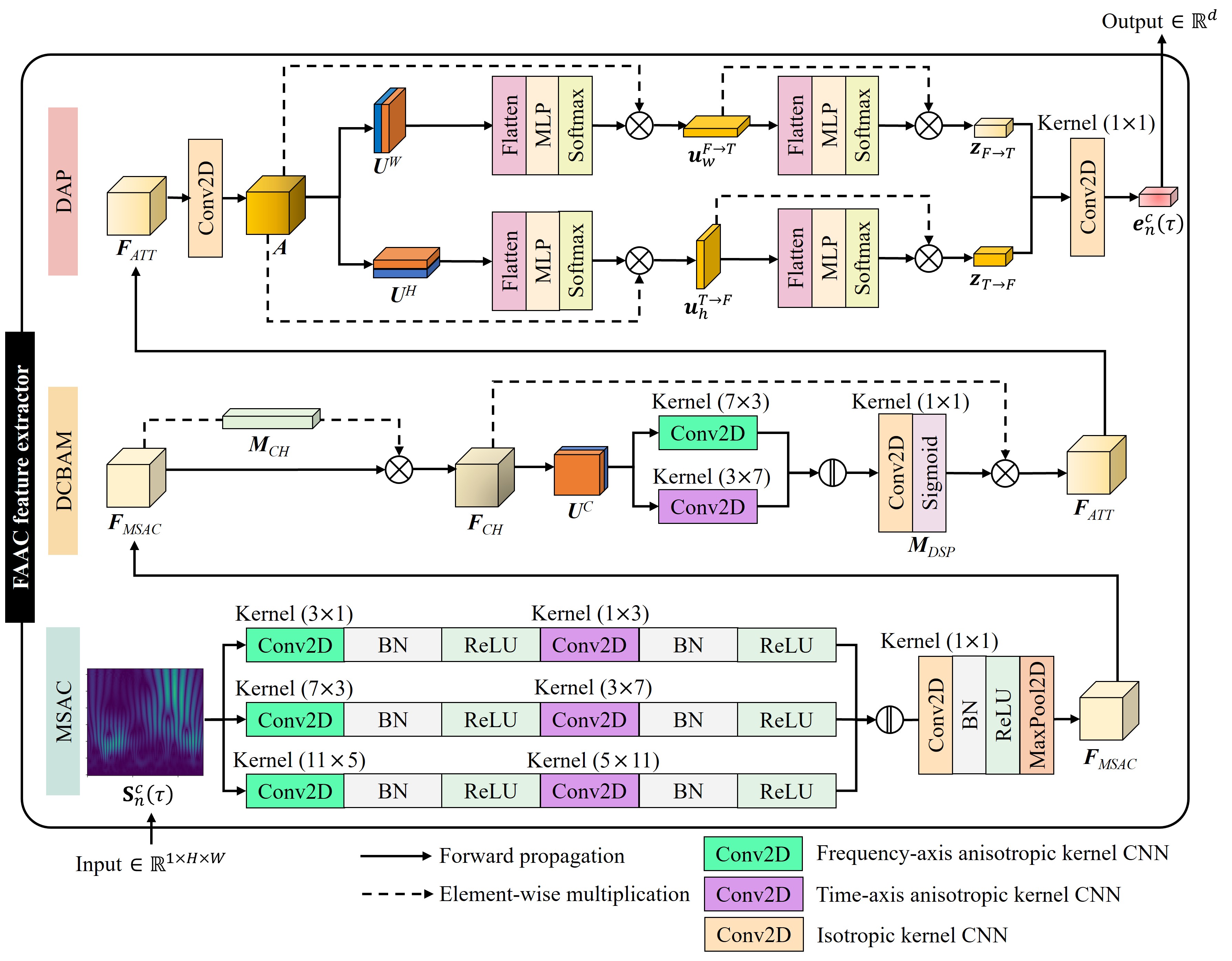}
	\caption{Architecture of the FAAC feature extractor, comprising multi-scale anisotropic convolution, dual-axis convolutional block attention module, and dynamic adaptive pooling.}
	\label{fig:figure5}
\end{figure}

At observation \(\tau\), the FAAC embedding for bearing \(n\) and segment \(c\) is defined as follows:

\begin{equation}
    \mathbf{e}_{n}^{c}(\tau) = \Phi_{\theta}\left( \mathbf{S}_{n}^{c}(\tau) \right) \in \mathbb{R}^{d}
    \tag{26}
    \label{eq:26}
\end{equation}

where \(\theta\) denotes the trainable parameters of FAAC, which are shared across observation periods and segments. The architecture of the FAAC feature extractor is shown in Figure~\ref{fig:figure5}.

\subsubsection{Multi-scale anisotropic convolution}

The input scalogram for a single segment is denoted as follows:

\begin{equation}
    \mathbf{F}^{(0)} = \mathbf{S}_{n}^{c}(\tau) \in \mathbb{R}^{1 \times H \times W}
    \tag{27}
    \label{eq:27}
\end{equation}

For kernel \(\mathbf{K} = \left( k_{F},k_{T} \right)\), \(k_{F}\) and \(k_{T}\) denote the kernel sizes along the frequency and local time axes, respectively. A convolutional unit with \(d_{o}\) output channels is defined as follows:

\begin{equation}
    \Gamma_{\mathbf{K}}^{\left( d_{o} \right)}\left( \mathbf{F} \right) = ReLU\left( BN\left( {Conv}_{\mathbf{K}}^{\left( d_{o} \right)}\left( \mathbf{F} \right) \right) \right)
    \tag{28}
    \label{eq:28}
\end{equation}

All the convolutions employed the same padding to preserve the spatial dimensions, with both stride and dilation set to one.

The MSAC comprises three parallel branches with different receptive field sizes. Within each branch, an anisotropic kernel elongated along the frequency axis is applied, followed by another anisotropic kernel elongated along the local time axis. The kernel pairs are expressed as follows:

\begin{equation}
    \begin{gathered}
            \left( \mathbf{K}_{1}^{(1)},\mathbf{K}_{1}^{(2)} \right) = \left( (3,1),(1,3) \right) \\
            \left( \mathbf{K}_{2}^{(1)},\mathbf{K}_{2}^{(2)} \right) = \left( (7,3),(3,7) \right) \\
            \left( \mathbf{K}_{3}^{(1)},\mathbf{K}_{3}^{(2)} \right) = \left( (11,5),(5,11) \right)
    \end{gathered}
    \tag{29}
    \label{eq:29}
\end{equation}

Let \(D_{s}\) denote the number of output channels in branch \(s\). The output of each branch is expressed as:

\begin{equation}
    \mathbf{Y}_{S} = \Gamma_{\mathbf{K}_{s}^{(2)}}^{\left( D_{s} \right)}\left( \Gamma_{\mathbf{K}_{s}^{(1)}}^{\left( D_{s} \right)}\left( \mathbf{F}^{(0)} \right) \right) \in \mathbb{R}^{D_{s} \times H \times W},\ \ s = 1,2,3
    \tag{30}
    \label{eq:30}
\end{equation}

The outputs of the three branches are concatenated along the channel axis:

\begin{equation}
    \mathbf{F}_{CAT} = \mathbf{Y}_{1}\|\mathbf{Y}_{2}\|\mathbf{Y}_{3} \in \mathbb{R}^{\widetilde{D} \times H \times W},\ \ \widetilde{D} = \sum_{s = 1}^{3}D_{s}
    \tag{31}
    \label{eq:31}
\end{equation}

The concatenated features are fused using a \(1 \times 1\) convolution with \(D\) output channels as follows:

\begin{equation}
    \overline{\mathbf{F}} = \Gamma_{(1,1)}^{(D)}\left( \mathbf{F}_{cat} \right)
    \tag{32}
    \label{eq:32}
\end{equation}

The spatial resolution is then reduced by \(2 \times 2\) max pooling:

\begin{equation}
    \mathbf{F}_{MSAC} = {Maxpool}_{2 \times 2}\left( \overline{\mathbf{F}} \right) \in \mathbb{R}^{D \times H^{'} \times W^{'}},\ \ H^{'} = \left\lfloor \frac{H}{2} \right\rfloor,\ \ W^{'} = \left\lfloor \frac{W}{2} \right\rfloor
    \tag{33}
    \label{eq:33}
\end{equation}

Here, \({Maxpool}_{2 \times 2}\) employs a stride of two.

\textbf{Dual-axis convolutional block attention module}

DCBAM retains the sequential channel--spatial attention structure of CBAM; however, it replaces the single spatial convolution with parallel anisotropic convolutions elongated along the frequency and local time axes. This architecture combines the directional spatial responses along the two axes of the scalogram.

First, the channel attention function \(\mathbf{M}_{CH}( \cdot )\) defined in Equation~\eqref{eq:7} is applied to obtain a channel-refined feature map:

\begin{equation}
    \mathbf{F}_{CH} = \mathbf{M}_{CH}\left( \mathbf{F}_{MSAC} \right) \odot \mathbf{F}_{MSAC}
    \tag{34}
    \label{eq:34}
\end{equation}

The channel attention map is broadcast along both spatial dimensions during element-wise multiplication.

Subsequently, average and max pooling were performed on \(\mathbf{F}_{CH}\) along the channel axis. Their outputs were concatenated along the axis to form a spatial descriptor.

\begin{equation}
    \mathbf{U}_{DSP} = AAP\left( \mathbf{F}_{CH};C \right)\| AMP\left( \mathbf{F}_{CH};C \right) \in \mathbb{R}^{2 \times H^{'} \times W^{'}}
    \tag{35}
    \label{eq:35}
\end{equation}

DCBAM applies parallel anisotropic convolutions to \(\mathbf{U}_{DSP}\) along the frequency and local time axes as follows:

\begin{equation}
    \begin{gathered}
            \mathbf{G}_{F} = {Conv}_{(7,3)}^{(1)}\left( \mathbf{U}_{DSP} \right),\ \ \mathbf{G}_{F} \in \mathbb{R}^{1 \times H^{'} \times W^{'}} \\
            \mathbf{G}_{T} = {Conv}_{(3,7)}^{(1)}\left( \mathbf{U}_{DSP} \right),\ \ \mathbf{G}_{T} \in \mathbb{R}^{1 \times H^{'} \times W^{'}}
    \end{gathered}
    \tag{36}
    \label{eq:36}
\end{equation}

where \(\mathbf{G}_{F}\) and \(\mathbf{G}_{T}\) denote the spatial responses obtained using the receptive fields elongated along the frequency and local time axes, respectively. Both convolutions employ the same padding.

The two-directional responses were concatenated along the channel axis and fused using a \(1 \times 1\) convolution, followed by sigmoid activation, to generate the spatial attention map:

\begin{equation}
    \mathbf{M}_{DSP} = \sigma\left\lbrack {Conv}_{(1,1)}^{(1)}\left( \mathbf{G}_{T}\|\mathbf{G}_{F} \right) \right\rbrack \in \mathbb{R}^{1 \times H^{'} \times W^{'}}
    \tag{37}
    \label{eq:37}
\end{equation}

Finally, the spatial attention map was applied to the channel-refined feature map:

\begin{equation}
    \mathbf{F}_{ATT} = \mathbf{M}_{DSP} \odot \mathbf{F}_{CH} \in \mathbb{R}^{D \times H^{'} \times W^{'}}
    \tag{38}
    \label{eq:38}
\end{equation}

\textbf{Dynamic adaptive pooling}

First, DAP applies a pointwise convolution to the attention-refined feature map:

\begin{equation}
    \mathbf{A} = {Conv}_{(1,1)}^{(D)}\left( \mathbf{F}_{att} \right)
    \tag{39}
    \label{eq:39}
\end{equation}

where \(\mathbf{A} \in \mathbb{R}^{D \times H^{'} \times W^{'}}\) is aggregated through frequency-to-time and time-to-frequency pathways. The indices \(h = 1,\ldots,H^{'}\) and \(w = 1,\ldots,W^{'}\) denote positions along the frequency and local-time axes, respectively, and \(\mathbf{A}_{:,h,w} \in \mathbb{R}^{D}\) denotes the channel feature vector at position \((h,w)\).

DAP aggregates the same feature map in two distinct orders: the frequency-to-time path first aggregates along the frequency axis, followed by the local time axis. The time-to-frequency path performs these operations in reverse order.

\textbf{Frequency-to-time path}

For each frequency position \(h\), the average and max descriptors are computed by pooling over the local time axis:

\begin{equation}
    {\overline{\mathbf{a}}}_{h}^{F} = AAP\left( \mathbf{A};W \right) \in \mathbb{R}^{D \times H^{'} \times 1}
    \tag{40}
    \label{eq:40}
\end{equation}

\begin{equation}
    {\widehat{\mathbf{a}}}_{h}^{F} = AMP\left( \mathbf{A};W \right) \in \mathbb{R}^{D \times H^{'} \times 1}
    \tag{41}
    \label{eq:41}
\end{equation}

These descriptors are then utilized to compute the frequency-axis attention weights \(\mathbf{\omega}_{h}^{F}\) as follows:

\begin{equation}
	\begin{gathered}
		\mathbf{U}^{W}
		= \overline{\mathbf{a}}_{h}^{F} \Vert \widehat{\mathbf{a}}_{h}^{F}
		\in \mathbb{R}^{2D \times H' \times 1} \\
		\widetilde{\mathbf{U}}^{W}
		= Flatten\left(\mathbf{U}^{W}\right)
		\in \mathbb{R}^{2DH'} \\
		\mathbf{s}^{F}
		= \mathbf{W}_{F}\widetilde{\mathbf{U}}^{W}
		+ \mathbf{b}_{F}
		\in \mathbb{R}^{H'} \\
		\omega_{h}^{F}
		= \frac{\exp\left(s_{h}^{F}\right)}
		{\sum_{h'=1}^{H'} \exp\left(s_{h'}^{F}\right)},
		\qquad h=1,\ldots,H'
	\end{gathered}
	\tag{42}
	\label{eq:42}
\end{equation}

where \(\mathbf{W}_{F} \in \mathbb{R}^{H^{'} \times 2DH^{'}}\) and \(\mathbf{b}_{F} \in \mathbb{R}^{H^{'}}\) are the trainable weight matrix and bias vector, respectively. The frequency-axis weights \(\mathbf{\omega}_{h}^{F}\) are used to aggregate the frequency axis at each local-time position \(\omega\), resulting in the vector \(\mathbf{u}_{\omega}^{F \rightarrow T}\):

\begin{equation}
    \mathbf{u}_{\omega}^{F \rightarrow T} = \sum_{h = 1}^{H^{'}}{\mathbf{\omega}_{h}^{F}\mathbf{A}_{:,h,:}} \in \mathbb{R}^{D \times W^{'}}
    \tag{43}
    \label{eq:43}
\end{equation}

Subsequently, the time-axis attention weights were computed by applying an affine transformation followed by a softmax function:

\begin{equation}
    \begin{gathered}
            {\widetilde{\mathbf{u}}}_{\omega}^{F \rightarrow T} = Flatten\left( \mathbf{u}_{\omega}^{F \rightarrow T} \right) \in \mathbb{R}^{DW^{'}} \\
            \mathbf{r}_{\omega}^{T|F} = \mathbf{W}_{T|F}{\widetilde{\mathbf{u}}}_{\omega}^{F \rightarrow T} + \mathbf{b}_{T|F} \in \mathbb{R}^{W^{'}} \\
            \mathbf{v}_{\omega}^{T|F} = \frac{\exp\left( \mathbf{r}_{\omega}^{T|F} \right)}{\sum_{\omega^{'} = 1}^{W^{'}}{\exp\left( \mathbf{r}_{\omega^{'}}^{T|F} \right)}} \in \mathbb{R}^{W^{'}}
    \end{gathered}
    \tag{44}
    \label{eq:44}
\end{equation}

where \(\mathbf{W}_{T|F} \in \mathbb{R}^{W^{'} \times DW^{'}}\) and \(\mathbf{b}_{T|F} \in \mathbb{R}^{W^{'}}\) denote a trainable weight matrix and bias vector, respectively.

Finally, the resulting time-axis attention weights \(\mathbf{v}_{\omega}^{T|F}\) were utilized to perform weighted aggregation over the remaining time axis:

\begin{equation}
    \mathbf{z}_{F \rightarrow T} = \sum_{\omega = 1}^{W^{'}}{\mathbf{v}_{\omega}^{T|F}\mathbf{u}_{\omega}^{F \rightarrow T}} \in \mathbb{R}^{D}
    \tag{45}
    \label{eq:45}
\end{equation}

The resulting representation \(\mathbf{z}_{F \rightarrow T}\) denotes a \(D\)-channel feature vector obtained by adaptively aggregating features along the frequency, followed by the time axis, using the corresponding attention weights.

\textbf{Time-to-frequency path}

The time-to-frequency path utilizes a similar aggregation principle as the FT path, with the order of the two axes reversed as follows:

\begin{equation}
    \begin{gathered}
            {\overline{\mathbf{a}}}_{w}^{T} = AAP\left( \mathbf{A};H \right) \in \mathbb{R}^{D \times 1 \times W^{'}} \\
            {\widehat{\mathbf{a}}}_{w}^{T} = AMP\left( \mathbf{A};H \right) \in \mathbb{R}^{D \times 1 \times W^{'}} \\
            \mathbf{U}^{H} = {\overline{\mathbf{a}}}_{w}^{T}\|{\widehat{\mathbf{a}}}_{w}^{T} \in \mathbb{R}^{2D \times 1 \times W^{'}} \\
            {\widetilde{\mathbf{U}}}^{H} = Flatten\left( \mathbf{U}^{H} \right) \in \mathbb{R}^{2DW^{'}} \\
            \mathbf{s}_{w}^{T} = \mathbf{W}_{T}{\widetilde{\mathbf{U}}}^{H} + \mathbf{b}_{T} \in \mathbb{R}^{W^{'}} \\
            \mathbf{\omega}_{w}^{T} = \frac{\exp\left( \mathbf{s}_{w}^{T} \right)}{\sum_{w^{'} = 1}^{W^{'}}{\exp\left( \mathbf{s}_{w^{'}}^{T} \right)}} \in \mathbb{R}^{W^{'}} \\
            \mathbf{u}_{h}^{T \rightarrow F} = \sum_{w = 1}^{W^{'}}{\mathbf{\omega}_{w}^{T}\mathbf{A}_{:,:,w}} \in \mathbb{R}^{D \times H^{'}} \\
            {\widetilde{\mathbf{u}}}_{h}^{T \rightarrow F} = Flatten\left( \mathbf{u}_{h}^{T \rightarrow F} \right) \in \mathbb{R}^{DH^{'}} \\
            \mathbf{r}_{h}^{F|T} = \mathbf{W}_{F|T}{\widetilde{\mathbf{u}}}_{h}^{T \rightarrow F} + \mathbf{b}_{F|T} \in \mathbb{R}^{H^{'}} \\
            \mathbf{v}_{h}^{F|T} = \frac{\exp\left( \mathbf{r}_{h}^{F|T} \right)}{\sum_{h^{'} = 1}^{H^{'}}{\exp\left( \mathbf{r}_{h^{'}}^{F|T} \right)}} \in \mathbb{R}^{H^{'}} \\
            \mathbf{z}_{T \rightarrow F} = \sum_{h = 1}^{H^{'}}{\mathbf{v}_{h}^{F|T}\mathbf{u}_{h}^{T \rightarrow F}} \in \mathbb{R}^{D}
    \end{gathered}
    \tag{46}
    \label{eq:46}
\end{equation}

The \(D\)-dimensional aggregated vectors \(\mathbf{z}_{F \rightarrow T}\) and \(\mathbf{z}_{T \rightarrow F}\) from the two paths are concatenated and projected to a \(d\)-dimensional latent representation, as follows:

\begin{equation}
    \mathbf{e}_{n}^{c}(\tau) = \mathbf{W}_{f}\left( \mathbf{z}_{F \rightarrow T}\|\mathbf{z}_{T \rightarrow F} \right) + \mathbf{b}_{f} \in \mathbb{R}^{d}
    \tag{47}
    \label{eq:47}
\end{equation}

The dimensions of the projection parameters are denoted as \(\mathbf{W}_{f} \in \mathbb{R}^{d \times 2D}\) and \(\mathbf{b}_{f} \in \mathbb{R}^{d}\).

\subsection{Ordered segment fusion and sequence construction}

The feature extractor \(\Phi_{\theta}\) in Equation~\eqref{eq:26} is shared across all segments. For each observation \(\tau\), the \(C\) segment embeddings are concatenated in their original segment order to form a unified latent representation:

\begin{equation}
    \mathbf{l}_{n}(\tau) = \mathbf{e}_{n}^{1}(\tau)\|\cdots\|\mathbf{e}_{n}^{C}(\tau) \in \mathbb{R}^{Cd}
    \tag{48}
    \label{eq:48}
\end{equation}

where \(\mathbf{l}_{n}(\tau)\) represents the entire vibration record acquired at observation index \(\tau\).

At the current observation index \(t\), a full input window comprises the most recent \(L\) observations, with indices \(\tau_{t,1} < \tau_{t,2} < \cdots < \tau_{t,L} = t\). For consecutive observation indices, \(\tau_{t,j} = t - L + j\), where \(j = 1,\ldots,L\). If fewer observations are available, left-padding is applied according to Equation~\eqref{eq:23}. The latent representations were arranged in acquisition order to construct the GRU input matrix:

\begin{equation}
    \left\lbrack \mathbf{Z}_{n}(t) \right\rbrack_{j,:} = {\mathbf{l}_{n}\left( \tau_{t,j} \right)}^{T},\ \ j = 1,\ldots,L,\ \ \mathbf{Z}_{n}(t) \in \mathbb{R}^{L \times Cd}
    \tag{49}
    \label{eq:49}
\end{equation}

\subsection{GRU-based temporal modeling and RUL regression}

The unidirectional GRU has hidden dimension \(d_{h}\). The initial hidden state is set to the zero vector for each input window, and the sequence is processed as follows:

\begin{equation}
    \begin{gathered}
            \mathbf{h}_{n,0}(t) = \mathbf{0} \in \mathbb{R}^{d_{h}} \\
            \mathbf{h}_{n,j}(t) = GRUCell\left( \left\lbrack {\left\lbrack \mathbf{Z}_{n}(t) \right\rbrack_{j,:}}^{T},\mathbf{h}_{n,j - 1}(t) \right\rbrack \right),\ \ j = 1,\ldots,L \\
            {\widetilde{\mathbf{h}}}_{n,j}(t) = LayerNorm\left( \mathbf{h}_{n,j}(t) \right)
    \end{gathered}
    \tag{50}
    \label{eq:50}
\end{equation}

Layer normalization is independently applied at each GRU output position. Only the normalized hidden state at the final position, \({\widetilde{\mathbf{h}}}_{n,L}(t) \in \mathbb{R}^{d_{h}}\), is forwarded to the RUL regression head.

The RUL regression head comprises fully connected transformations, ReLU activation, dropout, and sigmoid outputs.

\begin{equation}
    {\overline{\mathbf{h}}}_{n}(t) = Dropout\left( ReLU\left( \mathbf{W}_{a}{\widetilde{\mathbf{h}}}_{n,L}(t) + \mathbf{b}_{a} \right) \right)
    \tag{51}
    \label{eq:51}
\end{equation}

\begin{equation}
    {\widehat{r}}_{n}(t) = \sigma\left( \mathbf{W}_{r}{\overline{\mathbf{h}}}_{n}(t) + \mathbf{b}_{r} \right) \in (0,1)
    \tag{52}
    \label{eq:52}
\end{equation}

Let \(\mathcal{B}_{n}\) denote a mini-batch sampled from the \(n\)-th bearing. Each sample in \(\mathcal{B}_{n}\) denotes an input--target pair \(\left( \mathbf{X}_{n}(t),r_{n}(t) \right)\). The mean squared error (MSE) loss for \(\mathcal{B}_{n}\) is computed as follows:

\begin{equation}
    \mathcal{L}_{MSE}\left( \theta;\mathcal{B}_{n} \right) = \frac{1}{\left| \mathcal{B}_{n} \right|}\sum_{\left( \mathbf{X}_{n}(t),r_{n}(t) \right) \in \mathcal{B}_{n}}^{}\left\lbrack {\widehat{r}}_{n}(t;\theta) - r_{n}(t) \right\rbrack^{2}
    \tag{53}
    \label{eq:53}
\end{equation}

where \(\theta\) denotes all trainable model parameters, \({\widehat{r}}_{n}(t;\theta)\) denotes the RUL predicted from \(\mathbf{X}_{n}(t)\), and \(\left| \mathcal{B}_{n} \right|\) represents the number of samples in the mini-batch.

\begin{figure}[htbp]
	\centering
	\includegraphics[height=0.8\linewidth]{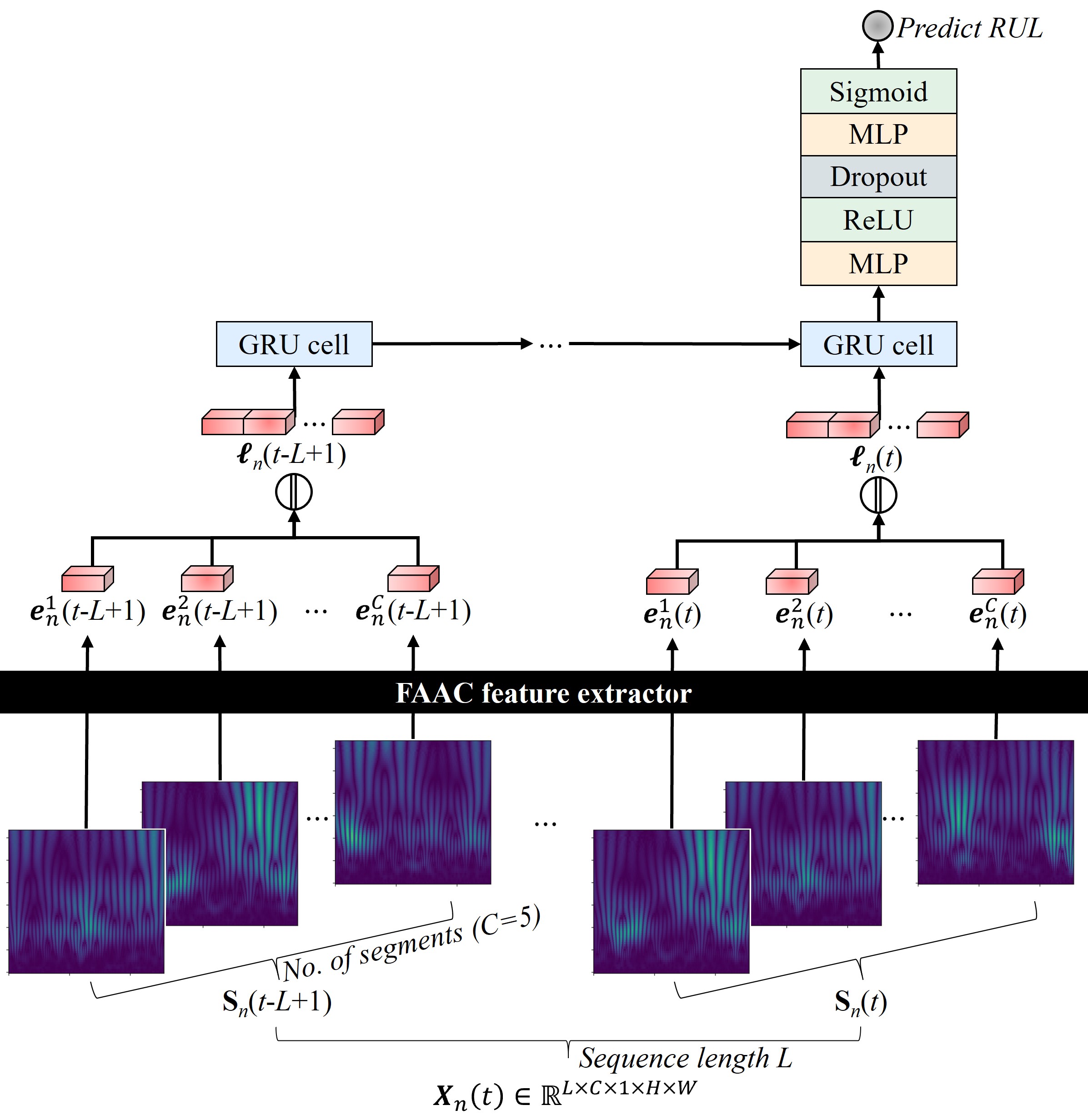}
	\caption{Proposed FAAC-GRU architecture.}
	\label{fig:figure6}
\end{figure}

The FAAC-GRU architecture for the segment-level input is shown in Figure~\ref{fig:figure6}.

\subsection{RUL smoothing and MC-dropout inference}

To mitigate short-term fluctuations in RUL predictions, an exponentially weighted moving average (EWMA) is applied to the output, denoted as \({\widehat{r}}_{n}(t;\theta)\) \cite{Jiang_2023_Dual}. Let \(t_{0}\) denote the initial evaluation time and \(\beta \in (0,1\rbrack\) denote the weight assigned to the current prediction. The EWMA is calculated as follows:

\begin{equation}
    {\widetilde{r}}_{n}\left( t_{0};\theta \right) = {\widehat{r}}_{n}\left( t_{0};\theta \right),\ \ {\widetilde{r}}_{n}(t;\theta) = \beta{\widehat{r}}_{n}(t;\theta) + (1 - \beta){\widetilde{r}}_{n}(t - 1;\theta),\ \ t > t_{0}
    \tag{54}
    \label{eq:54}
\end{equation}

where \(\beta\) controls the weight assigned to the current prediction relative to the previous smoothed value. A smaller \(\beta\) yields stronger smoothing and slower adaptation to changes, whereas a larger \(\beta\) results in faster responsiveness but increased sensitivity to short-term fluctuations. Therefore, the choice of \(\beta\) balances smoothing and responsiveness. In accordance with prior studies, we set \(\beta = 0.2\) \cite{Jiang_2023_Dual,Ye_2023_Selective}.

During MC-dropout inference, the trained FAAC-GRU parameters \(\theta\) remain fixed, and dropout is enabled exclusively in the regression head. Across \(N_{MC}\) stochastic forward passes, the raw prediction from pass \(q\) is expressed as follows:

\begin{equation}
    {\widehat{r}}_{n}^{(q)}\left( t_{n,k};\theta \right) = f_{\theta,\xi_{n,k}^{(q)}}\left( \mathbf{X}_{n}\left( t_{n,k} \right) \right),\ \ q = 1,\ldots,N_{MC},\ \ k = 1,\ldots,K_{n}
    \tag{55}
    \label{eq:55}
\end{equation}

where \(\xi_{n,k}^{(q)}\) denotes the dropout mask realization utilized in stochastic forward pass \(q\) at evaluation time \(t_{n,k}\).

For stochastic forward pass \(q\), the complete sequence of raw predictions is denoted as \({\widehat{\mathbf{r}}}_{n}^{(q)} = \left\lbrack {\widehat{r}}_{n}^{(q)}\left( t_{n,1};\theta \right),\ldots,{\widehat{r}}_{n}^{(q)}\left( t_{n,K_{n}};\theta \right) \right\rbrack^{T}\). Let \(\psi_{\beta}\) denote the EWMA operator in Equation~\eqref{eq:54}. Each stochastic prediction sequence is independently smoothed, yielding \({\widetilde{\mathbf{r}}}_{n}^{(q)} = \psi_{\beta}\left( {\widehat{\mathbf{r}}}_{n}^{(q)} \right)\) for \(q = 1,\ldots,N_{MC}\).

For \(\delta \in (0,1)\), the pointwise dropout uncertainty interval at time \(t_{n,k}\) is calculated from the smoothed samples as follows:

\begin{equation}
    \mathcal{I}_{n,1 - \delta}\left( t_{n,k} \right) = \left\lbrack Q_{\frac{\delta}{2}}\left( \left\{ {\widetilde{\mathbf{r}}}_{n}^{(q)}\left( t_{n,k} \right) \right\}_{q = 1}^{N_{MC}} \right),Q_{1 - \frac{\delta}{2}}\left( \left\{ {\widetilde{\mathbf{r}}}_{n}^{(q)}\left( t_{n,k} \right) \right\}_{q = 1}^{N_{MC}} \right) \right\rbrack
    \tag{56}
    \label{eq:56}
\end{equation}

where \(Q_{\frac{\delta}{2}}( \cdot )\) denotes the empirical \(\frac{\delta}{2}\)-quantile of the MC samples. For example, \(\delta = 0.05\) yields an interval bounded by the 2.5\textsuperscript{th} and 97.5\textsuperscript{th} percentiles.

\section{Experiments}
\label{sec:experiments}

\subsection{Data description}

\begin{figure}[htbp]
	\centering
	\includegraphics[width=\linewidth]{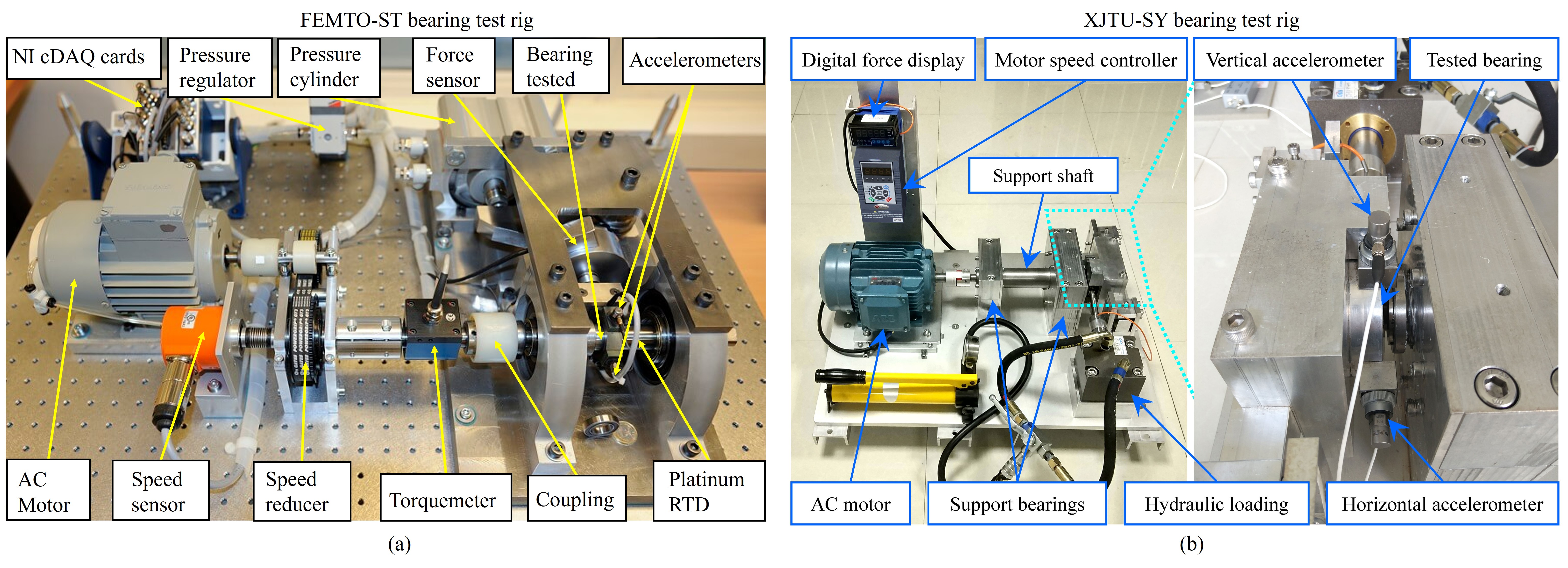}
	\caption{Bearing test rigs for the two public datasets. (a) FEMTO-ST bearing dataset, released for the 2012 PHM challenge, (b) XJTU-SY bearing dataset, provided by xi'an Jiaotong University.}
	\label{fig:figure7}
\end{figure}

The RUL prediction performance of the proposed framework was evaluated using two public bearing datasets. The two test rigs are shown in Figure~\ref{fig:figure7}, and the experimental conditions are described below.

\textbf{FEMTO-ST bearing dataset}

The FEMTO-ST bearing (FB) dataset, released for the 2012 PHM challenge, was collected using the PRONOSTIA test rig to facilitate research in rolling bearing fault diagnosis and prognostics \cite{Nectoux_2012_PRONOSTIA}. The dataset comprised 17 run-to-failure (RtF) experiments conducted under three operating conditions. The vibrations were acquired using accelerometers mounted in the horizontal and vertical directions. The rotational speeds and applied loads were as follows: Condition 1, 1800 rpm and 4000 N; Condition 2, 1650 rpm and 4200 N; and Condition 3, 1500 rpm and 5000 N. Vibration signals were sampled at 25.6 kHz for a duration of 0.1 s every 10 s.

\textbf{XJTU-SY bearing dataset}

The XJTU-SY bearing (XB) dataset was provided by the Institute of Design Science and Basic Component at Xi' an Jiaotong University (XJTU), Shaanxi, China, and Changing Sumyoung Technology Co., Ltd. (SY), Zhejiang, China \cite{Wang_2020_Hybrid}. This dataset contains complete RtF records for 15 rolling element bearings, obtained through accelerated degradation experiments under three operating conditions: Condition 1, 2100 rpm and 12,000 N; Condition 2, 2250 rpm and 11,000 N; and Condition 3, 2400 rpm and 10,000 N. Signals were sampled at 25.6 kHz for 1.28 s per record, yielding 32,768 samples per record, with a recording interval of 1 min.

\subsection{Evaluation metric}

The performance was evaluated using the mean absolute error (MAE), root mean squared error (RMSE), and score for RUL trajectories normalized to [0, 1] \cite{Jiang_2023_Dual,Deng_2024_Hybrid}. Let \(\Omega_{n}\) denote the test bearing set of evaluation time points. MAE and RMSE were calculated as follows:

\begin{equation}
    {MAE}_{n} = \frac{1}{\left| \Omega_{n} \right|}\sum_{t \in \Omega_{n}}^{}\left| {\widehat{r}}_{n}(t;\theta) - r_{n}(t;\theta) \right|
    \tag{57}
    \label{eq:57}
\end{equation}

\begin{equation}
    {RMSE}_{n} = \sqrt{\frac{1}{\left| \Omega_{n} \right|}\sum_{t \in \Omega_{n}}^{}\left( {\widehat{r}}_{n}(t;\theta) - r_{n}(t;\theta) \right)^{2}}
    \tag{58}
    \label{eq:58}
\end{equation}

For a constant error \(\left| {\widehat{r}}_{n}(t;\theta) - r_{n}(t;\theta) \right|\), when the predicted RUL is higher than the actual value, the score will be higher than when the predicted value is lower than the actual value. This asymmetric penalty function provides a more realistic evaluation of prognostic performance. The Score metric is defined as follows \cite{Zuo_2023_Hybrid}:

\begin{equation}
    {Score}_{n} = \left\{ \begin{array}{r}
            \sum_{t \in \Omega_{n}}^{}\left( e^{- \frac{{\widehat{r}}_{n}(t) - r_{n}(t)}{13}} - 1 \right),\ \ {\widehat{r}}_{n}(t) < r_{n}(t) \\
            \sum_{t \in \Omega_{n}}^{}\left( e^{\frac{{\widehat{r}}_{n}(t) - r_{n}(t)}{10}} - 1 \right),\ \ {\widehat{r}}_{n}(t) \geq r_{n}(t)
        \end{array} \right.\ 
    \tag{59}
    \label{eq:59}
\end{equation}

For the same absolute prediction error, the Score assigns a smaller penalty when the predicted failure time is earlier than the actual failure time. A larger penalty is assigned when the predicted failure time is later than the actual failure time, meaning that the model overestimates how long the bearing can continue to operate. Lower Score values indicate better performance.

\subsection{Experimental settings}

This section outlines the training and test splits, as well as the baseline models used for comparison.

\begin{table}[htbp]
    \centering
    \caption{Training and test splits for the FB dataset, following the PHM 2012 challenge protocol \cite{Nectoux_2012_PRONOSTIA}.}
    \label{tab:2}
    \small
    \setlength{\tabcolsep}{5pt}
    \renewcommand{\arraystretch}{1.2}
    \renewcommand{\multirowsetup}{\centering}
    \begin{tabularx}{\linewidth}{@{}*{4}{>{\centering\arraybackslash}X}@{}}
        \toprule
        Data set & Condition 1 & Condition 2 & Condition 3 \\
        \midrule
        \multirow[t]{2}{=}{Train set} & Bearing 1-1 (FB1-1) & Bearing 2-1 (FB2-1) & Bearing 3-1 (FB3-1) \\
        & Bearing 1-2 (FB1-2) & Bearing 2-2 (FB2-2) & Bearing 3-2 (FB3-2) \\
        \midrule
        \multirow[t]{5}{=}{Test set} & Bearing 1-3 (FB1-3) & Bearing 2-3 (FB2-3) & Bearing 3-3 (FB3-3) \\
        & Bearing 1-4 (FB1-4) & Bearing 2-4 (FB2-4) &  \\
        & Bearing 1-5 (FB1-5) & Bearing 2-5 (FB2-5) &  \\
        & Bearing 1-6 (FB1-6) & Bearing 2-6 (FB2-6) &  \\
        & Bearing 1-7 (FB1-7) & Bearing 2-7 (FB2-7) &  \\
        \bottomrule
    \end{tabularx}
\end{table}

\begin{table}[htbp]
    \centering
    \caption{Evaluation scheme for the XB dataset, limited to bearings with outer-race faults under Condition 1 and 2 \cite{Wang_2020_Hybrid}.}
    \label{tab:3}
    \small
    \setlength{\tabcolsep}{5pt}
    \renewcommand{\arraystretch}{1.2}
    \renewcommand{\multirowsetup}{\centering}
    \begin{tabularx}{\linewidth}{@{}*{3}{>{\centering\arraybackslash}X}@{}}
        \toprule
        Data set & Condition 1 & Condition 2 \\
        \midrule
        \multirow[t]{4}{=}{Leave-one-out cross validation} & Bearing 1-1 (XB1-1) & Bearing 2-2 (XB2-2) \\
        & Bearing 1-2 (XB1-2) & Bearing 2-4 (XB2-4) \\
        & Bearing 1-3 (XB1-3) & Bearing 2-5 (XB2-5) \\
        & Bearing 1-5 (XB1-5) &  \\
        \bottomrule
    \end{tabularx}
\end{table}

The evaluation schemes for the FB and XB datasets are listed in Tables~\ref{tab:2} and ~\ref{tab:3}, respectively. For the FB dataset, training and test splits adhered to the official PHM 2012 challenge guidelines \cite{Nectoux_2012_PRONOSTIA}. In the case of the XB dataset, only Conditions 1 and 2 were considered \cite{Deng_2024_Hybrid}, and the analysis was limited to bearings with outer-race faults \cite{Wang_2020_Hybrid}. Leave-one-out cross-validation was employed for the selected XB bearings, as listed in Table~\ref{tab:3}.

The training sequences in the RtF were nonstationary, with distributions that evolve. Following a previous study \cite{Tefera_2025_Cons}, each training RtF sequence was divided into three sampling intervals.

\begin{itemize}
    \item
    Healthy state: The initial 10\% of the RtF sequence, representing healthy operation.
    \item
    Slight degradation stage: The interval from 10 to 95\% of the RtF sequence, corresponding to the progressive degradation stage for sampling purposes.
    \item
    Sharp degradation stage: For sampling purposes, the final 5\% of the RtF sequence was treated as the stage of rapid degradation before failure.
\end{itemize}

Samples were randomly drawn from these intervals to form batches containing 20\% healthy, 70\% slightly degraded, and 10\% sharply degraded samples. After merging samples from all three intervals, each batch was partitioned such that 75\% was allocated for training and 25\% for validation. Early stopping was implemented, terminating training if the validation loss failed to improve by at least 0.0001 over 10 consecutive epochs.

\begin{table}[htbp]
    \centering
    \caption{Inputs and network architectures of the comparison models.}
    \label{tab:4}
    \small
    \setlength{\tabcolsep}{5pt}
    \renewcommand{\arraystretch}{1.2}
    \renewcommand{\multirowsetup}{\centering}
    \begin{tabularx}{\linewidth}{@{}>{\centering\arraybackslash}p{0.15\linewidth}>{\centering\arraybackslash}p{0.28\linewidth}>{\centering\arraybackslash}X@{}}
        \toprule
        \multirow[t]{2}{=}{Model} & \multicolumn{2}{c}{Description} \\
        & Input & Network architecture \\
        \midrule
        SAL-CNN \cite{Liu_2022_SALCNN} & STFT spectrogram. & CNN--CBAM feature extractor followed by an LSTM-based temporal encoder and a fully connected regression head \\
        CDCT \cite{Jiang_2023_Dual} & Raw vibration signal, Time-, and Frequency-domain feature. & Dual-branch causal CNN encoders followed by transformer-based feature integration and a fully connected regression head \\
        HA-CLSTM \cite{Zuo_2023_Hybrid} & Multichannel wavelet packet coefficient. & ConvLSTM backbone with channel and spatial attention, followed by global pooling and a fully connected regression head \\
        TT-CLSTM \cite{Niazi_2024_Multi} & Time-domain feature and CWT scalogram. & Parallel transformer and ConvLSTM branches with late feature fusion, followed by a fully connected regression head and Kalman filtering \\
        MLP-DT \cite{Deng_2024_Hybrid} & CWT scalogram. & MLP-based feature compression followed by an integrated DeepAR--transformer and a fully connected regression head \\
        \bottomrule
    \end{tabularx}
\end{table}

The input representations and network architectures of the five comparison models are listed in Table~\ref{tab:4}. These models were selected from DL-based approaches that learned the direct mapping from their respective input representations to the RUL. Specifically, CDCT \cite{Jiang_2023_Dual} directly mapped raw vibration signals and handcrafted time- and frequency-domain features to the RUL, whereas SAL-CNN \cite{Liu_2022_SALCNN}, HA-CLSTM \cite{Zuo_2023_Hybrid}, TT-CLSTM \cite{Niazi_2024_Multi}, and MLP-DT \cite{Deng_2024_Hybrid} directly mapped TFRs to the RUL.

\subsection{Experimental results}

This section presents quantitative results for RUL prediction, including visual comparisons of prediction trajectories and errors across RtF sequences, nominal MC-dropout-based uncertainty intervals, and assessments of computational complexity.

\begin{table}[htbp]
    \centering
    \caption{MAE and RMSE of the RUL prediction models for FB Condition 1.}
    \label{tab:5}
    \footnotesize
    \setlength{\tabcolsep}{3pt}
    \renewcommand{\arraystretch}{1.2}
    \begin{tabular}{@{}c*{12}{c}@{}}
        \toprule
        Test set & \multicolumn{2}{c}{SAL-CNN} & \multicolumn{2}{c}{CDCT} & \multicolumn{2}{c}{HA-CLSTM} & \multicolumn{2}{c}{TT-CLSTM} & \multicolumn{2}{c}{MLP-DT} & \multicolumn{2}{c}{FAAC-GRU} \\
        & MAE & RMSE & MAE & RMSE & MAE & RMSE & MAE & RMSE & MAE & RMSE & MAE & RMSE \\
        \midrule
        FB1-3 & 0.068 & 0.073 & 0.057 & 0.068 & 0.091 & 0.113 & 0.066 & 0.104 & \underline{0.043} & \underline{0.066} & \textbf{0.035} & \textbf{0.037} \\
        FB1-4 & 0.257 & 0.291 & 0.104 & \underline{0.138} & \underline{0.102} & 0.144 & 0.397 & 0.486 & 0.241 & 0.262 & \textbf{0.037} & \textbf{0.054} \\
        FB1-5 & 0.154 & 0.184 & 0.206 & 0.261 & 0.146 & \underline{0.170} & 0.281 & 0.343 & \underline{0.142} & 0.184 & \textbf{0.079} & \textbf{0.116} \\
        FB1-6 & 0.262 & 0.298 & 0.155 & 0.201 & \underline{0.128} & \underline{0.162} & 0.384 & 0.445 & 0.176 & 0.266 & \textbf{0.120} & \textbf{0.150} \\
        FB1-7 & 0.141 & 0.219 & \textbf{0.082} & \textbf{0.112} & 0.112 & 0.194 & 0.253 & 0.296 & 0.178 & 0.188 & \underline{0.111} & \underline{0.151} \\
        \midrule
        \(\mu\) & 0.176 & 0.213 & 0.121 & \underline{0.156} & \underline{0.116} & 0.157 & 0.276 & 0.335 & 0.156 & 0.193 & \textbf{0.076} & \textbf{0.102} \\
        \bottomrule
    \end{tabular}
\end{table}

\begin{table}[htbp]
    \centering
    \caption{MAE and RMSE of the RUL prediction models for FB Conditions 2 and 3.}
    \label{tab:6}
    \footnotesize
    \setlength{\tabcolsep}{3pt}
    \renewcommand{\arraystretch}{1.2}
    \begin{tabular}{@{}c*{12}{c}@{}}
        \toprule
        Test set & \multicolumn{2}{c}{SAL-CNN} & \multicolumn{2}{c}{CDCT} & \multicolumn{2}{c}{HA-CLSTM} & \multicolumn{2}{c}{TT-CLSTM} & \multicolumn{2}{c}{MLP-DT} & \multicolumn{2}{c}{FAAC-GRU} \\
        & MAE & RMSE & MAE & RMSE & MAE & RMSE & MAE & RMSE & MAE & RMSE & MAE & RMSE \\
        \midrule
        FB2-3 & \underline{0.137} & \underline{0.165} & 0.281 & 0.356 & 0.420 & 0.516 & 0.159 & 0.205 & 0.188 & 0.298 & \textbf{0.123} & \textbf{0.154} \\
        FB2-4 & \textbf{0.169} & \textbf{0.224} & 0.356 & 0.416 & 0.454 & 0.538 & 0.281 & 0.340 & \underline{0.220} & \underline{0.251} & 0.279 & 0.305 \\
        FB2-5 & \underline{0.198} & \underline{0.235} & 0.315 & 0.388 & 0.466 & 0.535 & 0.328 & 0.403 & 0.212 & 0.241 & \textbf{0.156} & \textbf{0.202} \\
        FB2-6 & \underline{0.135} & \textbf{0.149} & 0.287 & 0.350 & 0.397 & 0.473 & 0.317 & 0.356 & \textbf{0.118} & \underline{0.164} & 0.147 & 0.188 \\
        FB2-7 & 0.325 & 0.336 & 0.402 & 0.472 & 0.402 & 0.493 & \textbf{0.216} & \textbf{0.249} & \underline{0.278} & \underline{0.310} & 0.285 & 0.333 \\
        FB3-3 & 0.330 & 0.400 & \underline{0.270} & \underline{0.332} & 0.324 & 0.386 & 0.353 & 0.407 & 0.325 & 0.364 & \textbf{0.268} & \textbf{0.301} \\
        \midrule
        \(\mu\) & \underline{0.216} & \underline{0.252} & 0.319 & 0.386 & 0.411 & 0.490 & 0.276 & 0.327 & 0.223 & 0.271 & \textbf{0.210} & \textbf{0.247} \\
        \bottomrule
    \end{tabular}
\end{table}

A comparative analysis of the RUL prediction performances of the FB test bearings under the three conditions is presented in Tables~\ref{tab:5} and ~\ref{tab:6}. The relative performances of the comparison models varied across conditions. For Condition 1, HA-CLSTM achieved the lowest mean MAE among the comparison models and CDCT achieved the lowest mean RMSE; however, its prediction errors increased under Conditions 2 and 3, where SAL-CNN demonstrated the lowest mean MAE and RMSE among the comparison models for Conditions 2 and 3. These results indicate that the relative accuracies of the evaluated models were sensitive to the specific operating conditions. Notably, FAAC-GRU consistently achieved the lowest mean prediction errors across all conditions, with mean MAE and RMSE values of 0.076 and 0.102 for Condition 1, and 0.210 and 0.247 for Conditions 2 and 3. These results highlight the predictive performance across the operating conditions evaluated in this study.

\begin{table}[htbp]
    \centering
    \caption{MAE and RMSE of the RUL prediction models for XB Condition 1.}
    \label{tab:7}
    \footnotesize
    \setlength{\tabcolsep}{3pt}
    \renewcommand{\arraystretch}{1.2}
    \begin{tabular}{@{}c*{12}{c}@{}}
        \toprule
        Test set & \multicolumn{2}{c}{SAL-CNN} & \multicolumn{2}{c}{CDCT} & \multicolumn{2}{c}{HA-CLSTM} & \multicolumn{2}{c}{TT-CLSTM} & \multicolumn{2}{c}{MLP-DT} & \multicolumn{2}{c}{FAAC-GRU} \\
        & MAE & RMSE & MAE & RMSE & MAE & RMSE & MAE & RMSE & MAE & RMSE & MAE & RMSE \\
        \midrule
        XB1-1 & 0.137 & 0.164 & 0.130 & 0.182 & 0.165 & 0.193 & 0.177 & 0.215 & \textbf{0.126} & \textbf{0.143} & \underline{0.129} & \underline{0.161} \\
        XB1-2 & 0.228 & 0.256 & 0.205 & 0.265 & \underline{0.166} & \textbf{0.197} & 0.174 & \underline{0.227} & 0.295 & 0.322 & \textbf{0.163} & 0.201 \\
        XB1-3 & 0.213 & 0.241 & \textbf{0.136} & \textbf{0.157} & 0.233 & 0.263 & 0.267 & 0.304 & \underline{0.148} & \underline{0.186} & 0.167 & \underline{0.197} \\
        XB1-5 & 0.224 & 0.272 & 0.198 & \underline{0.230} & \underline{0.192} & 0.243 & 0.218 & 0.301 & 0.277 & 0.309 & \textbf{0.145} & \textbf{0.201} \\
        \midrule
        \(\mu\) & 0.201 & 0.233 & \underline{0.167} & \underline{0.209} & 0.189 & 0.224 & 0.209 & 0.261 & 0.211 & 0.240 & \textbf{0.151} & \textbf{0.190} \\
        \bottomrule
    \end{tabular}
\end{table}

\begin{table}[htbp]
    \centering
    \caption{MAE and RMSE of the RUL prediction models for XB Condition 2.}
    \label{tab:8}
    \footnotesize
    \setlength{\tabcolsep}{3pt}
    \renewcommand{\arraystretch}{1.2}
    \begin{tabular}{@{}c*{12}{c}@{}}
        \toprule
        Test set & \multicolumn{2}{c}{SAL-CNN} & \multicolumn{2}{c}{CDCT} & \multicolumn{2}{c}{HA-CLSTM} & \multicolumn{2}{c}{TT-CLSTM} & \multicolumn{2}{c}{MLP-DT} & \multicolumn{2}{c}{FAAC-GRU} \\
        & MAE & RMSE & MAE & RMSE & MAE & RMSE & MAE & RMSE & MAE & RMSE & MAE & RMSE \\
        \midrule
        XB2-2 & 0.180 & 0.226 & \underline{0.087} & \textbf{0.097} & 0.182 & 0.206 & 0.120 & 0.161 & 0.232 & 0.267 & \textbf{0.081} & \underline{0.101} \\
        XB2-4 & \underline{0.178} & \underline{0.205} & 0.243 & 0.282 & 0.248 & 0.291 & 0.201 & 0.234 & 0.238 & 0.265 & \textbf{0.168} & \textbf{0.190} \\
        XB2-5 & 0.240 & 0.285 & 0.185 & 0.219 & \textbf{0.119} & \textbf{0.162} & 0.161 & 0.211 & 0.430 & 0.509 & \underline{0.160} & \underline{0.199} \\
        \midrule
        \(\mu\) & 0.199 & 0.239 & 0.172 & \underline{0.199} & 0.183 & 0.220 & \underline{0.161} & 0.202 & 0.300 & 0.347 & \textbf{0.136} & \textbf{0.163} \\
        \bottomrule
    \end{tabular}
\end{table}

The MAE and RMSE of the XB test bearings under Conditions 1 and 2 are presented in Tables~\ref{tab:7} and ~\ref{tab:8}, respectively. The model rankings differed from those observed for the FB dataset. For example, the CDCT achieved the second-lowest mean RMSE under both XB conditions, despite its comparatively weaker performance on the FB dataset. Therefore, the relative performance of a given RUL prediction model can vary according to the dataset and operating conditions. FAAC-GRU achieved the lowest mean MAE and RMSE under both XB conditions, which was consistent with the mean error comparisons for the FB dataset. The other models achieved lower errors for certain individual test bearings. Therefore, the aggregate results corroborate the performance of the FAAC-GRU on the two datasets and operating conditions evaluated in this study.

\begin{table}[htbp]
    \centering
    \caption{Mean RMSE and score for the FB and XB test bearings.}
    \label{tab:9}
    \footnotesize
    \setlength{\tabcolsep}{4pt}
    \renewcommand{\arraystretch}{1.2}
    \begin{tabular}{@{}cc*{6}{c}@{}}
        \toprule
        Metric & Test sets & SAL-CNN & CDCT & HA-CLSTM & TT-CLSTM & MLP-DT & FAAC-GRU \\
        \midrule
        \multirow{2}{*}{RMSE} & FBs & \underline{0.234} & 0.281 & 0.338 & 0.330 & 0.235 & \textbf{0.181} \\
        & XBs & 0.235 & \underline{0.204} & 0.222 & 0.236 & 0.285 & \textbf{0.178} \\
        \midrule
        \multirow{2}{*}{Score} & FBs & 23.064 & 30.083 & 36.282 & 39.519 & \underline{19.003} & \textbf{16.246} \\
        & XBs & 2.643 & \underline{2.024} & 2.196 & 2.357 & 3.501 & \textbf{1.915} \\
        \bottomrule
    \end{tabular}
\end{table}

A comparison of the mean RMSE and scores across the FB and XB test bearings is presented in Table~\ref{tab:9}. Among the comparison models, SAL-CNN achieved the lowest mean RMSE on the FB dataset, whereas MLP-DT achieved the lowest Score. CDCT achieved the lowest values for both metrics on the XB dataset. These differences indicate that the relative performances of the comparison models varied depending on the dataset. However, the FAAC-GRU achieved the lowest mean RMSE and Score on both datasets, indicating that its predictive advantage was robust to changes in dataset and evaluation metric. This consistent performance underscores the effectiveness of the proposed architecture in both evaluated settings.

\begin{figure}[htbp]
	\centering
	\includegraphics[width=\linewidth]{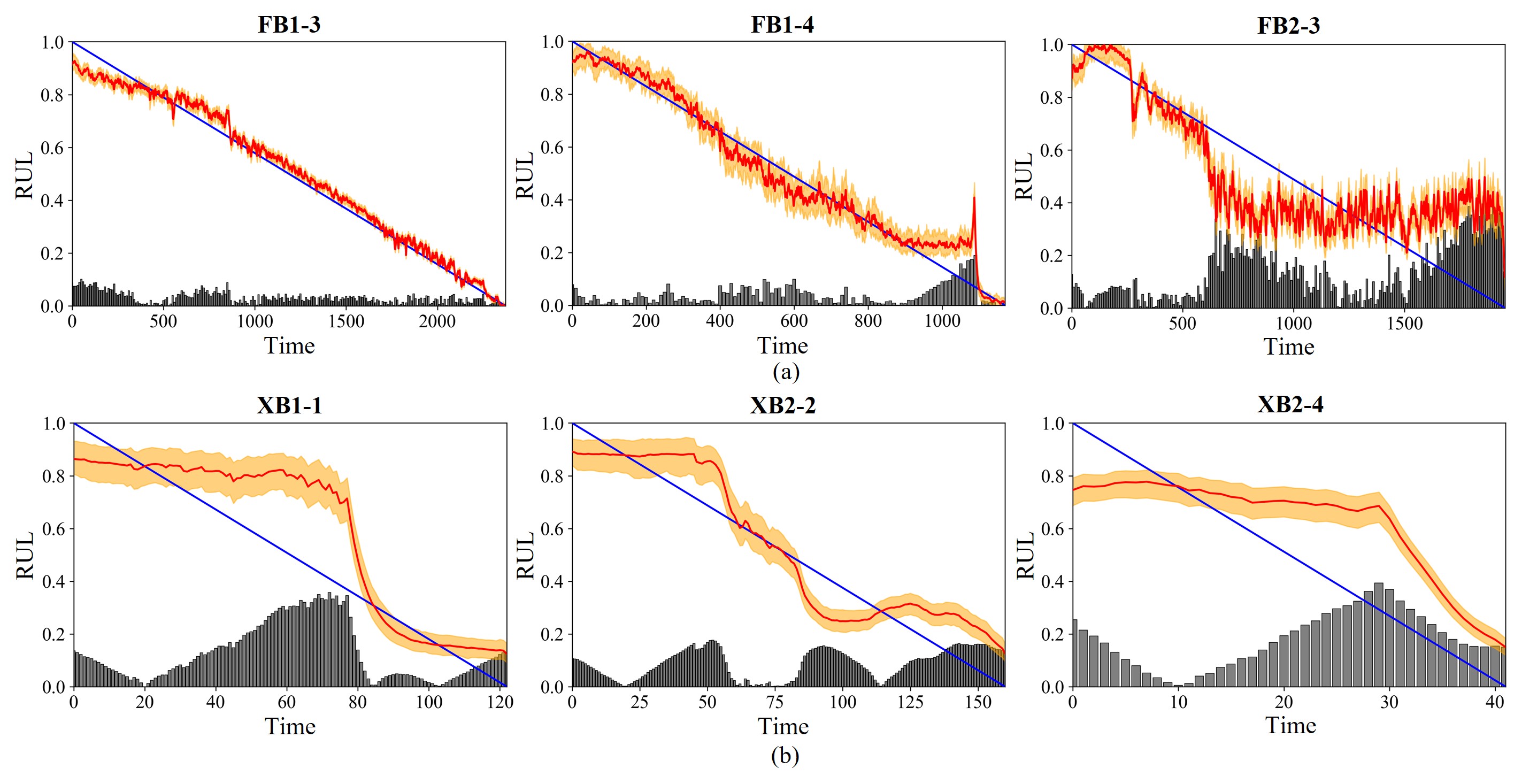}
	\caption{RUL predictions over the RtF sequences of selected FB and XB test bearings.}
	\label{fig:figure8}
\end{figure}

The RUL predictions for the representative test bearings are shown in Figure~\ref{fig:figure8}. The horizontal and vertical axes denote time and normalized RUL, respectively. The blue line represents the true RUL labels, whereas the red line represents the predicted mean RUL. The orange shaded region indicates the nominal 95\% confidence interval estimated via MC dropout. The black bars at the bottom of each subplot represent prediction errors at each time point.

For FB1-3 in Figure~\ref{fig:figure8}(a), the predicted trajectory followed the overall RUL trend with relatively minor errors. Similarly, the predictions for FB1-4 displayed a decreasing RUL trend, although the errors increased near failure. Larger errors were observed for FB2-3. As shown in Appendix~\ref{sec:appB}, the raw signals for FB1-3 and FB1-4 displayed comparatively gradual changes, whereas FB2-3 displayed substantial fluctuations during early operation. These signal characteristics likely contributed to the increased difficulty in predicting FB2-3. However, FAAC-GRU maintained a comparatively consistent RUL trend over time, even in the presence of such fluctuations.

\begin{figure}[htbp]
	\centering
	\includegraphics[width=\linewidth]{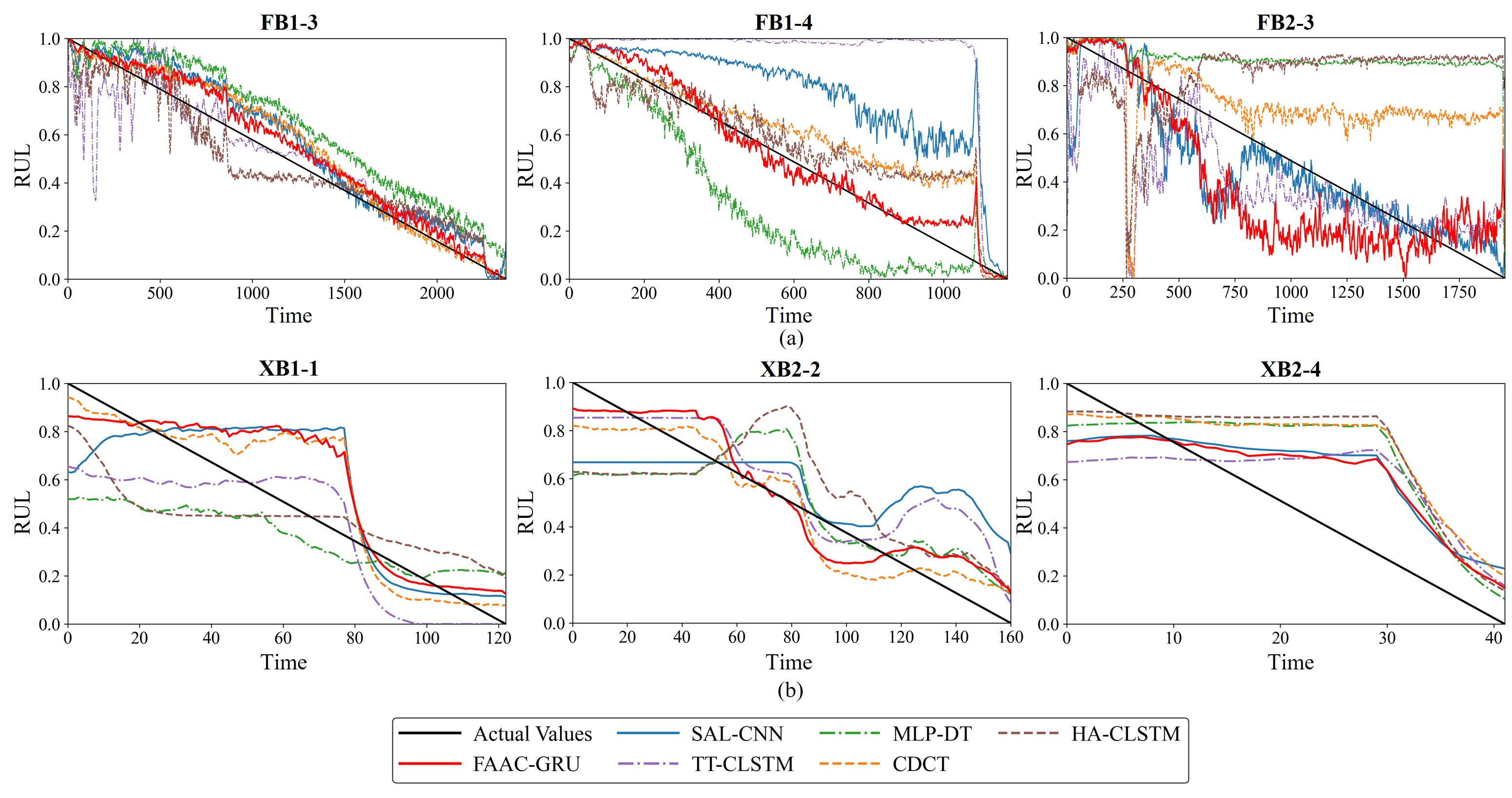}
	\caption{RUL prediction trajectories of the evaluated models for the (a) FB and (b) XB test bearings.}
	\label{fig:figure9}
\end{figure}

The RUL prediction trajectories for the FB and XB test bearings are shown in Figure~\ref{fig:figure9}(a) and (b), respectively. In each subplot, the black line denotes the true RUL, whereas the other lines represent predictions from different models. Prediction behavior varied across the bearings and models. In Figure~\ref{fig:figure9}(a), most models capture the decreasing RUL trend for FB1-3. SAL-CNN demonstrates comparatively strong performance on FB1-3 and FB2-3, but tends to overestimate the RUL throughout much of the FB1-4 sequence. CDCT closely follows the trajectories of FB1-3 and FB1-4; however, it demonstrated abrupt prediction shifts at early operational stages for FB2-3. A similar variability was observed for the XB bearings in Figure~\ref{fig:figure9}(b): although some models accurately tracked one bearing, they may overestimate (SAL-CNN and CDCT for XB1-1) or underestimate (TT-CLSTM, HA-CLSTM, and MLP-DT for XB2-2) on others. These examples highlight the differences in prediction behavior across the evaluated bearings.

The FAAC-GRU consistently aligns with the true RUL trajectories in the presented cases. For FB1-4, its predictions remain relatively stable and accurate throughout the sequence, whereas several comparison models demonstrated overestimation (SAL-CNN, TT-CLSTM, CDCT, and HA-CLSTM) or underestimation (MLP-DT). For FB2-3, other models demonstrated substantial fluctuations, particularly around time step 250, where an abrupt change occurred in the raw signal (see Appendix~\ref{sec:appB}, Figure~\ref{fig:figureb1}). However, FAAC-GRU maintained a relatively consistent, decreasing RUL trend through this transition.

\begin{table}[htbp]
    \centering
    \caption{Model parameter counts and computational cost per input sample}
    \label{tab:10}
    \small
    \setlength{\tabcolsep}{5pt}
    \renewcommand{\arraystretch}{1.2}
    \begin{tabular}{@{}cccc@{}}
        \toprule
        Model & \multicolumn{3}{c}{Computational complexity} \\
        & Parameters (M) & FLOPs (G) & Inference time (ms) \\
        \midrule
        SAL-CNN & 0.111 & 0.149 & 0.078 \\
        CDCT & 1.419 & 0.018 & 0.069 \\
        HA-CLSTM & 0.053 & 2.119 & 0.340 \\
        TT-CLSTM & 0.638 & 1.804 & 0.195 \\
        MLP-DT & 0.772 & 0.451 & 0.109 \\
        FAAC-GRU & 0.939 & 9.814 & 1.306 \\
        \bottomrule
    \end{tabular}
\end{table}

A comparison of the models in terms of the parameter count, floating-point operations (FLOPs), and inference time per sample is presented in Table~\ref{tab:10}. The mean inference time was calculated over 1,000 runs for each model under identical experimental conditions. All experiments were conducted using Python 3.8 and PyTorch, on an NVIDIA GeForce RTX 4080 GPU, CUDA 12.6, and driver version 560.94. For computational comparison, TFR inputs were resized to (64, 64) using bilinear interpolation, and the input sequence length for models utilizing multiple sequential inputs was fixed at 5. The detailed layer arrangement and output shapes of the FAAC-GRU are presented in Appendix~\ref{sec:appC} and Table~\ref{tab:c1}.

Among the evaluated models, HA-CLSTM had the lowest parameter count (0.053 M), whereas CDCT achieved the lowest operation count (0.018 G FLOPs) and the shortest inference time (0.069 ms). Notably, HA-CLSTM also demonstrated the longest inference time (0.340 ms) among the five comparison models. These findings indicate that the parameter count alone does not determine the inference cost; the number of network operations and their implementation significantly influence the runtime. FAAC-GRU comprised 0.939 M parameters, required 9.814 G FLOPs, and achieved a mean inference time of 1.306 ms per sample. It had the highest operation count and longest measured inference time among the models evaluated, despite CDCT having more parameters. The high operation count and comparatively long inference time represent a computational limitation of FAAC-GRU, particularly for deployment on resource-constrained hardware.

\section{Ablation study}
\label{sec:ablat}

This section presents the ablation experiments to evaluate the contribution of each FAAC--GRU component. Section~\ref{sec:5.1} compares attention modules, Section~\ref{sec:5.2} compares feature-map aggregation methods, and Section~\ref{sec:5.3} compares isotropic and anisotropic convolutional kernel designs.

\subsection{Channel block attention module design}
\label{sec:5.1}

This section compares the performance of CBAM and DCBAM within FAAC-GRU, with all other architectural components held constant for a fair comparison.

\begin{table}[htbp]
    \centering
    \caption{RUL prediction performance of FAAC-GRU with CBAM or DCBAM on FB test bearings under three operating conditions.}
    \label{tab:11}
    \footnotesize
    \setlength{\tabcolsep}{3pt}
    \renewcommand{\arraystretch}{1.2}
    \begin{tabular}{@{}c*{4}{c}@{\hspace{0.5em}}c*{4}{c}@{}}
        \toprule
        Test set & \multicolumn{2}{c}{CBAM} & \multicolumn{2}{c}{DCBAM} & Test set & \multicolumn{2}{c}{CBAM} & \multicolumn{2}{c}{DCBAM} \\
        Condition 1 & MAE & RMSE & MAE & RMSE & Condition 2\&3 & MAE & RMSE & MAE & RMSE \\
        \midrule
        FB1-3 & 0.039 & 0.048 & \textbf{0.035} & \textbf{0.037} & FB2-3 & 0.139 & 0.167 & \textbf{0.123} & \textbf{0.154} \\
        FB1-4 & 0.058 & 0.066 & \textbf{0.037} & \textbf{0.054} & FB2-4 & 0.344 & 0.411 & \textbf{0.279} & \textbf{0.305} \\
        FB1-5 & \textbf{0.076} & \textbf{0.111} & 0.079 & 0.116 & FB2-5 & 0.175 & 0.209 & \textbf{0.156} & \textbf{0.202} \\
        FB1-6 & \textbf{0.099} & \textbf{0.137} & 0.120 & 0.150 & FB2-6 & \textbf{0.088} & \textbf{0.125} & 0.147 & 0.188 \\
        FB1-7 & 0.127 & 0.163 & \textbf{0.111} & \textbf{0.151} & FB2-7 & 0.337 & 0.397 & \textbf{0.285} & \textbf{0.333} \\
        &  &  &  &  & FB3-3 & 0.289 & 0.358 & \textbf{0.268} & \textbf{0.301} \\
        \midrule
        \(\mu\) & 0.080 & 0.105 & \textbf{0.076} & \textbf{0.102} & \(\mu\) & 0.229 & 0.278 & \textbf{0.210} & \textbf{0.247} \\
        \bottomrule
    \end{tabular}
\end{table}

\begin{table}[htbp]
    \centering
    \caption{RUL prediction performance of FAAC-GRU with CBAM or DCBAM on XB test bearings under two operating conditions.}
    \label{tab:12}
    \footnotesize
    \setlength{\tabcolsep}{3pt}
    \renewcommand{\arraystretch}{1.2}
    \begin{tabular}{@{}c*{4}{c}@{\hspace{0.5em}}c*{4}{c}@{}}
        \toprule
        Test set & \multicolumn{2}{c}{CBAM} & \multicolumn{2}{c}{DCBAM} & Test set & \multicolumn{2}{c}{CBAM} & \multicolumn{2}{c}{DCBAM} \\
        Condition 1 & MAE & RMSE & MAE & RMSE & Condition 2 & MAE & RMSE & MAE & RMSE \\
        \midrule
        XB1-1 & \textbf{0.112} & \textbf{0.155} & 0.129 & 0.161 & XB2-2 & 0.084 & 0.105 & \textbf{0.081} & \textbf{0.101} \\
        XB1-2 & 0.186 & 0.216 & \textbf{0.163} & \textbf{0.201} & XB2-4 & \textbf{0.160} & \textbf{0.183} & 0.168 & 0.190 \\
        XB1-3 & 0.209 & 0.224 & \textbf{0.167} & \textbf{0.197} & XB2-5 & 0.176 & 0.207 & \textbf{0.160} & \textbf{0.199} \\
        XB1-5 & 0.150 & 0.206 & \textbf{0.145} & \textbf{0.201} &  &  &  &  &  \\
        \midrule
        \(\mu\) & 0.164 & 0.200 & \textbf{0.151} & \textbf{0.190} & \(\mu\) & 0.140 & 0.165 & \textbf{0.136} & \textbf{0.163} \\
        \bottomrule
    \end{tabular}
\end{table}

A comparative analysis of the CBAM and DCBAM for the FB and XB test bearings is presented in Tables~\ref{tab:11} and ~\ref{tab:12}, respectively. Under FB Condition 1, CBAM achieved MAE (0.080) and RMSE (0.105) and DCBAM achieved MAE (0.076) and RMSE (0.102), although discrepancies were observed at the level of individual bearings. In FB conditions 2 and 3, DCBAM reduced MAE from 0.229 to 0.210 and mean RMSE from 0.278 to 0.247. DCBAM had a lower MAE for FB2-3, FB2-4, FB2-5, FB2-7, and FB3-3, with particularly significant reductions for FB2-4 and FB2-7; however, CBAM yielded better results on FB2-6. Similar trends were observed in the XB dataset. For XB Condition 1, the MAE and RMSE decreased from 0.164 and 0.200 to 0.151 and 0.190, respectively. Under XB Condition 2, the MAE and RMSE decreased from 0.140 to 0.136 and 0.165 to 0.163.

Overall, replacing CBAM with DCBAM consistently maintained or enhanced the mean performance across all condition groups, with the most significant improvements observed under FB Conditions 2 and 3. However, these benefits were not uniformly distributed across individual bearings. The observed differences in mean error were generally modest, providing limited empirical justification for incorporating directional operations in the attention module.

\subsection{Feature map aggregation design}
\label{sec:5.2}

The approach used to aggregate feature maps into a compact vector directly influences the representational capacity available to subsequent layers and the computational cost of the model. Three aggregation methods were compared: flattening (FLAT), GAP, and the proposed DAP. The prediction performance and computational cost were jointly evaluated, as the aggregation methods generate distinct representations and alter the dimensionality of subsequent layers.

\begin{table}[htbp]
    \centering
    \caption{Parameter counts and computational cost per sample for three feature map aggregation variants of FAAC-GRU.}
    \label{tab:13}
    \small
    \setlength{\tabcolsep}{5pt}
    \renewcommand{\arraystretch}{1.2}
    \begin{tabular}{@{}cccc@{}}
        \toprule
        Model & \multicolumn{3}{c}{Computational complexity} \\
        & No. of parameters (M) & No. of FLOPs (G) & Inference time (ms) \\
        \midrule
        FLAT & 5.083 & 10.018 & 1.329 \\
        GAP & 0.743 & 9.811 & 1.301 \\
        DAP & 0.939 & 9.814 & 1.306 \\
        \bottomrule
    \end{tabular}
\end{table}

A comparison of the computational costs of the three aggregation variants is presented in Table~\ref{tab:13}. The FLAT method demonstrated the highest parameter count (5.083 M) and required the most operations (10.018 G FLOPs). Notably, the flattening operation itself merely reshapes the feature map; the increased computational burden stems from the larger representation processed by the subsequent learned layers. By contrast, GAP utilized 0.905 M parameters and required 9.811 G FLOPs, whereas DAP involved 0.939 M parameters and 9.814 G FLOPs, reflecting only a marginal increase in both parameters and operations relative to GAP. The measured inference times for FLAT, GAP, and DAP were 1.329 ms, 1.301 ms, and 1.306 ms, respectively.

\begin{table}[htbp]
    \centering
    \caption{RUL prediction performance of three FAAC-GRU aggregation variants on FB test bearings under three operating conditions.}
    \label{tab:14}
    \small
    \setlength{\tabcolsep}{5pt}
    \renewcommand{\arraystretch}{1.2}
    \begin{tabular}{@{}cc*{6}{c}@{}}
        \toprule
        Condition & Test set & \multicolumn{2}{c}{FLAT} & \multicolumn{2}{c}{GAP} & \multicolumn{2}{c}{DAP} \\
        &  & MAE & RMSE & MAE & RMSE & MAE & RMSE \\
        \midrule
        FB Condition 1 & FB1-3 & \underline{0.039} & \underline{0.041} & 0.046 & 0.055 & \textbf{0.035} & \textbf{0.037} \\
        & FB1-4 & 0.209 & 0.256 & \underline{0.109} & \underline{0.140} & \textbf{0.037} & \textbf{0.054} \\
        & FB1-5 & 0.109 & 0.163 & \textbf{0.053} & \textbf{0.109} & \underline{0.079} & \underline{0.116} \\
        & FB1-6 & \underline{0.135} & \underline{0.185} & 0.161 & 0.195 & \textbf{0.120} & \textbf{0.150} \\
        & FB1-7 & \underline{0.148} & \underline{0.187} & 0.155 & 0.198 & \textbf{0.111} & \textbf{0.151} \\
        \cmidrule(l){2-8}
        & \(\mu\) & 0.128 & 0.166 & \underline{0.105} & \underline{0.139} & \textbf{0.076} & \textbf{0.102} \\
        \midrule
        FB Condition 2\&3 & FB2-3 & 0.187 & 0.215 & \underline{0.176} & \underline{0.201} & \textbf{0.123} & \textbf{0.154} \\
        & FB2-4 & \underline{0.333} & \underline{0.387} & 0.368 & 0.414 & \textbf{0.279} & \textbf{0.305} \\
        & FB2-5 & 0.201 & 0.234 & \textbf{0.137} & \textbf{0.168} & \underline{0.156} & \underline{0.202} \\
        & FB2-6 & \textbf{0.092} & \textbf{0.116} & 0.153 & 0.192 & \underline{0.147} & \underline{0.188} \\
        & FB2-7 & 0.318 & 0.359 & \underline{0.300} & \underline{0.358} & \textbf{0.285} & \textbf{0.333} \\
        & FB3-3 & \underline{0.275} & \underline{0.328} & 0.296 & 0.377 & \textbf{0.268} & \textbf{0.301} \\
        \cmidrule(l){2-8}
        & \(\mu\) & \underline{0.234} & \underline{0.273} & 0.238 & 0.285 & \textbf{0.210} & \textbf{0.247} \\
        \bottomrule
    \end{tabular}
\end{table}

\begin{table}[htbp]
    \centering
    \caption{RUL prediction performance of three FAAC-GRU aggregation variants on XB test bearings under two operating conditions.}
    \label{tab:15}
    \small
    \setlength{\tabcolsep}{5pt}
    \renewcommand{\arraystretch}{1.2}
    \begin{tabular}{@{}cc*{6}{c}@{}}
        \toprule
        Condition & Test set & \multicolumn{2}{c}{FLAT} & \multicolumn{2}{c}{GAP} & \multicolumn{2}{c}{DAP} \\
        &  & MAE & RMSE & MAE & RMSE & MAE & RMSE \\
        \midrule
        XB Condition 1 & XB1-1 & 0.147 & 0.183 & \textbf{0.124} & \textbf{0.160} & \underline{0.129} & \underline{0.161} \\
        & XB1-2 & 0.204 & 0.248 & \underline{0.178} & \underline{0.206} & \textbf{0.163} & \textbf{0.201} \\
        & XB1-3 & 0.199 & 0.227 & \underline{0.189} & \underline{0.209} & \textbf{0.167} & \textbf{0.197} \\
        & XB1-5 & 0.148 & \textbf{0.171} & \textbf{0.142} & \underline{0.200} & \underline{0.145} & 0.201 \\
        \cmidrule(l){2-8}
        & \(\mu\) & 0.174 & 0.207 & \underline{0.158} & \underline{0.194} & \textbf{0.151} & \textbf{0.190} \\
        \midrule
        XB Condition 2 & XB2-2 & 0.081 & 0.102 & \textbf{0.079} & \textbf{0.096} & 0.081 & \underline{0.101} \\
        & XB2-4 & \textbf{0.145} & \textbf{0.175} & 0.179 & 0.202 & \underline{0.168} & \underline{0.190} \\
        & XB2-5 & 0.179 & 0.207 & \underline{0.169} & \textbf{0.192} & \textbf{0.160} & \underline{0.199} \\
        \cmidrule(l){2-8}
        & \(\mu\) & \textbf{0.135} & \textbf{0.161} & 0.142 & 0.163 & \underline{0.136} & \underline{0.163} \\
        \bottomrule
    \end{tabular}
\end{table}

A comparison of the three aggregation methods for the FB and XB bearings is presented in Tables~\ref{tab:14} and ~\ref{tab:15}, respectively. Under FB Condition 1, DAP achieved the lowest MAE (0.076) and RMSE (0.102), compared with 0.105 and 0.139 for GAP and 0.128 and 0.166 for FLAT. Differences among the aggregation methods were particularly pronounced for FB1-4, where the MAE values for FLAT, GAP, and DAP were 0.209, 0.109, and 0.037, respectively. Under FB Conditions 2 and 3, DAP reduced the MAE and RMSE by 0.028 and 0.038, respectively, compared with GAP, and by 0.024 and 0.026, respectively, compared with FLAT. These findings support the effectiveness of adaptive aggregation for RUL prediction on the FB dataset. However, the improvement was not consistent across all cases: DAP produces larger errors than GAP for certain bearings, and FLAT achieved the lowest MAE and RMSE under the XB Condition 2. Nearly uniform attention weights can result in behaviors similar to constant average pooling, whereas insufficient weights on relevant features may limit the advantages of adaptive aggregation.

\begin{figure}[htbp]
	\centering
	\includegraphics[width=\linewidth]{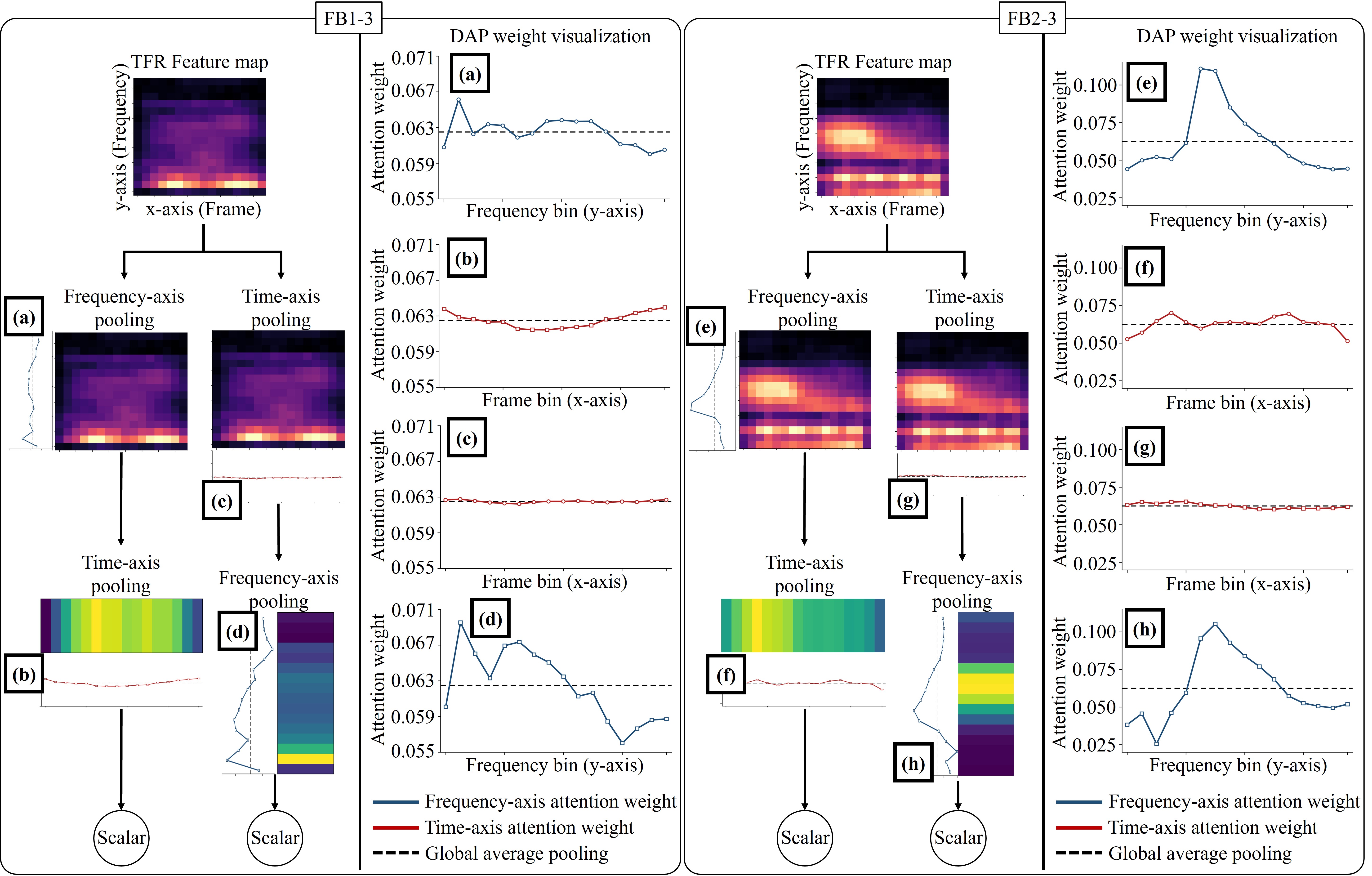}
	\caption{DAP attention weights for FB1-3 and FB2-3. Input samples were selected at 25\% of each RtF sequence; the first feature channel is shown.}
	\label{fig:figure10}
\end{figure}

\begin{figure}[htbp]
	\centering
	\includegraphics[width=\linewidth]{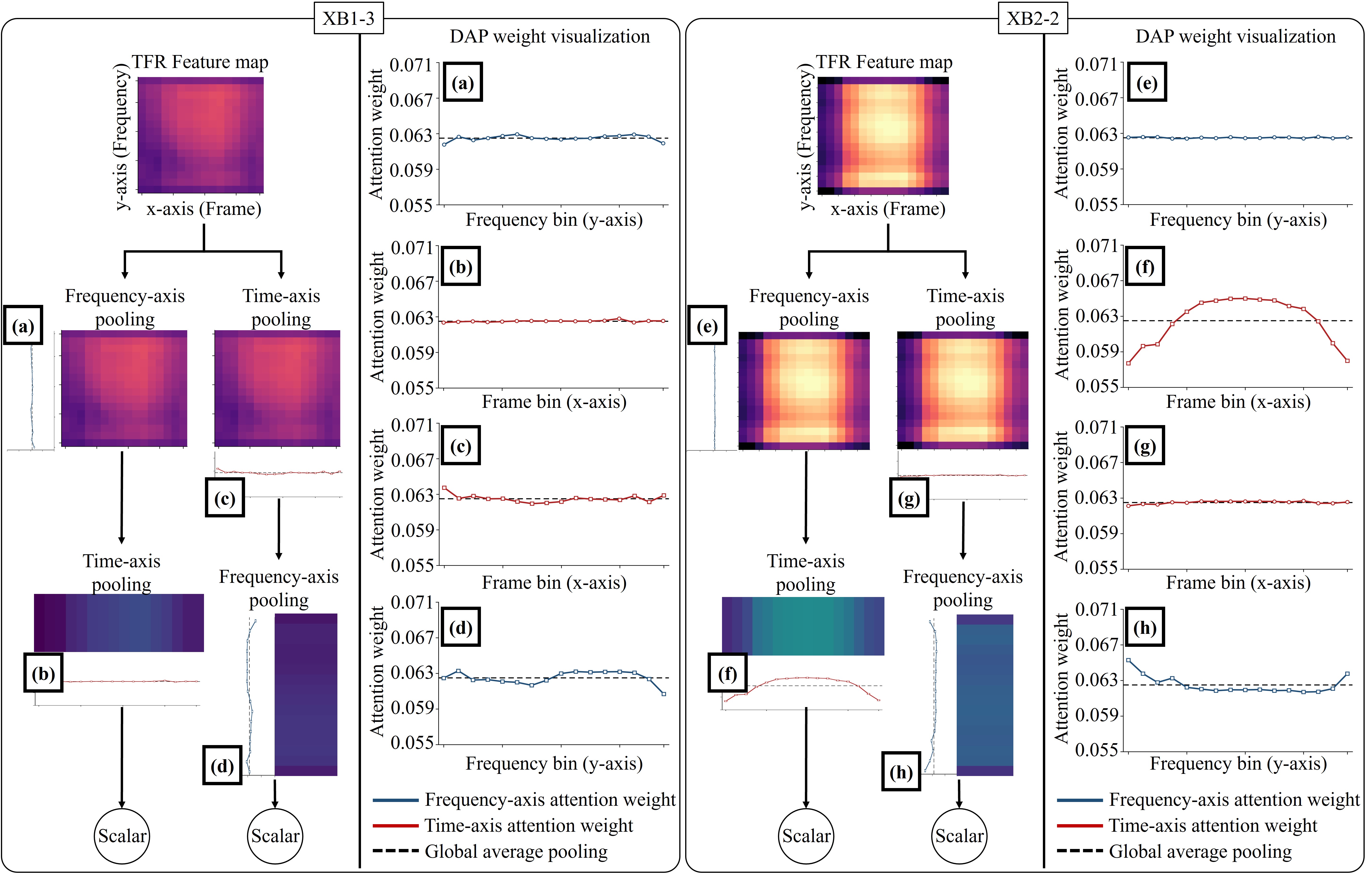}
	\caption{DAP attention weights for XB1-3 and XB2-2. Input samples were selected at 25\% of each RtF sequence; the first feature channel is shown.}
	\label{fig:figure11}
\end{figure}

The attention weights employed by DAP to aggregate the feature maps in the trained FAAC-GRU are shown in Figures~\ref{fig:figure10} and ~\ref{fig:figure11}. DAP utilizes four attention mechanisms: frequency- and time-axis attention in the frequency-to-time path, and time- and frequency-axis attention in the time-to-frequency path. The corresponding weight distributions for each test bearing are presented, with enlarged views provided on the right. The frequency- and time-axis attention weights are plotted in blue and red (Figures (a)--(h)), respectively, with the GAP indicated by black dashed lines.

The feature maps and DAP aggregation for the samples selected at 25\% of the FB1-3 and FB2-3 RtF sequences are shown in Figure~\ref{fig:figure10}. For FB1-3, strong activations were observed toward the low-frequency end of the feature map. The frequency-axis weights in panel (a) are higher at these positions, whereas the time-axis weights in panels (b) and (c) demonstrate less variation, which is consistent with activations extending across the local time axis. The frequency-axis weights in panel (d) further emphasize selected positions following the time-axis aggregation in panel (c). For FB2-3, the weights in panel (e) emphasize the activated frequency-axis positions, and those in panel (h) emphasize intermediate positions in the vector obtained from panel (g). The time-axis weights in panel (f) vary across positions but approach uniformity in certain regions. These examples demonstrate input-dependent aggregation, with the frequency-axis weights displaying more significant variability than those along the time axis.

The corresponding DAP results for XB1-3 and XB2-2 are shown in Figure~\ref{fig:figure11}. The feature map for XB1-3 displays relatively diffuse activations across frequency-axis positions, and the weight distributions in panels (a)--(d) were close to those of the GAP. By contrast, for XB2-2, the activation extended across much of the feature map. The weights in panels (e) and (g) were nearly uniform, whereas panels (f) and (h) assigned greater weights to the selected positions. For example, weights tend to be higher near the center than at the boundaries of the plotted axis, aligning with regions of stronger feature map activation. These findings indicate that DAP is capable of producing either near-uniform or selective aggregation, depending on the input.

\subsection{Convolution kernel design}
\label{sec:5.3}

This section compares the isotropic kernel variant (ISO) and anisotropic kernel variant (ANISO) of the convolutional feature extractor. Both variants utilize identical DCBAM and DAP modules, ensuring that attention and feature aggregation remain consistent during kernel evaluation.

\begin{table}[htbp]
    \centering
    \caption{RUL prediction performance of ISO and ANISO on FB test bearings under three operating conditions.}
    \label{tab:16}
    \footnotesize
    \setlength{\tabcolsep}{3pt}
    \renewcommand{\arraystretch}{1.2}
    \begin{tabular}{@{}c*{4}{c}@{\hspace{0.5em}}c*{4}{c}@{}}
        \toprule
        Test set & \multicolumn{2}{c}{ISO} & \multicolumn{2}{c}{ANISO} & Test set & \multicolumn{2}{c}{ISO} & \multicolumn{2}{c}{ANISO} \\
        Condition 1 & MAE & RMSE & MAE & RMSE & Condition 2\&3 & MAE & RMSE & MAE & RMSE \\
        \midrule
        FB1-3 & 0.046 & 0.052 & \textbf{0.035} & \textbf{0.037} & FB2-3 & 0.165 & 0.204 & \textbf{0.123} & \textbf{0.154} \\
        FB1-4 & 0.147 & 0.168 & \textbf{0.037} & \textbf{0.054} & FB2-4 & 0.295 & 0.338 & \textbf{0.279} & \textbf{0.305} \\
        FB1-5 & 0.085 & 0.127 & \textbf{0.079} & \textbf{0.116} & FB2-5 & 0.187 & 0.212 & \textbf{0.156} & \textbf{0.202} \\
        FB1-6 & 0.123 & 0.165 & \textbf{0.120} & \textbf{0.150} & FB2-6 & \textbf{0.118} & \textbf{0.133} & 0.147 & 0.188 \\
        FB1-7 & 0.149 & 0.179 & \textbf{0.111} & \textbf{0.151} & FB2-7 & 0.319 & 0.374 & \textbf{0.285} & \textbf{0.333} \\
        &  &  &  &  & FB3-3 & \textbf{0.261} & \textbf{0.300} & 0.268 & 0.301 \\
        \midrule
        \(\mu\) & 0.110 & 0.138 & \textbf{0.076} & \textbf{0.102} & \(\mu\) & 0.224 & 0.260 & \textbf{0.210} & \textbf{0.247} \\
        \bottomrule
    \end{tabular}
\end{table}

\begin{table}[htbp]
    \centering
    \caption{RUL prediction performance of ISO and ANISO on XB test bearings under two operating conditions.}
    \label{tab:17}
    \footnotesize
    \setlength{\tabcolsep}{3pt}
    \renewcommand{\arraystretch}{1.2}
    \begin{tabular}{@{}c*{4}{c}@{\hspace{0.5em}}c*{4}{c}@{}}
        \toprule
        Test set & \multicolumn{2}{c}{ISO} & \multicolumn{2}{c}{ANISO} & Test set & \multicolumn{2}{c}{ISO} & \multicolumn{2}{c}{ANISO} \\
        Condition 1 & MAE & RMSE & MAE & RMSE & Condition 2 & MAE & RMSE & MAE & RMSE \\
        \midrule
        XB1-1 & 0.146 & 0.189 & \textbf{0.129} & \textbf{0.161} & XB2-2 & 0.109 & 0.112 & \textbf{0.081} & \textbf{0.101} \\
        XB1-2 & 0.228 & 0.255 & \textbf{0.163} & \textbf{0.201} & XB2-4 & \textbf{0.157} & \textbf{0.170} & 0.168 & 0.190 \\
        XB1-3 & 0.197 & 0.221 & \textbf{0.167} & \textbf{0.197} & XB2-5 & 0.181 & 0.216 & \textbf{0.160} & \textbf{0.199} \\
        XB1-5 & 0.161 & 0.214 & \textbf{0.145} & \textbf{0.201} &  &  &  &  &  \\
        \midrule
        \(\mu\) & 0.183 & 0.220 & \textbf{0.151} & \textbf{0.190} & \(\mu\) & 0.149 & 0.166 & \textbf{0.136} & \textbf{0.163} \\
        \bottomrule
    \end{tabular}
\end{table}

A comparison of ISO and ANISO on the FB and XB test bearings is presented in Tables~\ref{tab:16} and ~\ref{tab:17}, respectively. In general, ANISO consistently achieved lower error rates. Under FB Condition 1, the MAE and RMSE decreased by 0.034 and 0.036. The most significant improvement was observed for FB1-4, where ANISO achieved an MAE of 0.037 and RMSE of 0.054, compared with 0.147 and 0.168 for ISO. Under FB Conditions 2 and 3, the ANISO consistently demonstrated lower errors for FB2-3, FB2-4, FB2-5, and FB2-7. By contrast, ISO outperformed ANISO on FB2-6 and FB3-3. For the XB dataset, ANISO achieved reductions in both the mean errors under Condition 1 and 2. ISO demonstrated superior performance on XB2-4. Collectively, these findings highlight the empirical advantages of the anisotropic kernel design under the evaluated conditions, although certain individual bearings presented exceptions.

\begin{figure}[H]
	\centering
	\includegraphics[width=0.7\linewidth]{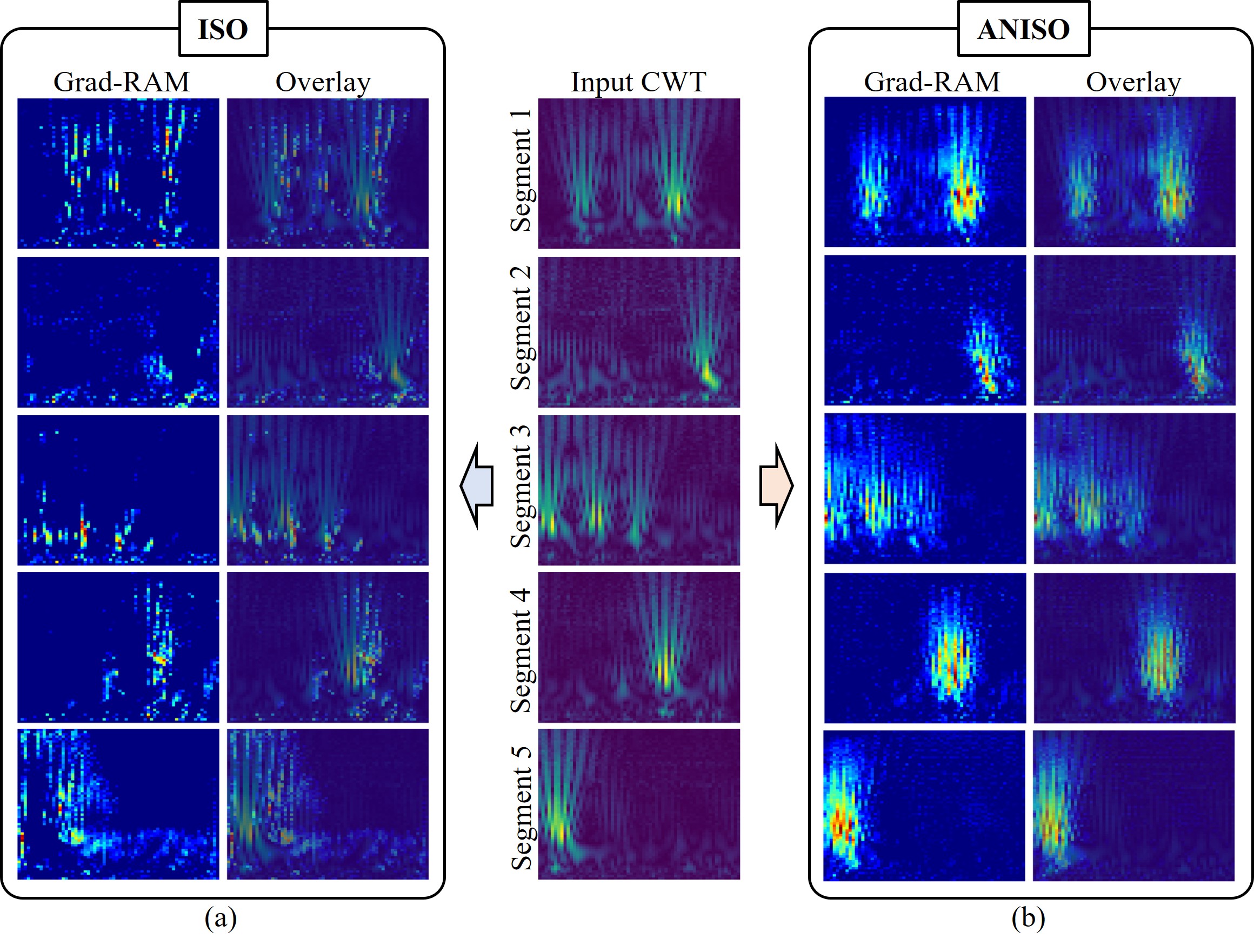}
	\caption{Input CWT scalograms (5 segments) and (a) ISO (b) ANISO Grad-RAM maps for FB1-4 at the end of its RtF sequence.}
	\label{fig:figure12}
\end{figure}

A comparison of the input CWT scalograms and corresponding gradient-weighted regression activation map (Grad-RAM) \cite{Wang_2023_TFReg} maps for (a) ISO and (b) ANISO using five segments extracted from FB1-4 at the end of its RtF sequence is shown in Figure~\ref{fig:figure12}. The horizontal and vertical axes represent the local time and frequency bins, respectively. Each example included an input scalogram, a Grad-RAM map, and an overlay. Grad-RAM was computed using feature maps from the first block of the FAAC feature extractor to analyze local time--frequency attribution patterns at an early stage of convolutional feature extraction.

In both panels, the Grad-RAM responses partially coincide with high-intensity regions in the input CWT scalograms. The ISO maps in panel (a) appear to be more spatially diffuse, with activations extending beyond the primary regions of interest. The ANISO maps in panel (b) are more concentrated within specific time--frequency regions and demonstrate sharper contrast with their surroundings. These observations indicate that the two kernel designs yield distinct spatial distributions of model attribution.

\section{Conclusion}
\label{sec:conclus}

This study introduced FAAC-GRU, a novel approach for RUL prediction from TFRs with directional structures. The proposed model incorporated an explicit anisotropic inductive bias through anisotropic convolutions, DCBAM-based channel attention, and DAP-based adaptive feature aggregation, enabling the extraction of localized time--frequency patterns. The GRU subsequently modeled temporal dependencies within these representations to facilitate accurate RUL estimation. During inference, MC dropout was employed to generate prediction intervals for the RUL estimates. Experimental evaluations on two publicly available bearing datasets demonstrate that FAAC-GRU achieved lower reported mean prediction errors compared with baseline models across various operating conditions. Ablation studies highlight the significant contributions of the anisotropic kernel design and DAP, particularly on the FB dataset, although performance gains were not uniform across all individual bearings. Attention visualization revealed input-dependent weighting of feature-map positions, whereas kernel comparisons empirically validate the capacity of the model for directional feature extraction. The inclusion of DCBAM resulted in marginal improvements in mean performance, indicating a more limited impact in the current experimental setting. In terms of computational efficiency, FAAC-GRU incurs higher FLOPs and longer inference times relative to comparison models, although it maintained millisecond-scale inference on the tested GPU. For deployment in resource-constrained environments, model compression methods, such as quantization or pruning, may be advantageous. Comprehensive assessment of end-to-end real-time performance under target deployment conditions remains an area for future investigation. A further limitation of this work is the exclusive use of TFRs as input representations. The magnitude-based representation and subsequent resizing used in this study may discard phase information and fine-scale signal details. Although vibration inputs preserve the recorded waveform, they also introduce noise and variability linked to operating conditions, potentially affecting prediction robustness. Future research will explore multi-domain fusion architectures that integrate complementary representations while mitigating overreliance on any single domain.

\bibliographystyle{unsrt}
\bibliography{references}

@article{An_2015_Prac, title={Practical options for selecting data-driven or physics-based prognostics algorithms with reviews}, volume={133}, ISSN={0951-8320}, url={http://dx.doi.org/10.1016/j.ress.2014.09.014}, DOI={10.1016/j.ress.2014.09.014}, journal={Reliability Engineering \& System Safety}, publisher={Elsevier BV}, author={An, Dawn and Kim, Nam H. and Choi, Joo-Ho}, year={2015}, month=Jan, pages={223--236} }

@article{Ayman_2025_Feat, title={Feature learning for bearing prognostics: A comprehensive review of machine/deep learning methods, challenges, and opportunities}, volume={245}, ISSN={0263-2241}, url={http://dx.doi.org/10.1016/j.measurement.2024.116589}, DOI={10.1016/j.measurement.2024.116589}, journal={Measurement}, publisher={Elsevier BV}, author={Ayman, Ahmed and Onsy, Ahmed and Attallah, Omneya and Brooks, Hadley and Morsi, Iman}, year={2025}, month=Mar, pages={116589} }

@article{Chen_2025_Hybrid, title={Remaining useful life prediction using a hybrid transfer learning-based adaptive Wiener process model}, volume={260}, ISSN={0951-8320}, url={http://dx.doi.org/10.1016/j.ress.2025.110975}, DOI={10.1016/j.ress.2025.110975}, journal={Reliability Engineering \& System Safety}, publisher={Elsevier BV}, author={Chen, Xiaowu and Liu, Zhen and Wu, Kunping and Sheng, Hanmin and Cheng, Yuhua}, year={2025}, month=Aug, pages={110975} }

@article{Cubillo_2016_Review, title={A review of physics-based models in prognostics: Application to gears and bearings of rotating machinery}, volume={8}, ISSN={1687-8140}, url={http://dx.doi.org/10.1177/1687814016664660}, DOI={10.1177/1687814016664660}, number={8}, journal={Advances in Mechanical Engineering}, publisher={SAGE Publications}, author={Cubillo, Adrian and Perinpanayagam, Suresh and Esperon-Miguez, Manuel}, year={2016}, month=Aug }

@article{Deng_2024_Hybrid, title={An intelligent hybrid deep learning model for rolling bearing remaining useful life prediction}, volume={40}, ISSN={1477-2671}, url={http://dx.doi.org/10.1080/10589759.2024.2385074}, DOI={10.1080/10589759.2024.2385074}, number={6}, journal={Nondestructive Testing and Evaluation}, publisher={Informa UK Limited}, author={Deng, Linfeng and Li, Wei and Yan, Xinhui}, year={2024}, month=jul, pages={2670--2697} }

@article{Feng_2013_Recent, title={Recent advances in time–frequency analysis methods for machinery fault diagnosis: A review with application examples}, volume={38}, ISSN={0888-3270}, url={http://dx.doi.org/10.1016/j.ymssp.2013.01.017}, DOI={10.1016/j.ymssp.2013.01.017}, number={1}, journal={Mechanical Systems and Signal Processing}, publisher={Elsevier BV}, author={Feng, Zhipeng and Liang, Ming and Chu, Fulei}, year={2013}, month=jul, pages={165--205} }

@article{Fink_2020_Poten, title={Potential, challenges and future directions for deep learning in prognostics and health management applications}, volume={92}, ISSN={0952-1976}, url={http://dx.doi.org/10.1016/j.engappai.2020.103678}, DOI={10.1016/j.engappai.2020.103678}, journal={Engineering Applications of Artificial Intelligence}, publisher={Elsevier BV}, author={Fink, Olga and Wang, Qin and Svensén, Markus and Dersin, Pierre and Lee, Wan-Jui and Ducoffe, Melanie}, year={2020}, month=jun, pages={103678} }

@article{Gabrielli_2024_Phys, title={Physics-based prognostics of rolling-element bearings: The equivalent damaged volume algorithm}, volume={215}, ISSN={0888-3270}, url={http://dx.doi.org/10.1016/j.ymssp.2024.111435}, DOI={10.1016/j.ymssp.2024.111435}, journal={Mechanical Systems and Signal Processing}, publisher={Elsevier BV}, author={Gabrielli, Alberto and Battarra, Mattia and Mucchi, Emiliano and Dalpiaz, Giorgio}, year={2024}, month=jun, pages={111435} }

@inproceedings{Gal2016MCdrop, title={Dropout as a bayesian approximation: Representing model uncertainty in deep learning}, booktitle={international conference on machine learning}, publisher={PMLR}, author={Gal, Yarin and Ghahramani, Zoubin}, year={2016}, pages={1050--1059} }

@inproceedings{Gazizulin_2015_Towards, title={Towards a physics based prognostic model for bearing - Spall initiation and propagation}, url={http://dx.doi.org/10.1109/aero.2015.7118995}, DOI={10.1109/aero.2015.7118995}, booktitle={2015 IEEE Aerospace Conference}, publisher={IEEE}, author={Gazizulin, Dmitri and Kogan, Gideon and Klein, Renata and Bortman, Jacob}, year={2015}, month=Mar, pages={1--10} }

@article{He_2025_Phys, title={Physics-informed neural network supported wiener process for degradation modeling and reliability prediction}, volume={258}, ISSN={0951-8320}, url={http://dx.doi.org/10.1016/j.ress.2025.110906}, DOI={10.1016/j.ress.2025.110906}, journal={Reliability Engineering \& System Safety}, publisher={Elsevier BV}, author={He, Zhongze and Wang, Shaoping and Shi, Jian and Liu, Di and Duan, Xiaochuan and Shang, Yaoxing}, year={2025}, month=jun, pages={110906} }

@article{Jiang_2023_Dual, title={A new convolutional dual-channel Transformer network with time window concatenation for remaining useful life prediction of rolling bearings}, volume={56}, ISSN={1474-0346}, url={http://dx.doi.org/10.1016/j.aei.2023.101966}, DOI={10.1016/j.aei.2023.101966}, journal={Advanced Engineering Informatics}, publisher={Elsevier BV}, author={Jiang, Li and Zhang, Tianao and Lei, Wei and Zhuang, Kejia and Li, Yibing}, year={2023}, month=Apr, pages={101966} }

@article{Li_2024_Review, title={A review on physics-informed data-driven remaining useful life prediction: Challenges and opportunities}, volume={209}, ISSN={0888-3270}, url={http://dx.doi.org/10.1016/j.ymssp.2024.111120}, DOI={10.1016/j.ymssp.2024.111120}, journal={Mechanical Systems and Signal Processing}, publisher={Elsevier BV}, author={Li, Huiqin and Zhang, Zhengxin and Li, Tianmei and Si, Xiaosheng}, year={2024}, month=Mar, pages={111120} }

@article{Li_2018_DCNN, title={Remaining useful life estimation in prognostics using deep convolution neural networks}, volume={172}, ISSN={0951-8320}, url={http://dx.doi.org/10.1016/j.ress.2017.11.021}, DOI={10.1016/j.ress.2017.11.021}, journal={Reliability Engineering \& System Safety}, publisher={Elsevier BV}, author={Li, Xiang and Ding, Qian and Sun, Jian-Qiao}, year={2018}, month=Apr, pages={1--11} }

@misc{Liu_2022_SALCNN, author={Liu, Bingguo and Gao, Zhuo and Lu, Binghui and Dong, Hangcheng and An, Zeru}, title={Sal-cnn: Estimate the remaining useful life of bearings using time-frequency information}, howpublished = {arXiv preprint arXiv:2204.05045}, year={2022}, archivePrefix = {arXiv}, eprint={2204.05045} }

@article{Lv_2024_Hybrid, title={A hybrid method combining Lévy process and neural network for predicting remaining useful life of rotating machinery}, volume={61}, ISSN={1474-0346}, url={http://dx.doi.org/10.1016/j.aei.2024.102490}, DOI={10.1016/j.aei.2024.102490}, journal={Advanced Engineering Informatics}, publisher={Elsevier BV}, author={Lv, Shuai and Liu, Shujie and Li, Hongkun and Wang, Yu and Liu, Gengshuo and Dai, Wei}, year={2024}, month=Aug, pages={102490} }

@article{Ma_2020_DeepW, title={Deep wavelet sequence-based gated recurrent units for the prognosis of rotating machinery}, volume={20}, ISSN={1741-3168}, url={http://dx.doi.org/10.1177/1475921720933155}, DOI={10.1177/1475921720933155}, number={4}, journal={Structural Health Monitoring}, publisher={SAGE Publications}, author={Ma, Meng and Mao, Zhu}, year={2020}, month=jul, pages={1794--1804} }

@article{Nandi_2005_Condi, title={Condition Monitoring and Fault Diagnosis of Electrical Motors---A Review}, volume={20}, ISSN={0885-8969}, url={http://dx.doi.org/10.1109/tec.2005.847955}, DOI={10.1109/tec.2005.847955}, number={4}, journal={IEEE Transactions on Energy Conversion}, publisher={Institute of Electrical and Electronics Engineers (IEEE)}, author={Nandi, S. and Toliyat, H.A. and Li, X.}, year={2005}, month=Dec, pages={719--729} }

@inproceedings{Nectoux_2012_PRONOSTIA, author={Nectoux, Patrick and Gouriveau, Rafael and Medjaher, Kamal and Ramasso, Emmanuel and Chebel-Morello, Brigitte and Zerhouni, Noureddine and Varnier, Christophe}, title={PRONOSTIA: An experimental platform for bearings accelerated degradation tests}, booktitle={IEEE International Conference on Prognostics and Health Management, PHM'12.}, year={2012}, pages={1--8}, note={IEEE Catalog Number: CPF12PHM-CDR} }

@article{Niazi_2024_Multi, title={Multi-scale time series analysis using TT-ConvLSTM technique for bearing remaining useful life prediction}, volume={206}, ISSN={0888-3270}, url={http://dx.doi.org/10.1016/j.ymssp.2023.110888}, DOI={10.1016/j.ymssp.2023.110888}, journal={Mechanical Systems and Signal Processing}, publisher={Elsevier BV}, author={Niazi, Sajawal Gul and Huang, Tudi and Zhou, Hongming and Bai, Song and Huang, Hong-Zhong}, year={2024}, month=Jan, pages={110888} }

@article{QIU_2002_DAMAGE, title={DAMAGE MECHANICS APPROACH FOR BEARING LIFETIME PROGNOSTICS}, volume={16}, ISSN={0888-3270}, url={http://dx.doi.org/10.1006/mssp.2002.1483}, DOI={10.1006/mssp.2002.1483}, number={5}, journal={Mechanical Systems and Signal Processing}, publisher={Elsevier BV}, author={QIU, JING and SETH, BRIJ B. and LIANG, STEVEN Y. and ZHANG, CHENG}, year={2002}, month=sep, pages={817--829} }

@article{Rejith_2023_Bearings, title={Bearings for aerospace applications}, volume={181}, ISSN={0301-679X}, url={http://dx.doi.org/10.1016/j.triboint.2023.108312}, DOI={10.1016/j.triboint.2023.108312}, journal={Tribology International}, publisher={Elsevier BV}, author={Rejith, R and Kesavan, D. and Chakravarthy, P and Narayana Murty, S.V.S.}, year={2023}, month=Mar, pages={108312} }

@article{Rezakhaniha_2011_Experi, title={Experimental investigation of collagen waviness and orientation in the arterial adventitia using confocal laser scanning microscopy}, volume={11}, ISSN={1617-7940}, url={http://dx.doi.org/10.1007/s10237-011-0325-z}, DOI={10.1007/s10237-011-0325-z}, number={3-4}, journal={Biomechanics and Modeling in Mechanobiology}, publisher={Springer Science and Business Media LLC}, author={Rezakhaniha, R. and Agianniotis, A. and Schrauwen, J. T. C. and Griffa, A. and Sage, D. and Bouten, C. V. C. and van de Vosse, F. N. and Unser, M. and Stergiopulos, N.}, year={2011}, month=jul, pages={461--473} }

@article{Shi_2025_Dynamic, title={Dynamic Energy Sparse Self-Attention Based on Informer for Remaining Useful Life of Rolling Bearings}, volume={13}, ISSN={2169-3536}, url={http://dx.doi.org/10.1109/access.2025.3594077}, DOI={10.1109/access.2025.3594077}, journal={IEEE Access}, publisher={Institute of Electrical and Electronics Engineers (IEEE)}, author={Shi, Cheng and Li, Qifei and Chen, Hui and Song, Yuanwei and Le, Qianqi}, year={2025}, pages={139616--139630} }

@article{Tefera_2025_Cons, title={Constraint-Guided Learning of Data-driven Health Indicator Models: An Application on Bearings}, volume={16}, ISSN={2153-2648}, url={http://dx.doi.org/10.36001/ijphm.2025.v16i2.4268}, DOI={10.36001/ijphm.2025.v16i2.4268}, number={2}, journal={International Journal of Prognostics and Health Management}, publisher={PHM Society}, author={Tefera, Yonas and Van Baelen, Quinten and Meire, Maarten and Luca, Stijn and Karsmakers, Peter}, year={2025}, month=Aug }

@article{Wang_2020_Hybrid, title={A Hybrid Prognostics Approach for Estimating Remaining Useful Life of Rolling Element Bearings}, volume={69}, ISSN={1558-1721}, url={http://dx.doi.org/10.1109/tr.2018.2882682}, DOI={10.1109/tr.2018.2882682}, number={1}, journal={IEEE Transactions on Reliability}, publisher={Institute of Electrical and Electronics Engineers (IEEE)}, author={Wang, Biao and Lei, Yaguo and Li, Naipeng and Li, Ningbo}, year={2020}, month=Mar, pages={401--412} }

@article{Wang_2019_Deepse, title={Deep separable convolutional network for remaining useful life prediction of machinery}, volume={134}, ISSN={0888-3270}, url={http://dx.doi.org/10.1016/j.ymssp.2019.106330}, DOI={10.1016/j.ymssp.2019.106330}, journal={Mechanical Systems and Signal Processing}, publisher={Elsevier BV}, author={Wang, Biao and Lei, Yaguo and Li, Naipeng and Yan, Tao}, year={2019}, month=Dec, pages={106330} }

@article{Wang_2023_TFReg, title={TFRegNCI: Interpretable Noncovalent Interaction Correction Multimodal Based on Transformer Encoder Fusion}, volume={63}, ISSN={1549-960X}, url={http://dx.doi.org/10.1021/acs.jcim.2c01283}, DOI={10.1021/acs.jcim.2c01283}, number={3}, journal={Journal of Chemical Information and Modeling}, publisher={American Chemical Society (ACS)}, author={Wang, Donghan and Li, Wenze and Dong, Xu and Li, Hongzhi and Hu, LiHong}, year={2023}, month=Jan, pages={782--793} }

@article{Weichsel_2009_Quanti, title={A quantitative measure for alterations in the actin cytoskeleton investigated with automated high‐throughput microscopy}, volume={77A}, ISSN={1552-4930}, url={http://dx.doi.org/10.1002/cyto.a.20818}, DOI={10.1002/cyto.a.20818}, number={1}, journal={Cytometry Part A}, publisher={Wiley}, author={Weichsel, Julian and Herold, Nikolas and Lehmann, Maik J. and Kräusslich, Hans‐Georg and Schwarz, Ulrich S.}, year={2009}, month=Nov, pages={52--63} }

@inbook{Woo_2018_CBAM, title={CBAM: Convolutional Block Attention Module}, ISBN={9783030012342}, ISSN={1611-3349}, url={http://dx.doi.org/10.1007/978-3-030-01234-2_1}, DOI={10.1007/978-3-030-01234-2_1}, booktitle={Computer Vision – ECCV 2018}, publisher={Springer International Publishing}, author={Woo, Sanghyun and Park, Jongchan and Lee, Joon-Young and Kweon, In So}, year={2018}, pages={3--19} }

@article{Xiao_2022_Self, title={Self-attention-based adaptive remaining useful life prediction for IGBT with Monte Carlo dropout}, volume={239}, ISSN={0950-7051}, url={http://dx.doi.org/10.1016/j.knosys.2021.107902}, DOI={10.1016/j.knosys.2021.107902}, journal={Knowledge-Based Systems}, publisher={Elsevier BV}, author={Xiao, Dengyu and Qin, Chengjin and Ge, Jianwen and Xia, Pengcheng and Huang, Yixiang and Liu, Chengliang}, year={2022}, month=Mar, pages={107902} }

@article{Yoo_2018_Novel, title={A Novel Image Feature for the Remaining Useful Lifetime Prediction of Bearings Based on Continuous Wavelet Transform and Convolutional Neural Network}, volume={8}, ISSN={2076-3417}, url={http://dx.doi.org/10.3390/app8071102}, DOI={10.3390/app8071102}, number={7}, journal={Applied Sciences}, publisher={MDPI AG}, author={Yoo, Youngji and Baek, Jun-Geol}, year={2018}, month=jul, pages={1102} }

@article{Zhang_2023_Digital, title={Digital twin-driven partial domain adaptation network for intelligent fault diagnosis of rolling bearing}, volume={234}, ISSN={0951-8320}, url={http://dx.doi.org/10.1016/j.ress.2023.109186}, DOI={10.1016/j.ress.2023.109186}, journal={Reliability Engineering \& System Safety}, publisher={Elsevier BV}, author={Zhang, Yongchao and Ji, J.C. and Ren, Zhaohui and Ni, Qing and Gu, Fengshou and Feng, Ke and Yu, Kun and Ge, Jian and Lei, Zihao and Liu, Zheng}, year={2023}, month=jun, pages={109186} }

@article{Zhao_2021_Feature, title={Feature Extraction for Data-Driven Remaining Useful Life Prediction of Rolling Bearings}, volume={70}, ISSN={1557-9662}, url={http://dx.doi.org/10.1109/tim.2021.3059500}, DOI={10.1109/tim.2021.3059500}, journal={IEEE Transactions on Instrumentation and Measurement}, publisher={Institute of Electrical and Electronics Engineers (IEEE)}, author={Zhao, Huimin and Liu, Haodong and Jin, Yang and Dang, Xiangjun and Deng, Wu}, year={2021}, pages={1--10} }

@article{Zhong_2025_RULMulti, title={Remaining useful life prediction of rolling bearing based on multi-region hypergraph self-attention network}, volume={225}, ISSN={0888-3270}, url={http://dx.doi.org/10.1016/j.ymssp.2025.112331}, DOI={10.1016/j.ymssp.2025.112331}, journal={Mechanical Systems and Signal Processing}, publisher={Elsevier BV}, author={Zhong, Jianhua and Jiang, Haifeng and Gu, Kairong and Zhong, Jianfeng and Zhong, Shuncong}, year={2025}, month=Feb, pages={112331} }

@article{Zuo_2023_Hybrid, title={A hybrid attention-based multi-wavelet coefficient fusion method in RUL prognosis of rolling bearings}, volume={237}, ISSN={0951-8320}, url={http://dx.doi.org/10.1016/j.ress.2023.109337}, DOI={10.1016/j.ress.2023.109337}, journal={Reliability Engineering \& System Safety}, publisher={Elsevier BV}, author={Zuo, Tao and Zhang, Kai and Zheng, Qing and Li, Xianxin and Li, Zhixuan and Ding, Guofu and Zhao, Minghang}, year={2023}, month=sep, pages={109337} }

@article{Gao_2024_Long, title={Long-term temporal attention neural network with adaptive stage division for remaining useful life prediction of rolling bearings}, volume={251}, ISSN={0951-8320}, url={http://dx.doi.org/10.1016/j.ress.2024.110218}, DOI={10.1016/j.ress.2024.110218}, journal={Reliability Engineering \& System Safety}, publisher={Elsevier BV}, author={Gao, Pengjie and Wang, Junliang and Shi, Ziqi and Ming, Weiwei and Chen, Ming}, year={2024}, month=Nov, pages={110218} }

@article{Hou_2022_High, title={High-speed train wheel set bearing fault diagnosis and prognostics: Fingerprint feature recognition method based on acoustic emission}, volume={171}, ISSN={0888-3270}, url={http://dx.doi.org/10.1016/j.ymssp.2022.108947}, DOI={10.1016/j.ymssp.2022.108947}, journal={Mechanical Systems and Signal Processing}, publisher={Elsevier BV}, author={Hou, Dongming and Qi, Hongyuan and Wang, Cuiping and Han, Defu}, year={2022}, month=May, pages={108947} }

@article{Mallat_1992_Singular, title={Singularity detection and processing with wavelets}, volume={38}, ISSN={1557-9654}, url={http://dx.doi.org/10.1109/18.119727}, DOI={10.1109/18.119727}, number={2}, journal={IEEE Transactions on Information Theory}, publisher={Institute of Electrical and Electronics Engineers (IEEE)}, author={Mallat, S. and Hwang, W.L.}, year={1992}, month=Mar, pages={617--643} }

@article{Wei_2025_RUL, title={Remaining Useful Life Prediction Method for Bearings Based on Pruned Exact Linear Time State Segmentation and Time--Frequency Diagram}, volume={25}, ISSN={1424-8220}, url={http://dx.doi.org/10.3390/s25061950}, DOI={10.3390/s25061950}, number={6}, journal={Sensors}, publisher={MDPI AG}, author={Wei, Xu and Fan, Jingjing and Wang, Huahua and Cai, Lulu}, year={2025}, month=Mar, pages={1950} }

@article{Ye_2023_Selective, title={A Selective Adversarial Adaptation Network for Remaining Useful Life Prediction of Machines Under Different Working Conditions}, volume={17}, ISSN={2373-7816}, url={http://dx.doi.org/10.1109/jsyst.2022.3183134}, DOI={10.1109/jsyst.2022.3183134}, number={1}, journal={IEEE Systems Journal}, publisher={Institute of Electrical and Electronics Engineers (IEEE)}, author={Ye, Zhuang and Yu, Jianbo}, year={2023}, month=Mar, pages={62--71} }

\section*{Appendix}
\appendix

\setcounter{figure}{0}
\renewcommand{\thefigure}{A\arabic{figure}}

\section{Theoretical properties of adaptive aggregation and directional coherence}
\label{sec:appA}

\subsection{Non-injectivity of GAP and input-adaptive weighting of DAP}
\label{sec:appA1}

Let \(\mathbf{A} \in \mathbb{R}^{D \times H^{'} \times W^{'}}\) denote the input feature map to the DAP module, where \(D\) denotes the number of channels and \(H^{'}\) and \(W^{'}\) represent the sizes of the frequency and local time axes, respectively. Let \(\Lambda = \left\{ 1,\ldots,H^{'} \right\} \times \left\{ 1,\ldots,W^{'} \right\}\) denote the set of spatial locations and \(N = H^{'}W^{'}\) its cardinality. GAP assigns equal weight \(\frac{1}{N}\) to each spatial location, yielding the following output for \(\mathbf{A}\) in channel \(d\):

\begin{equation}
    \left\lbrack z_{GAP}\left( \mathbf{A} \right) \right\rbrack_{d} = \frac{1}{N}\sum_{h = 1}^{H^{'}}{\sum_{w = 1}^{W^{'}}A_{d,h,w}},\ \ d = 1,\ldots,D
    \tag{A1.1}
    \label{eq:A1.1}
\end{equation}

To demonstrate the inability of GAP to distinguish between certain spatial activation distributions, assume \(N > 1\) and let \(\mathcal{S \subset}\Lambda\) be a local region containing \(m\) locations, with \(\rho = \frac{m}{N}\) and \(1 \leq m \leq N\). Let \(s > 0\) denote the activation magnitude. Therefore, the local region occupies only a fraction of the entire spatial domain; the following two feature maps defined below are distinct.

We define a spatially concentrated feature map \(\mathbf{A}^{\left( \mathcal{S} \right)}\), where activations are confined to region \(\mathcal{S}\) in every channel:

\begin{equation}
    A_{d,h,w}^{\left( \mathcal{S} \right)} = \left\{ \begin{array}{r}
            s,\ \ (h,w)\mathcal{\in S} \\
            0,\ \ (h,w)\mathcal{\notin S}
        \end{array} \right.\ ,\ \ d = 1,\ldots,D
    \tag{A1.2}
    \label{eq:A1.2}
\end{equation}

We define a spatially diffuse feature map \(\mathbf{A}^{(diff)}\) by distributing the same total activation uniformly across the entire spatial domain:

\begin{equation}
    A_{d,h,w}^{(diff)} = \rho s,\ \ \forall(h,w) \in \Lambda,\ \ d = 1,\ldots,D
    \tag{A1.3}
    \label{eq:A1.3}
\end{equation}

where \(\rho\) denotes the fraction of spatial locations occupied by \(\mathcal{S}\). By construction, both feature maps have identical sums of spatial activations in each channel:

\begin{equation}
    \sum_{h = 1}^{H^{'}}{\sum_{w = 1}^{W^{'}}A_{d,h,w}^{\left( \mathcal{S} \right)}} = ms = N\rho s = \sum_{h = 1}^{H^{'}}{\sum_{w = 1}^{W^{'}}A_{d,h,w}^{(diff)}}
    \tag{A1.4}
    \label{eq:A1.4}
\end{equation}

Consequently, for every channel \(d\),

\begin{equation}
    \left\lbrack z_{GAP}\left( \mathbf{A}^{\left( \mathcal{S} \right)} \right) \right\rbrack_{d} = \rho s = \left\lbrack z_{GAP}\left( \mathbf{A}^{(diff)} \right) \right\rbrack_{d}
    \tag{A1.5}
    \label{eq:A1.5}
\end{equation}

This demonstrates that two feature maps with different spatial activation distributions can yield identical GAP outputs. GAP retains only the spatial mean of each channel and is insensitive to differences in spatial concentrations when the mean values are identical. Similarly, consider two distinct regions \(\mathcal{S}_{1}\) and \(\mathcal{S}_{2}\), with equal cardinality and activation magnitudes. Then,

\begin{equation}
    \left\lbrack z_{GAP}\left( \mathbf{A}^{\left( \mathcal{S}_{1} \right)} \right) \right\rbrack_{d} = \left\lbrack z_{GAP}\left( \mathbf{A}^{\left( \mathcal{S}_{2} \right)} \right) \right\rbrack_{d},\ \ if\ \left| \mathcal{S}_{1} \right| = \left| \mathcal{S}_{2} \right|
    \tag{A1.6}
    \label{eq:A1.6}
\end{equation}

This indicates that GAP does not encode the spatial locations of activations.

To compare DAP with GAP, we considered the time-to-frequency path defined in the main text. This path first aggregates along the time dimension using the time-axis softmax weights \(\mathbf{w}_{w}^{T}\left( \mathbf{A} \right)\):

\begin{equation}
    \mathbf{u}_{h}^{T \rightarrow F} = \sum_{w = 1}^{W^{'}}{\mathbf{w}_{w}^{T}\left( \mathbf{A} \right)A_{:,h,w}}
    \tag{A1.7}
    \label{eq:A1.7}
\end{equation}

It then aggregates along the frequency dimension using frequency-axis softmax weights \(\mathbf{v}_{h}^{F|T}\left( \mathbf{A} \right)\), computed from the first aggregation as follows:

\begin{equation}
    \mathbf{z}_{T \rightarrow F} = \sum_{h = 1}^{H^{'}}{\mathbf{v}_{h}^{F|T}\left( \mathbf{A} \right)\mathbf{u}_{h}^{T \rightarrow F}}
    \tag{A1.8}
    \label{eq:A1.8}
\end{equation}

The superscript \(F|T\) denotes the frequency axis weights computed after aggregation along the time axis. Substituting Equation~\eqref{eq:A1.7} into Equation~\eqref{eq:A1.8} yields

\begin{equation}
    \mathbf{z}_{T \rightarrow F}\left( \mathbf{A} \right) = \sum_{h = 1}^{H^{'}}{\mathbf{v}_{h}^{F|T}\left( \mathbf{A} \right)\sum_{w = 1}^{W^{'}}{\mathbf{w}_{w}^{T}\left( \mathbf{A} \right)A_{:,h,w}}}
    \tag{A1.9}
    \label{eq:A1.9}
\end{equation}

\begin{equation}
    = \sum_{h = 1}^{H^{'}}{\sum_{w = 1}^{W^{'}}{\mathbf{v}_{h}^{F|T}\left( \mathbf{A} \right)\mathbf{w}_{w}^{T}\left( \mathbf{A} \right)A_{:,h,w}}}
    \tag{A1.10}
    \label{eq:A1.10}
\end{equation}

The effective spatial weight applied by the time-to-frequency path at location \((h,w)\) is:

\[\pi_{h,w}^{T \rightarrow F}\left( \mathcal{A} \right) = \mathbf{v}_{h}^{F|T}\left( \mathbf{A} \right)\mathbf{w}_{w}^{T}\left( \mathbf{A} \right)\]

Because the weights along each axis are normalized via softmax, the resulting spatial weights are nonnegative and sum to one. For a fixed input, the effective weight matrix is given by the outer product of the two axis-specific weight vectors, making it separable.

The total weight assigned by the time-to-frequency path to region \(\mathcal{S}\) is defined as follows:

\begin{equation}
    M_{T \rightarrow F}\left( \mathcal{S;}\mathbf{A} \right) = \sum_{(h,w)\mathcal{\in S}}^{}{\pi_{h,w}^{T \rightarrow F}\left( \mathbf{A} \right)}
    \tag{A1.11}
    \label{eq:A1.11}
\end{equation}

For the spatially concentrated feature map \(\mathbf{A}^{\left( \mathcal{S} \right)}\), the time-to-frequency path output in channel \(d\) is expressed as follows:

\begin{equation}
    \begin{gathered}
            \left\lbrack z_{T \rightarrow F}\left( \mathbf{A}^{\left( \mathcal{S} \right)} \right) \right\rbrack_{d} = \sum_{(h,w)\mathcal{\in S}}^{}{\mathbf{\pi}_{h,w}^{T \rightarrow F}\left( \mathbf{A}^{\left( \mathcal{S} \right)} \right)}s \\
            = M_{T \rightarrow F}\left( \mathcal{S;}\mathbf{A}^{\left( \mathcal{S} \right)} \right)s
    \end{gathered}
    \tag{A1.12}
    \label{eq:A1.12}
\end{equation}

The GAP output for the same feature map is

\begin{equation}
    \left\lbrack z_{GAP}\left( \mathbf{A}^{\left( \mathcal{S} \right)} \right) \right\rbrack_{d} = \rho s
    \tag{A1.13}
    \label{eq:A1.13}
\end{equation}

Therefore, the difference between the two outputs is expressed as follows:

\begin{equation}
    \left\lbrack {z_{T \rightarrow F}\left( \mathbf{A}^{\left( \mathcal{S} \right)} \right) - z}_{GAP}\left( \mathbf{A}^{\left( \mathcal{S} \right)} \right) \right\rbrack_{d} = \left\lbrack M_{T \rightarrow F}\left( \mathcal{S;}\mathbf{A}^{\left( \mathcal{S} \right)} \right) - \rho \right\rbrack s
    \tag{A1.14}
    \label{eq:A1.14}
\end{equation}

GAP assigns region \(\mathcal{S}\) a total weight equal to its spatial occupancy fraction, \(\rho = \frac{m}{N}\). Therefore, if \(M_{T \rightarrow F}\left( \mathcal{S;}\mathbf{A}^{\left( \mathcal{S} \right)} \right) > \rho\), the time-to-frequency path assigns a greater total weight to the local region \(\mathcal{S}\) than GAP does. When both axis-specific weight vectors are uniform, \(\mathbf{w}_{w}^{T} = \frac{1}{W^{'}}\) and \(\mathbf{v}_{h}^{F|T} = \frac{1}{H^{'}}\), ensuring that \(\mathbf{\pi}_{h,w}^{T \rightarrow F} = \frac{1}{H^{'}W^{'}} = \frac{1}{N}\) and the path output equals the GAP output. Therefore, GAP applies fixed uniform weights, whereas DAP can adjust its effective spatial weights based on the input, allowing it to emphasize specific local regions. This result characterizes a single aggregation path; it does not imply that the learned weights necessarily emphasize degradation-relevant regions, that DAP is injective, or that spatial information is fully preserved after pooling and subsequent projection.

\subsection{Quantitative analysis of directional structure in TFRs}
\label{sec:appA2}

\subsubsection{Local structure tensor coherency and dominant orientation}
\label{sec:appA2.1}

The local directional structure of a single-channel feature map \(\mathcal{F} \in \mathbb{R}^{F \times T}\) is quantified using a two-dimensional structure tensor. In this representation, the local time coordinate \(t\) is mapped to the horizontal axis, whereas the frequency coordinate \(f\) is assigned to the vertical axis. Assuming unit grid spacing in \(\mathcal{F}\), partial derivatives at interior grid points are approximated using central differences:

\begin{equation}
    \mathcal{F}_{t}(f,t) ≔ \frac{\partial\mathcal{F}(f,t)}{\partial t} \approx \frac{\mathcal{F}(f,t + 1) - \mathcal{F}(f,t - 1)}{2}
    \tag{A2.1}
    \label{eq:A2.1}
\end{equation}

\begin{equation}
    \mathcal{F}_{f}(f,t) ≔ \frac{\partial\mathcal{F}(f,t)}{\partial f} \approx \frac{\mathcal{F}(f + 1,t) - \mathcal{F}(f - 1,t)}{2}
    \tag{A2.2}
    \label{eq:A2.2}
\end{equation}

The outermost rows and columns are omitted, as central differences cannot be computed at these boundaries. The structure tensor for a local window \(\Omega\) is defined as follows \cite{Rezakhaniha_2011_Experi,Weichsel_2009_Quanti}:

\begin{equation}
    \mathbf{g}(f,t) = \left\lbrack \begin{array}{r}
            \mathcal{F}_{t}(f,t) \\
            \mathcal{F}_{f}(f,t)
        \end{array} \right\rbrack
    \tag{A2.3}
    \label{eq:A2.3}
\end{equation}

\begin{equation}
    \mathbf{J}_{\Omega} = \sum_{(f,t) \in \Omega}^{}{\mathbf{g}(f,t)\mathbf{g}^{T}(f,t)} = \begin{bmatrix}
            J_{tt} & J_{tf} \\
            J_{tf} & J_{ff}
    \end{bmatrix}
    \tag{A2.4}
    \label{eq:A2.4}
\end{equation}

where \(\mathbf{J}_{\Omega}\) denotes a symmetric positive semidefinite matrix, as it is constructed from the sum of outer products of the local gradient vectors \(\mathbf{g}(f,t)\). Let \(\lambda_{1} \geq \lambda_{2} \geq 0\) denote the eigenvalues of \(\mathbf{J}_{\Omega}\). For windows with nonzero total gradient energy, the local structure tensor coherency is defined as \cite{Rezakhaniha_2011_Experi}

\begin{equation}
    C_{\Omega} = \frac{\lambda_{1} - \lambda_{2}}{\lambda_{1} + \lambda_{2}} = \frac{\sqrt{\left( J_{tt} - J_{ff} \right)^{2} + 4J_{tf}^{2}}}{J_{tt} + J_{ff}}
    \tag{A2.5}
    \label{eq:A2.5}
\end{equation}

Therefore, \(0 \leq C_{\Omega} \leq 1\). Values of \(C_{\Omega}\) approaching one indicate strongly anisotropic local gradient energy, whereas values near zero correspond to similar gradient energies along the two principal directions.

The eigenvector associated with \(\lambda_{1}\), denoted by \(\mathbf{e}_{1}\), represents the dominant gradient direction---that is, the direction along which \(\mathcal{F}\) demonstrates the most significant variation within the window. The orientation of a local linear structure is orthogonal to this direction and is therefore determined by the eigenvector associated with \(\lambda_{2}\). Let \(\vartheta_{\Omega}\) denote the structural orientation, measured from the positive \(t\)-axis toward the positive \(f\)-axis. For distinct eigenvalues, this orientation is defined modulo a half-turn and can be expressed as follows\footnote{\url{https://github.com/Biomedical-Imaging-Group/OrientationJ/blob/master/src/main/java/orientation/StructureTensor.java}} \cite{Rezakhaniha_2011_Experi}:

\begin{equation}
    \vartheta_{\Omega} = \frac{1}{2}{atan2}\left( - 2J_{tf},J_{ff} - J_{tt} \right)mod\ \pi,\ \ \vartheta_{\Omega} \in \lbrack 0,\pi)
    \tag{A2.6}
    \label{eq:A2.6}
\end{equation}

where \(\vartheta_{\Omega} \approx 0\) or \(\vartheta_{\Omega} \approx \pi\) radians corresponds with structures aligned with the local time axis, whereas \(\vartheta_{\Omega} \approx \frac{\pi}{2}\) radians indicates alignment with the frequency axis. These orientations are defined with respect to the sampled image coordinates.

\subsubsection{Quantitative results for (64, 64) TFRs}
\label{sec:appA2.2}

\begin{figure}[H]
	\centering
	\includegraphics[scale=0.5]{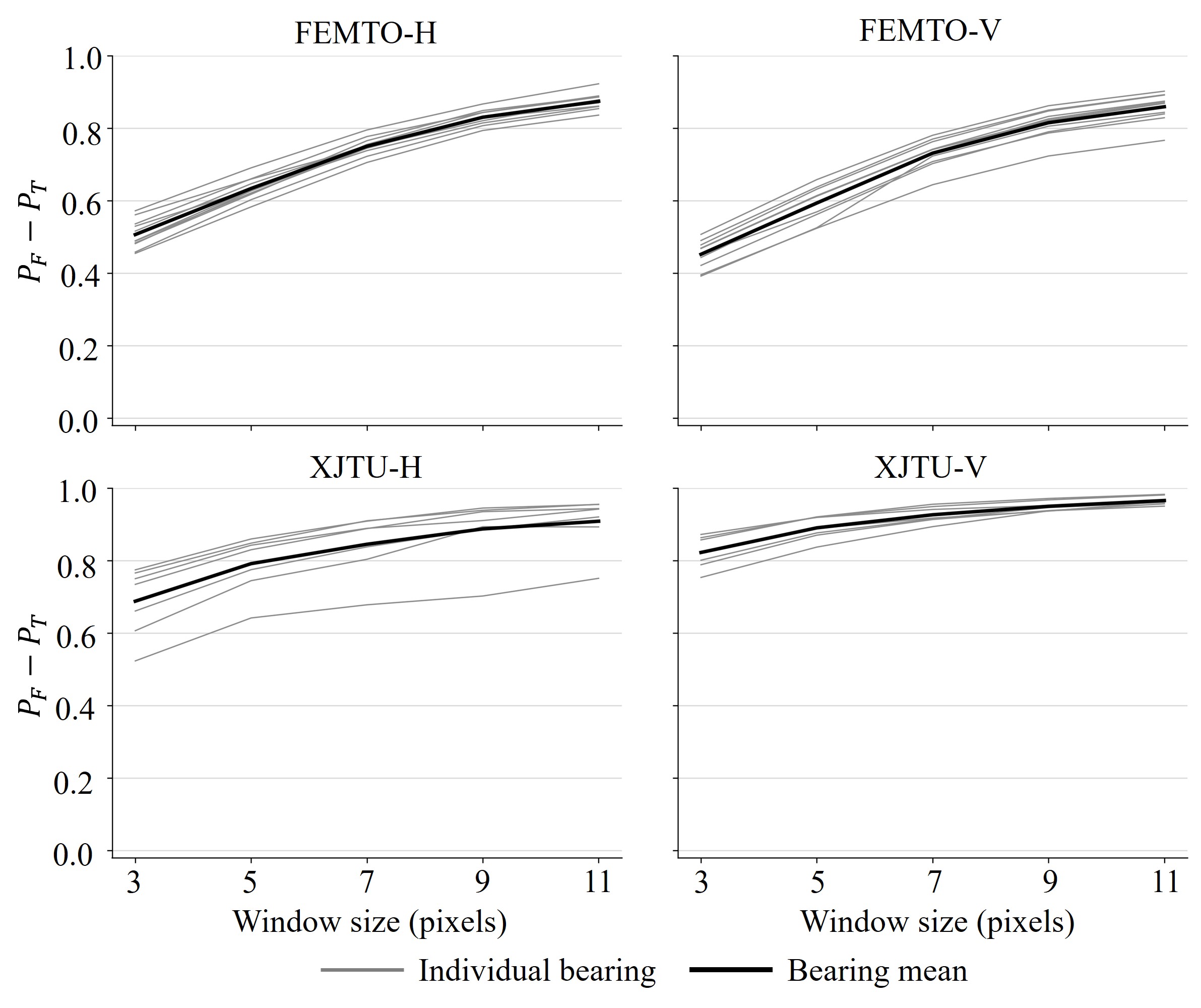}
	\caption{Axis preference of local structures.}
	\label{fig:figurea1}
\end{figure}

The analysis was conducted on (64, 64) CWT scalograms used as model inputs. Following central differencing, the remaining (62, 62) interior grid was partitioned, beginning at the upper-left corner, into non-overlapping local windows \(w \in \left\{ 3,5,7,9,11 \right\}\). Incomplete windows at the boundaries were excluded. Bearing-level measures presented in Figures~\ref{fig:figurea1} and ~\ref{fig:figurea3} were obtained by aggregating results from local windows within each map, averaging across the five segments of each observation, and subsequently averaging across observations within each of the ten intervals of the normalized RtF time.

The analysis utilized test sets from FEMTO and XJTU-SY, comprising 11 FEMTO and seven XJTU-SY bearings. Here, H and V denote the horizontal and vertical vibration measurement channels, respectively. Dataset--channel combinations are denoted as FEMTO-H, FEMTO-V, XJTU-H, and XJTU-V.

A comparison of the alignment of the local structures along the time and frequency axes is shown in Figure~\ref{fig:figurea1}. The anisotropic gradient energy within a local window \(\Omega\) is defined as \(D_{\Omega} = \lambda_{1,\Omega} - \lambda_{2,\Omega}\). The fractions associated with the two axes were determined by the angular distances \(d_{T}\) and \(d_{F}\) from the time and frequency axes, respectively.

\begin{equation}
    \begin{gathered}
            P_{T} = \frac{\sum_{\Omega}^{}{D_{\Omega}\mathbb{I}\left\{ d_{T}\left( \vartheta_{\Omega} \right) \leq 15^{{^\circ}} \right\}}}{\sum_{\Omega}^{}D_{\Omega}} \\
            P_{F} = \frac{\sum_{\Omega}^{}{D_{\Omega}\mathbb{I}\left\{ d_{F}\left( \vartheta_{\Omega} \right) \leq 15^{{^\circ}} \right\}}}{\sum_{\Omega}^{}D_{\Omega}}
    \end{gathered}
    \tag{A2.7}
    \label{eq:A2.7}
\end{equation}

The angular distances to the two axes are defined as follows:

\begin{equation}
    \begin{gathered}
            d_{T}(\vartheta) = \min(\vartheta,\pi - \vartheta) \\
            d_{F}(\vartheta) = \left| \vartheta - \frac{\pi}{2} \right|
    \end{gathered}
    \tag{A2.8}
    \label{eq:A2.8}
\end{equation}

The axis preference index was defined as \(P_{F} - P_{T}\). Therefore, \(P_{F} - P_{T} > 0\) indicates that a greater proportion of anisotropic gradient energy is associated with local structures aligned with the frequency axis, as determined by the specified angular tolerance. The thin gray lines in Figure~\ref{fig:figurea1} represent individual bearings, whereas the thick black line represents the mean.

The mean values of \(P_{F} - P_{T}\) for w = 5 were 0.633, 0.593, 0.791, and 0.890 for FEMTO-H, FEMTO-V, XJTU-H, and XJTU-V, respectively. Furthermore, \(P_{F} - P_{T} > 0\) in all 180 combinations of 18 bearings, two measurement channels, and five window sizes. These findings demonstrate a consistent preference for frequency-axis alignment in the analyzed (64, 64) TFRs across all bearings and window sizes considered.

\begin{figure}[H]
	\centering
	\includegraphics[scale=0.5]{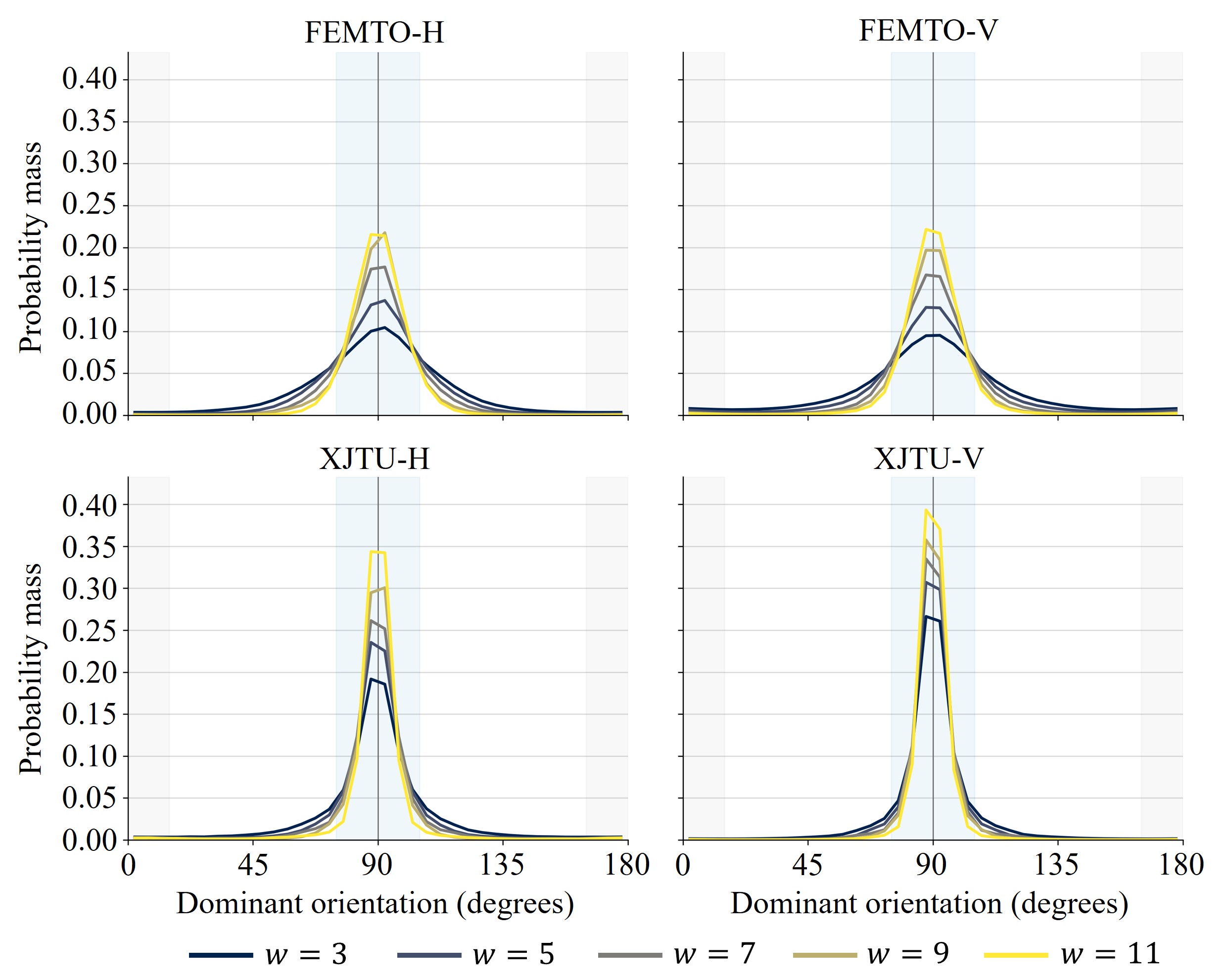}
	\caption{Distribution of dominant orientation.}
	\label{fig:figurea2}
\end{figure}

The distribution of local anisotropic gradient energy over the structural orientations for each window size is shown in Figure~\ref{fig:figurea2}. The number of complete local windows in a single TFR was \(N_{w} = \left\lfloor \frac{62}{w} \right\rfloor^{2}\). Therefore, for \(w \in \left\{ 3,5,7,9,11 \right\}\), the corresponding counts are \(N_{w} \in \left\{ 400,144,64,36,25 \right\}\). The orientation ranges \(\left\lbrack 0^{{^\circ}},180^{{^\circ}} \right)\) were divided into 36 bins with a width of \(5^{{^\circ}}\), denoted as \(\mathcal{B}_{k}\):

\begin{equation}
    \mathcal{B}_{k} = \left\lbrack 5(k - 1)^{{^\circ}},5k^{{^\circ}} \right),\ \ k = 1,\ldots,36
    \tag{A2.9}
    \label{eq:A2.9}
\end{equation}

For bearing \(i\) and TFR map \(m\), the fraction of anisotropic gradient energy assigned to the orientation bin \(\mathcal{B}_{k}\) is calculated as follows:

\begin{equation}
    h_{i,m,w,k} = \frac{\sum_{j = 1}^{N_{w}}{D_{i,m,w,j}\mathbb{I}\left\{ \vartheta_{i,m,w,j} \in \mathcal{B}_{k} \right\}}}{\sum_{j = 1}^{N_{w}}D_{i,m,w,j}}
    \tag{A2.10}
    \label{eq:A2.10}
\end{equation}

where \(h_{i,m,w,k}\) denotes the normalized anisotropic gradient energy assigned to each \(5^{{^\circ}}\) bin and satisfies \(\sum_{k = 1}^{36}h_{i,m,w,k} = 1\) when the total anisotropic gradient energy is nonzero. These values collectively form a discrete energy-weighted orientation distribution.

For each bearing, the distributions were first averaged over all segment--observation TFR maps. The resulting bearing-level distributions were then averaged across all bearings to obtain a distribution for each dataset--channel combination as follows:

\begin{equation}
    {\overline{h}}_{w,k} = \frac{1}{I}\sum_{i = 1}^{I}\left( \frac{1}{M_{i}}\sum_{m = 1}^{M_{i}}h_{i,m,w,k} \right)
    \tag{A2.11}
    \label{eq:A2.11}
\end{equation}

where \(M_{i}\) denotes the total number of segment--observation TFR maps for bearing \(i\), and \(I\) represents the number of bearings. Therefore, Figure~\ref{fig:figurea2} assigns equal weights to each bearing after averaging the map-level distributions.

Across all dataset--channel combinations, the normalized anisotropic gradient energy was predominantly concentrated near \(\ 90^{{^\circ}}\), indicating a preference for structures aligned with the frequency axis, which corresponds to the vertical direction in the TFRs. This observation is consistent with the positive values of $P_F-P_T$ in Figure~\ref{fig:figurea1}.

The difference in the map-level coherency between the original TFRs and pixel-permuted reference maps is shown in Figure~\ref{fig:figurea3}. Map-level coherency is defined as follows:

\begin{equation}
    C_{map} = \frac{\sum_{\Omega}^{}{E_{\Omega}C_{\Omega}}}{\sum_{\Omega}^{}E_{\Omega}} = \frac{\sum_{\Omega}^{}\left( \lambda_{1,\Omega} - \lambda_{2,\Omega} \right)}{\sum_{\Omega}^{}\left( \lambda_{1,\Omega} + \lambda_{2,\Omega} \right)},\ \ E_{\Omega} = \lambda_{1,\Omega} + \lambda_{2,\Omega}
    \tag{A2.12}
    \label{eq:A2.12}
\end{equation}

Let \(C_{TFR}\) denote the coherency of the original TFR and \(C_{PERM}\) that of a reference map obtained by randomly permuting the pixel locations within the same map. The differences are defined as follows:

\begin{equation}
    \Delta C = C_{TFR} - C_{PERM}
    \tag{A2.13}
    \label{eq:A2.13}
\end{equation}

The thin gray and thick black lines in Figure~\ref{fig:figurea3} represent the individual bearings and their mean, respectively.

\begin{figure}[H]
	\centering
	\includegraphics[scale=0.5]{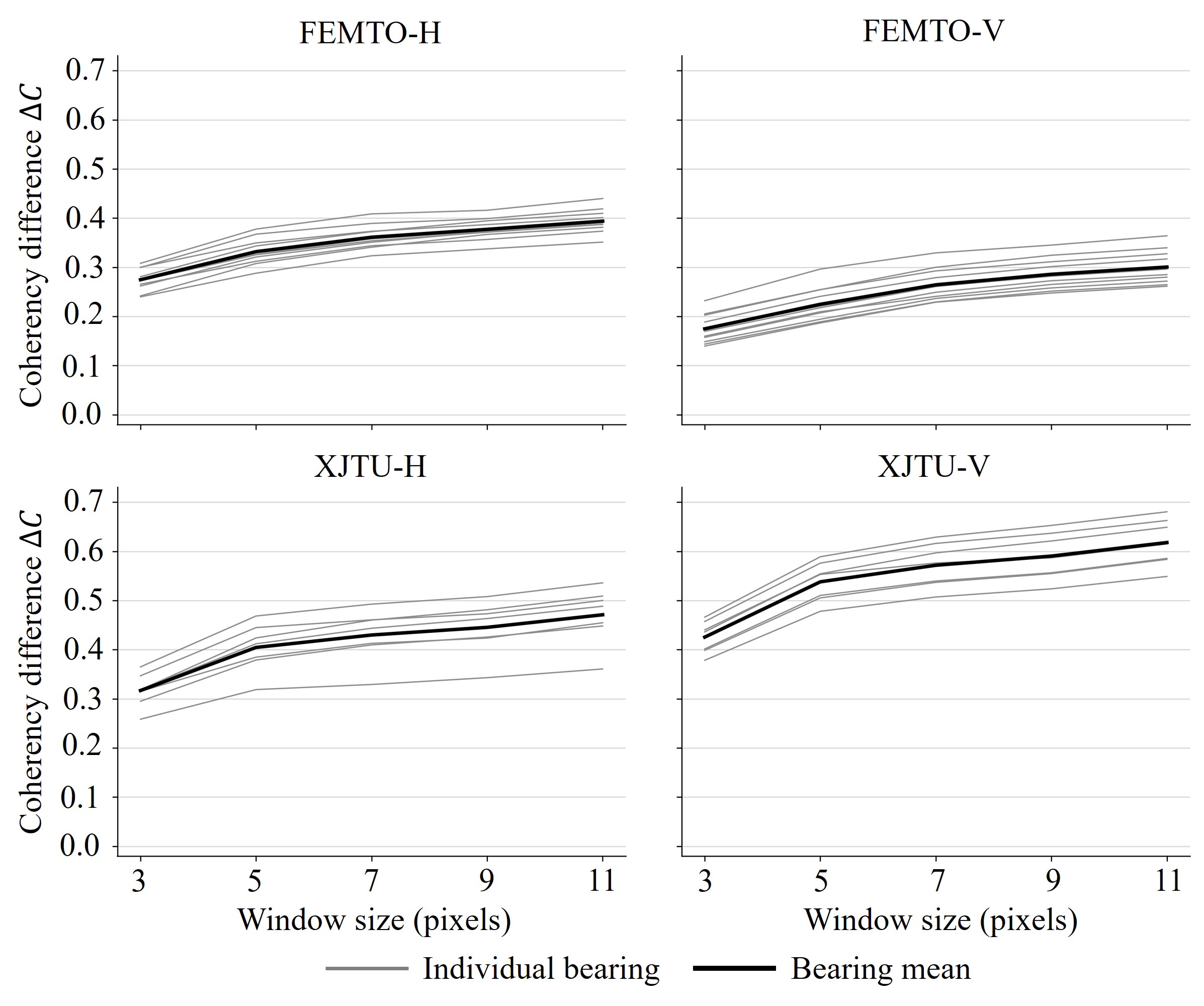}
	\caption{Difference in coherency between TFRs and pixel-permuted maps.}
	\label{fig:figurea3}
\end{figure}

For \(w = 5\), the mean values of \(\Delta C\) were 0.331, 0.224, 0.404, and 0.538 for FEMTO-H, FEMTO-V, XJTU-H, and XJTU-V, respectively. Across all combinations of 18 bearings, two measurement channels, and five window sizes, \(\Delta C > 0\). This indicates that, under the specified aggregation procedure, the original TFRs displayed higher local structure tensor coherency compared with the pixel-permuted maps, which maintained pixel values but disrupted spatial arrangement. This analysis characterizes spatial organization; however, it does not, in isolation, establish statistical significance or demonstrate that the observed structures encode degradation.

\section{Visualization of raw bearing vibration signals.}
\label{sec:appB}

\setcounter{figure}{0}
\renewcommand{\thefigure}{B\arabic{figure}}

\begin{figure}[H]
	\centering
	\includegraphics[width=\linewidth]{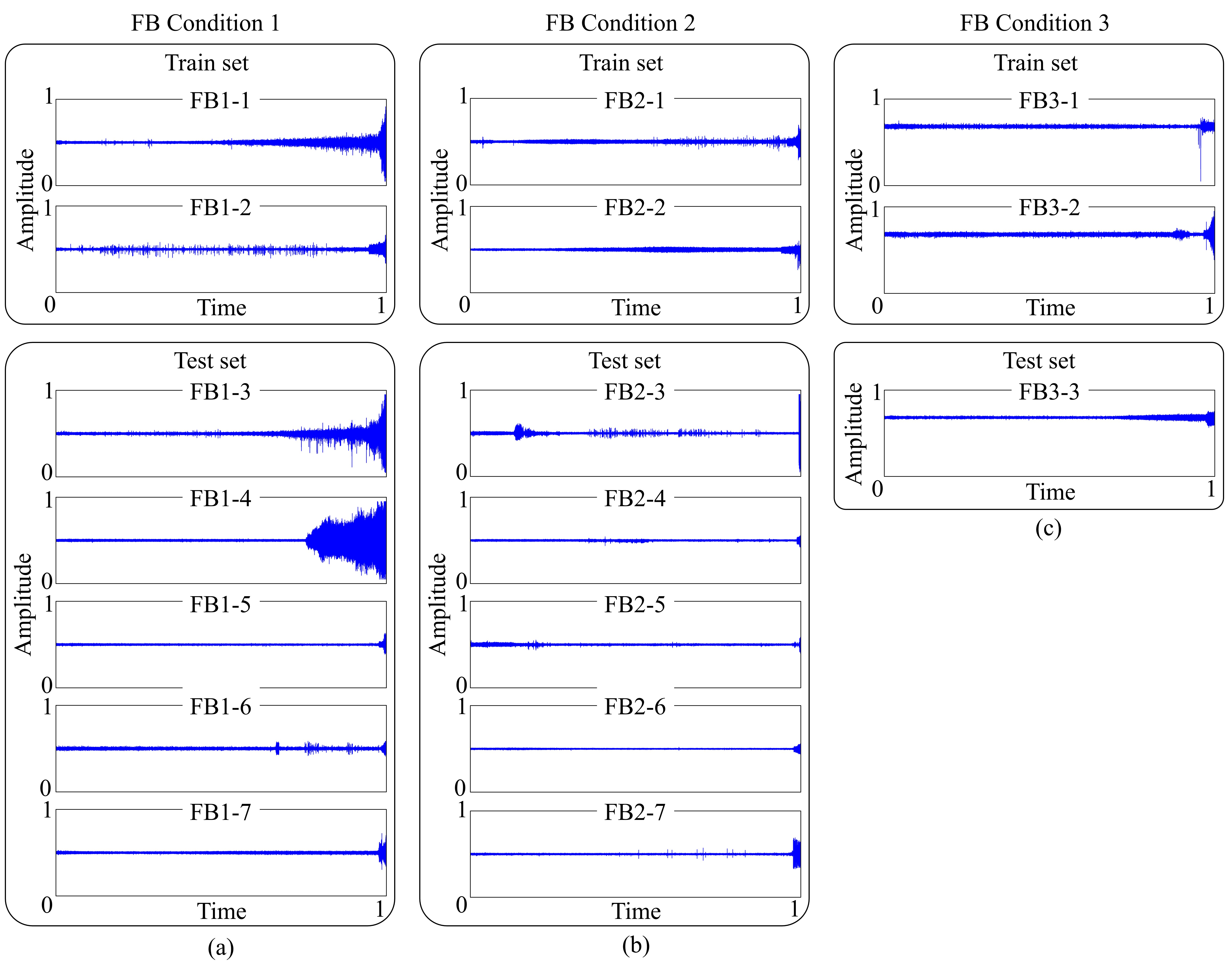}
	\caption{Raw vibration signals over the RtF for FB under three operating conditions.}
	\label{fig:figureb1}
\end{figure}

\begin{figure}[H]
	\centering
	\includegraphics[width=0.6\linewidth]{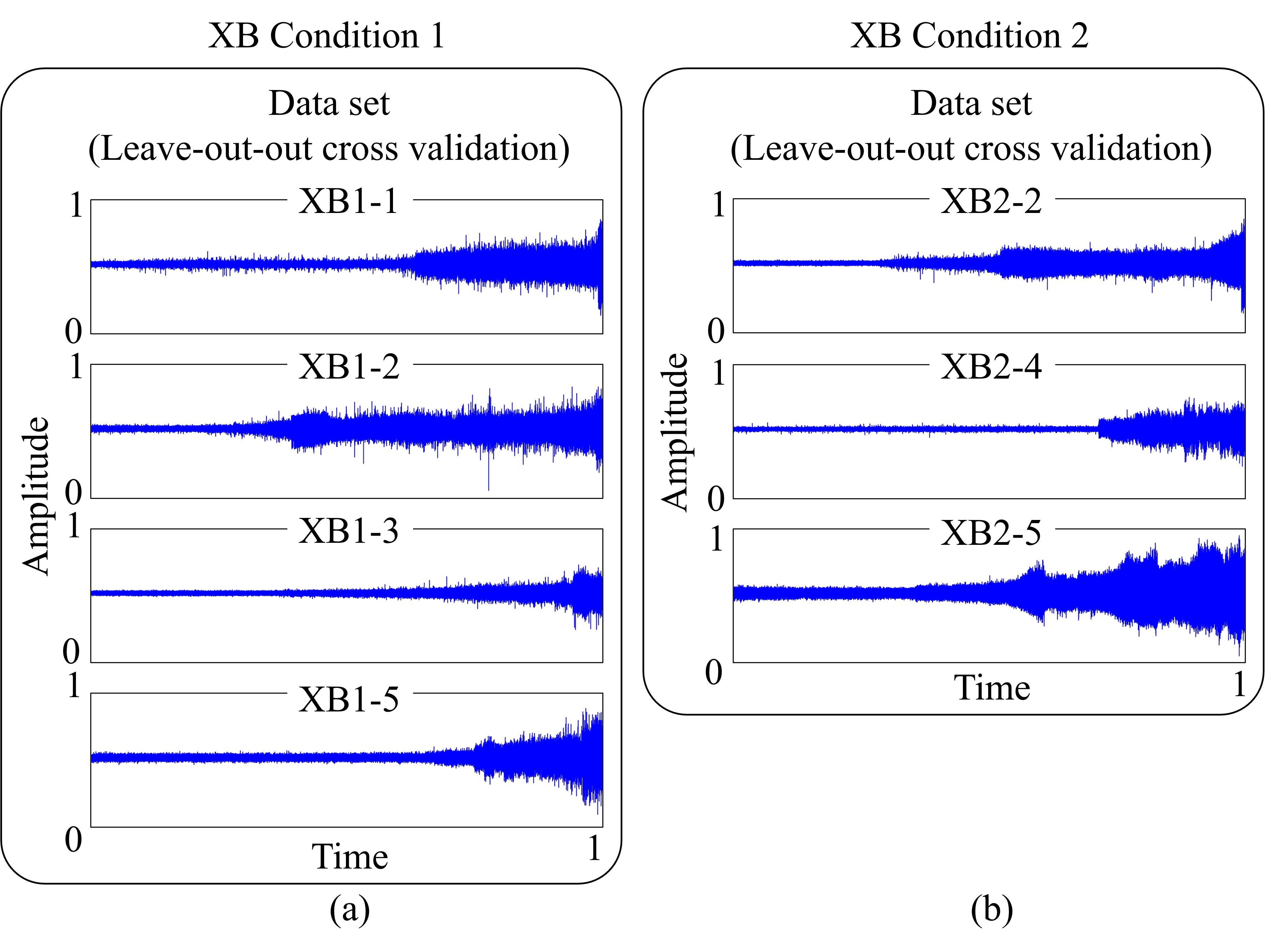}
	\caption{Raw vibration signals over the RtF for XB under two operating conditions.}
	\label{fig:figureb2}
\end{figure}

The vibration signals over RtF for all bearings in the FB and XB datasets are shown in Figures~\ref{fig:figureb1} and ~\ref{fig:figureb2}, respectively. Amplitude min-max scaling parameters were estimated from the training set and applied unchanged to the test set. For many bearings, the amplitude distribution evolved progressively as failure approached, although the trajectories were not uniformly monotonic. Certain bearings deviated from this general trend; for example, FB1-2 demonstrated recurring abrupt amplitude excursions, whereas FB2-3 displayed large amplitude fluctuations early in operation. Signal traces alone cannot establish whether these fluctuations result from measurement noise, operational transients or bearing conditions.

\section{FAAC-GRU architecture.}
\label{sec:appC}

The computational structure of FAAC-GRU for input TFR sequences of length \(L\) with \(C\) segments per observation is presented in Table~\ref{tab:c1}, where \(B\) denotes the batch size. In the reported configuration, \(C\) denotes the number of segments rather than the number of feature channels. The computational complexity of this architecture is presented in Table~\ref{tab:10}.

\setcounter{table}{0}
\renewcommand{\thetable}{C\arabic{table}}
\begin{table}[htbp]
    \centering
    \caption{FAAC-GRU layer configuration and output structures.}
    \label{tab:c1}
    \small
    \setlength{\tabcolsep}{5pt}
    \renewcommand{\arraystretch}{1.2}
    \begin{tabularx}{\linewidth}{@{}p{0.25\linewidth}Xp{0.27\linewidth}@{}}
        \toprule
        Section & Layer / Operation & Output shape \\
        \midrule
        Input & Segment-wise TFR sequence & (B, L, C, 1, 64, 64) \\
        Segment batching & Merge B, L, C for shared extraction & (BLC, 1, 64, 64) \\
        Shared FAAC feature extractor & MSAC block 1 & (BLC, 32, 64, 64) \\
        & (2, 2) max pooling & (BLC, 32, 32, 32) \\
        & MSAC block 2 & (BLC, 64, 32, 32) \\
        & (2, 2) max pooling & (BLC, 64, 16, 16) \\
        & MSAC block 3 & (BLC, 128, 16, 16) \\
        & DCBAM & (BLC, 128, 16, 16) \\
        & (1, 1) feature projection without batch normalization or activation & (BLC, 128, 16, 16) \\
        & DAP & (BLC, 128) \\
        Restore segment axis & Restore B, L, C & (B, L, C, 128) \\
        Segment fusion & Ordered concatenation & (B, L, Cx128) = (B, L, 640) \\
        Temporal encoder & Unidirectional GRU & (B, L, 128) \\
        & Layer normalization of the final hidden state & (B, 128) \\
        Regression head & Fully connected layer--ReLU--Dropout & (B, 128) \\
        & Fully connected layer--Sigmoid & (B, 1) \\
        Output & Output & (B, 1) \\
        \bottomrule
    \end{tabularx}
\end{table}

\end{document}